\pdfoutput=1
\documentclass[letterpaper, 10pt, conference]{ieeeconf}

\IEEEoverridecommandlockouts
\usepackage{graphics}
\usepackage{amsmath}
\usepackage{amssymb}

\usepackage{amsthm}
\usepackage{xcolor}
\usepackage{tikz}
\usetikzlibrary{arrows.meta, calc, positioning, decorations.pathreplacing, decorations.pathmorphing, patterns, backgrounds}
\usepackage[utf8]{inputenc}
\usepackage[T1]{fontenc}

\newcommand{\rgbd}{RGB\nobreakdash-D}

\newcommand{\sparsh}{SPARSH}
\newcommand{\digit}{DIGIT}

\theoremstyle{plain}
\newtheorem{theorem}{Theorem}
\newtheorem{proposition}[theorem]{Proposition}
\newtheorem{lemma}[theorem]{Lemma}
\newtheorem{corollary}[theorem]{Corollary}
\theoremstyle{definition}
\newtheorem{definition}{Definition}
\newtheorem{assumption}{Assumption}
\theoremstyle{remark}

\usepackage[skins,breakable]{tcolorbox}
\tcbset{statement/.style 2 args={enhanced, breakable, frame hidden,
    colback=#1, boxrule=0pt, arc=4pt,
    skin first=enhanced, skin middle=enhanced, skin last=enhanced,
    overlay={\begin{scope}
        \clip[rounded corners=4pt] (frame.south west) rectangle (frame.north east);
        \fill[#2] (frame.south west) rectangle ([xshift=1.8pt]frame.north west);
      \end{scope}},
    left=8pt, right=7pt, top=5pt, bottom=5pt, boxsep=0pt,
    before skip=7pt, after skip=7pt,
    before upper={\clubpenalty=10000 \widowpenalty=10000
      \displaywidowpenalty=10000 \predisplaypenalty=10000
      \postdisplaypenalty=0
      \csname @beginparpenalty\endcsname=10000}}}
\tcolorboxenvironment{definition}{statement={black!4}{black!45}}
\tcolorboxenvironment{assumption}{statement={yellow!12}{yellow!55!black}}
\tcolorboxenvironment{theorem}{statement={blue!6}{blue!55!black}}
\tcolorboxenvironment{proposition}{statement={teal!7}{teal!60!black}}
\tcolorboxenvironment{lemma}{statement={violet!6}{violet!55!black}}
\tcolorboxenvironment{corollary}{statement={green!7}{green!45!black}}
\usepackage{xurl}
\usepackage[colorlinks=true, linkcolor=blue!50!black, citecolor=blue!50!black,
            urlcolor=blue!50!black]{hyperref}

\newcommand{\R}{\mathbb{R}}
\newcommand{\fld}[1]{\boldsymbol{#1}}
\newcommand{\Obj}{\mathcal{O}}
\newcommand{\Pose}{\mathcal{P}_{s}}
\newcommand{\PoseT}{\hat{\mathcal{P}}_{s}}
\newcommand{\Grid}{\Lambda}
\newcommand{\Surf}{\Sigma}
\newcommand{\Dp}{\fld{D}}
\newcommand{\Sg}{\fld{S}}
\newcommand{\F}{\fld{F}}
\newcommand{\f}{\fld{F}_{s}}
\newcommand{\pen}{\fld{Z}}
\newcommand{\Phir}{\Phi_{\mathrm{render}}}
\newcommand{\Phil}{\Phi_{\mathrm{lift}}}
\newcommand{\Phic}{\Phi_{\mathrm{complete}}}
\newcommand{\Phid}{\Phi_{\mathrm{depth}}}
\newcommand{\Phitr}{\Phi_{\mathrm{track}}}
\newcommand{\Imgz}{\mathcal{Y}_0}
\newcommand{\Img}{\mathcal{Y}}
\newcommand{\Cam}{\mathcal{P}_{r}}
\newcommand{\Config}{\mathcal{G}}
\newcommand{\Objp}{\Obj_{\mathrm{p}}}
\newcommand{\SurfH}{\hat\Sigma}
\newcommand{\Phit}{\Phi_{\mathrm{touch}}}
\newcommand{\Phif}{\Phi_{\mathrm{force}}}
\newcommand{\norm}[1]{\lVert #1 \rVert}
\newcommand{\ninf}[1]{\lVert #1 \rVert_{\infty}}
\newcommand{\fhat}{\hat{\fld{F}}_{s}}
\newcommand{\fstar}{\fld{F}_{s}^{*}}

\hypersetup{pdftitle={Geo-Force: A Theory of Force Sensing from Object Geometry},
  pdfauthor={Abdullah Ayad}}

\title{\LARGE \bf
Geo-Force: A Theory of Force Sensing from Object Geometry
}

\author{%
  Abdullah Ayad%
  \thanks{A. Ayad (e-mail: {\tt\small abdallahayman772@gmail.com}) is an
  independent researcher. Part of this work was done while with the
  Department of CSAI, University of Technology Nuremberg, 90461 Nuremberg,
  Germany.}%
}

\begin{document}

\maketitle
\thispagestyle{empty}
\pagestyle{empty}

\begin{abstract}
Robots need contact forces to handle delicate objects, but tactile sensors
are fragile and hard to integrate, whereas depth sensors, such as
\rgbd{} cameras, are widely available. A virtual touch sensor that keeps the
sensing surface of a tactile sensor but replaces its electronics with a touch
signal predicted from the target object's geometry, as measured by a depth sensor,
could close this gap, yet it is unclear what such a virtual sensor computes
and when its output can be trusted. We give a formal answer. We define the
\emph{Geometric Force} as the force on each taxel of a virtual touch sensor,
computed in three steps: rendering the surface to be touched as seen from the
virtual sensor, predicting the touch signal with a touch map, learned or
simulated, and estimating the force from that signal. We prove three
properties of the Geometric Force. First, its force vectors only rotate
under a change of coordinate frame. Second, it vanishes when the rendered
surface does not touch the virtual sensor. Third, it is stable: its error
grows at most linearly with the errors of the three steps for Lipschitz
maps. The
rendering error, in turn, depends on how well the surface is sampled or
reconstructed and how accurately the sensor is placed. We then show a
fundamental limit: geometry does not reveal the object's stiffness, so no
touch map can predict touch exactly for objects of the same shape and different
stiffness. Under an elastic contact model, all object-specific knowledge
resides in the touch map. From these results, we derive requirements on the
depth sensor and the objects, design rules, and the conditions under which
such a virtual sensor fails.
\end{abstract}

\section{INTRODUCTION}

Touch-sensitive robotic tasks often rely on custom hardware such as
underactuated grippers, vacuum grippers, or fingertip tactile sensors, which
makes them system-specific. Compared to vision and language,
touch data in robotics remains scarce. Tactile perception is complicated by
the heterogeneity of sensor morphology, signal representation,
dimensionality, and processing pipelines~\cite{albini2025representingdatarobotictactile}.
Collecting consistent tactile data is hard as well: contact forces can
degrade both sensors and objects. Depth sensors, in contrast, are cheap,
standardized, and already mounted on most robots. We therefore ask:
\emph{what can a robot know about contact force from the geometry of an
object alone?}

Touch sensors, regardless of their measurement principle (e.g., optics,
magnetism, capacitance), measure interactions between an external object and a
compliant sensing surface. Optical tactile sensors such as
\digit{}~\cite{Lambeta2020DIGIT} have gained traction because they provide
rich spatial information, which motivated the notion of \emph{taxels},
analogous to pixels in images. Building, integrating, and calibrating such
sensors for every robotic platform remains a significant barrier to adoption.
Replacing them requires virtual alternatives of comparable fidelity, for which
physical simulators~\cite{wang2022tacto,si2022taxim} and cross-modal models
learned from easier-to-collect modalities~\cite{tactgen2025,wu_vitacgen_2025}
are candidate solutions. Cross-modal models that map vision or depth to touch
can, in principle, provide tactile feedback where only cameras are available,
which is attractive because \rgbd{} cameras capture 3D geometry directly.

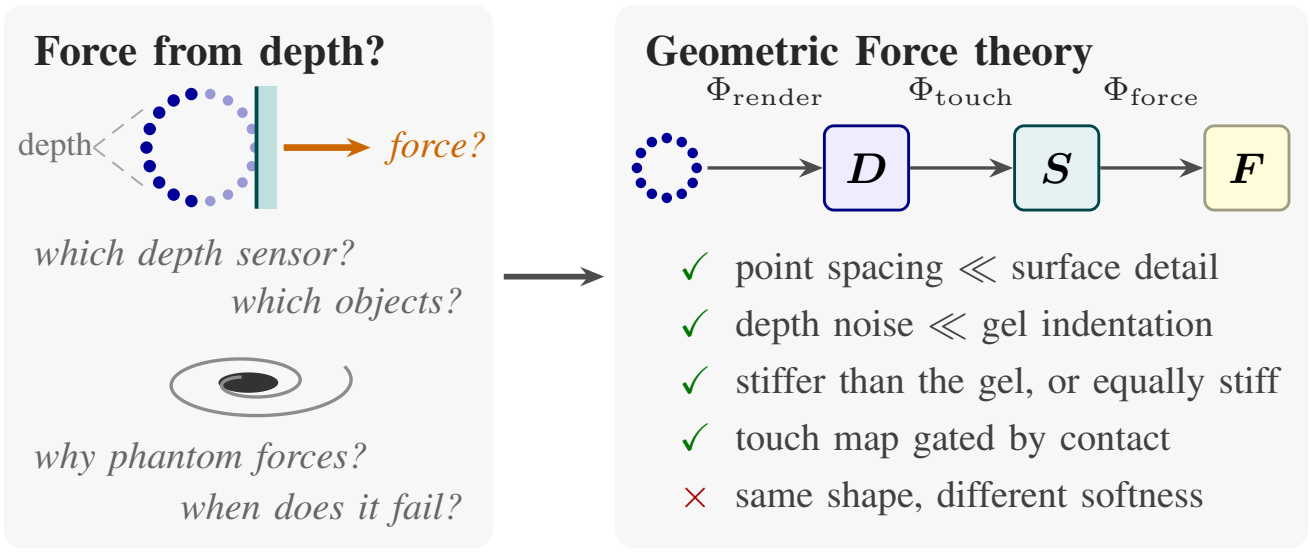
\begin{figure}[t]
\centering
\resizebox{\columnwidth}{!}{%
\begin{tikzpicture}[>={Stealth[length=5pt, width=4pt]},
    /utils/exec={%
      \colorlet{panelfill}{black!3}%
      \colorlet{depL}{blue!50!black}\colorlet{tchL}{teal!55!black}%
      \colorlet{frcL}{yellow!35!black!70}\colorlet{frcA}{orange!80!black}%
      \colorlet{yes}{green!45!black}\colorlet{no}{red!65!black}},
    head/.style={font=\small\bfseries, text=black!85, anchor=north west},
    q/.style={font=\footnotesize\itshape, text=black!60},
    box/.style={rounded corners=2pt, line width=0.6pt, minimum width=0.62cm,
                minimum height=0.62cm, font=\small},
    map/.style={font=\scriptsize, text=black!85},
    take/.style={font=\footnotesize, text=black!75, anchor=west}]
  \fill[panelfill, rounded corners=4pt] (0,0) rectangle (3.6,4.0);
  \fill[panelfill, rounded corners=4pt] (4.5,0) rectangle (9.6,4.0);
  \node[head] at (0.1,3.9) {Force from depth?};
  \node[head] at (4.6,3.9) {Geometric Force theory};
  \node[font=\scriptsize, text=black!55] at (0.38,2.95) {depth};
  \draw[black!35, densely dashed] (0.66,2.95) -- (1.05,3.25);
  \draw[black!35, densely dashed] (0.66,2.95) -- (1.05,2.65);
  \foreach \a in {100,120,...,260} {
    \fill[blue!60!black] ({1.45+0.4*cos(\a)},{2.95+0.4*sin(\a)}) circle (1.3pt);
  }
  \foreach \a in {-80,-60,...,80} {
    \fill[blue!60!black!40] ({1.45+0.4*cos(\a)},{2.95+0.4*sin(\a)}) circle (1.1pt);
  }
  \fill[teal!25] (1.86,2.5) rectangle (2.01,3.4);
  \draw[teal!60!black, thick] (1.86,2.5) -- (1.86,3.4);
  \draw[->, frcA, line width=1.4pt] (2.06,2.95) -- (2.7,2.95);
  \node[font=\footnotesize\itshape, text=frcA, anchor=west] at (2.72,2.95) {force?};
  \node[q, anchor=west] at (0.1,2.15)  {which depth sensor?};
  \node[q, anchor=east] at (3.5,1.8)  {which objects?};
  \fill[black!80] (1.8,1.22) ellipse (0.22 and 0.075);
  \draw[black!45, line width=0.7pt, domain=2:13, samples=200]
    plot ({1.8+0.06*\x*cos(\x r)}, {1.22+0.022*\x*sin(\x r)});
  \node[q, anchor=west] at (0.1,0.66)   {why phantom forces?};
  \node[q, anchor=east] at (3.5,0.28)  {when does it fail?};
  \draw[->, line width=1.2pt, black!70] (3.68,2.0) -- (4.42,2.0);
  \foreach \a in {0,30,...,330} {
    \fill[blue!60!black] ({4.88+0.22*cos(\a)},{2.8+0.22*sin(\a)}) circle (1.1pt);
  }
  \coordinate (in) at (5.18,2.8);
  \node[box, fill=blue!7, draw=depL] (D) at (6.35,2.8) {\(\Dp\)};
  \node[box, fill=teal!10, draw=tchL] (S) at (7.75,2.8) {\(\Sg\)};
  \node[box, fill=yellow!18, draw=frcL] (F) at (9.15,2.8) {\(\F\)};
  \draw[->, line width=0.7pt, black!70] (in) -- (D) node[midway, above=9.5pt, map] {\(\Phir\)};
  \draw[->, line width=0.7pt, black!70] (D) -- (S) node[midway, above=9.5pt, map] {\(\Phit\)};
  \draw[->, line width=0.7pt, black!70] (S) -- (F) node[midway, above=9.5pt, map] {\(\Phif\)};
  \node[take] at (4.85,2.05) {\textcolor{yes}{\(\checkmark\)}\; point spacing \(\ll\) surface detail};
  \node[take] at (4.85,1.63) {\textcolor{yes}{\(\checkmark\)}\; depth noise \(\ll\) gel indentation};
  \node[take] at (4.85,1.21) {\textcolor{yes}{\(\checkmark\)}\; stiffer than the gel, or equally stiff};
  \node[take] at (4.85,0.79) {\textcolor{yes}{\(\checkmark\)}\; touch map gated by contact};
  \node[take] at (4.85,0.37) {\textcolor{no}{\(\times\)}\; same shape, different softness};
\end{tikzpicture}}
\caption{Escaping the depth-to-force rabbit hole. Estimating the force on a
touch sensor from depth measurements raises open-ended questions (left):
which depth sensor, which objects, why phantom forces, and when the estimate fails.
The Geometric Force theory reduces these four questions to one chain (right):
target object's geometry \(\rightarrow\) rendered depth patch
\(\rightarrow\) predicted touch signal \(\rightarrow\) force. Analyzing this chain sets requirements on the depth sensor and the
objects through its error bounds, removes phantom forces by gating the touch
map with contact (\(\checkmark\)), and identifies stiffness,
invisible to geometry, as the cause of failure (\(\times\)).}
\label{fig:rabbithole}
\end{figure}
\begin{figure*}[t]
\centering
\begin{tikzpicture}[scale=0.9, every node/.append style={transform shape},
    >={Stealth[length=4pt, width=3.5pt]},
    /utils/exec={%
      \colorlet{panel}{black!3}%
      \colorlet{objfill}{orange!15}\colorlet{objline}{orange!55!black}%
      \colorlet{rigidfill}{orange!40}%
      \colorlet{gelfill}{teal!12}\colorlet{gelline}{teal!60!black}%
      \colorlet{depL}{blue!50!black}\colorlet{tchL}{teal!55!black}%
      \colorlet{frcL}{yellow!35!black!70}\colorlet{frcA}{orange!80!black}%
      \colorlet{no}{red!65!black}},
    head/.style={font=\small\bfseries, text=black!85, anchor=north west},
    lbl/.style={font=\scriptsize, text=black!65, align=center},
    ptitle/.style={font=\small, text=black!85, anchor=south west},
    pdesc/.style={font=\scriptsize, text=black!65, anchor=north west, align=left},
    flow/.style={->, line width=0.8pt, draw=black!70},
    map/.style={font=\scriptsize, text=black!85}]
  \foreach \xa/\xb in {0/4.2, 4.4/10.6, 10.8/15.6, 15.8/19.6} {
    \fill[panel, rounded corners=4pt] (\xa,0) rectangle (\xb,4.4);
  }
  \node[head] at (0.12,4.3)  {(a) Setup};
  \node[head] at (4.52,4.3)  {(b) Geometric Force};
  \node[head] at (10.92,4.3) {(c) Guarantees};
  \node[head] at (15.92,4.3) {(d) Stiffness};
  \coordinate (lens) at ($(0.9,3.0)+(-43.6:0.22)$);
  \fill[blue!8] (lens) -- ++(-31:2.8) -- ($(lens)+(-56:2.8)$) -- cycle;
  \begin{scope}[shift={(0.9,3.0)}, rotate=-43.6]
    \fill[black!65, rounded corners=1pt] (-0.42,-0.17) rectangle (0.1,0.17);
    \fill[black!45] (0.1,-0.1) rectangle (0.22,0.1);
  \end{scope}
  \node[lbl] at (0.72,2.25) {depth sensor};
  \fill[black!22] (2.2,2.15) rectangle (3.7,2.4);
  \fill[black!22] (2.2,0.45) rectangle (2.45,2.15);
  \fill[black!22] (3.45,0.45) rectangle (3.7,2.15);
  \fill[black!22] (2.85,2.4) rectangle (3.05,2.75);
  \fill[objfill] (2.95,1.05) circle (0.5);
  \draw[objline, thick] (2.95,1.05) circle (0.5);
  \foreach \xa/\xb in {2.45/2.55, 3.35/3.45} {
    \fill[teal!30] (\xa,0.7) rectangle (\xb,1.4);
    \draw[gelline, densely dashed] (\xa,0.7) rectangle (\xb,1.4);
  }
  \node[lbl, anchor=east] (vts) at (1.95,0.75) {virtual touch\\sensors};
  \draw[black!45, thin] (vts.east) -- (2.45,0.95);
  \fill[black!4] (4.7,1.9) rectangle (5.5,2.7);
  \shade[inner color=black!45, outer color=black!4] (5.1,2.3) ellipse (0.3 and 0.25);
  \draw[black!50, line width=0.6pt] (4.7,1.9) rectangle (5.5,2.7);
  \fill[blue!5] (6.3,1.9) rectangle (7.1,2.7);
  \shade[inner color=blue!55, outer color=blue!5] (6.7,2.3) circle (0.32);
  \draw[depL, line width=0.6pt] (6.3,1.9) rectangle (7.1,2.7);
  \fill[teal!5] (7.9,1.9) rectangle (8.7,2.7);
  \shade[inner color=teal!65, outer color=teal!5] (8.3,2.3) ellipse (0.32 and 0.27);
  \draw[tchL, line width=0.6pt] (7.9,1.9) rectangle (8.7,2.7);
  \fill[yellow!10] (9.5,1.9) rectangle (10.3,2.7);
  \foreach \i/\j/\len in {-0.1/-0.1/0.18, -0.1/0.1/0.18, 0.1/-0.1/0.18, 0.1/0.1/0.18,
      -0.3/-0.1/0.08, -0.3/0.1/0.08, 0.3/-0.1/0.08, 0.3/0.1/0.08,
      -0.1/-0.3/0.08, 0.1/-0.3/0.08, -0.1/0.3/0.08, 0.1/0.3/0.08} {
    \draw[->, frcA, line width=0.6pt]
      ({9.9+\i-0.3*\len},{2.3+\j-0.5*\len}) -- ++({0.6*\len},{\len});
  }
  \foreach \i/\j in {-0.3/-0.3, -0.3/0.3, 0.3/-0.3, 0.3/0.3} {
    \fill[frcA] ({9.9+\i},{2.3+\j}) circle (0.6pt);
  }
  \draw[frcL, line width=0.6pt] (9.5,1.9) rectangle (10.3,2.7);
  \draw[flow] (5.55,2.3) -- (6.25,2.3) node[midway, above=12.5pt, map] {\(\Phir\)};
  \draw[flow] (7.15,2.3) -- (7.85,2.3) node[midway, above=12.5pt, map] {\(\Phit\)};
  \draw[flow] (8.75,2.3) -- (9.45,2.3) node[midway, above=12.5pt, map] {\(\Phif\)};
  \foreach \cx/\name in {5.1/depth image, 6.7/depth patch, 8.3/touch signal, 9.9/force} {
    \node[lbl] at (\cx,1.5) {\strut\name};
  }
  \foreach \cy in {3.13, 1.88, 0.63} {
    \fill[white, rounded corners=3pt] (11.0,{\cy-0.48}) rectangle (12.3,{\cy+0.48});
    \draw[black!15, rounded corners=3pt] (11.0,{\cy-0.48}) rectangle (12.3,{\cy+0.48});
  }
  \begin{scope}[shift={(11.42,2.83)}]
    \draw[->, black!60] (0,0) -- (0.48,0);
    \draw[->, black!60] (0,0) -- (0,0.48);
    \draw[->, frcA, line width=0.8pt] (0.1,0.08) -- ++(0.17,0.28);
    \draw[->, black!45, line width=0.6pt] (5:0.62) arc (5:85:0.62);
  \end{scope}
  \node[ptitle, anchor=west, align=left] at (12.55,3.13) {rotates with the\\coordinate frame};
  \begin{scope}[shift={(11.65,1.88)}]
    \fill[gelfill] (-0.5,-0.4) rectangle (0.5,-0.26);
    \draw[gelline, thick] (-0.5,-0.26) -- (0.5,-0.26);
    \fill[objfill] (0,0.12) circle (0.26);
    \draw[objline, thick] (0,0.12) circle (0.26);
  \end{scope}
  \node[ptitle, anchor=west] at (12.55,1.88) {no contact, no force};
  \begin{scope}[shift={(11.65,0.63)}]
    \fill[frcA!15] (-0.42,-0.32) -- (0.4,0.3) -- (0.4,-0.32) -- cycle;
    \draw[->, black!60] (-0.42,-0.32) -- (0.5,-0.32);
    \draw[->, black!60] (-0.42,-0.32) -- (-0.42,0.38);
    \draw[frcA, line width=0.8pt] (-0.42,-0.32) -- (0.4,0.3);
  \end{scope}
  \node[ptitle, anchor=west, align=left] at (12.55,0.63) {force error \(\le\) sum of\\weighted map errors};
  \foreach \cx/\name in {16.75/rigid, 18.65/soft} {
    \node[lbl] at (\cx,2.75) {\strut\name};
  }
  \begin{scope}[shift={(16.75,1.55)}, scale=0.7]
    \fill[gelfill] (-1.1,-0.35) rectangle (1.1,0);
    \draw[gelline, thick] (-1.1,0) -- (-0.397,0);
    \draw[gelline, thick] (0.397,0) -- (1.1,0);
    \fill[rigidfill] (0,0.45) circle (0.6);
    \draw[objline, thick] (0,0.45) circle (0.6);
  \end{scope}
  \begin{scope}[shift={(18.65,1.55)}, scale=0.7]
    \fill[gelfill] (-1.1,-0.35) rectangle (1.1,0);
    \draw[gelline, thick] (-1.1,0) -- (-0.346,0);
    \draw[gelline, thick] (0.346,0) -- (1.1,0);
    \fill[objfill] ($(0,0.45)+(-54.8:0.6)$) arc (-54.8:234.8:0.6) -- cycle;
    \draw[objline, thick] ($(0,0.45)+(-54.8:0.6)$) arc (-54.8:234.8:0.6) -- cycle;
  \end{scope}
  \node[font=\LARGE, text=no] at (17.7,1.85) {\(=\)};
  \node[lbl] at (17.7,0.65) {same predicted force,\\different real force};
  \fill[panel, rounded corners=4pt] (0,-2.4) rectangle (19.6,-0.2);
  \node[head, anchor=west] at (0.12,-1.3) {(e) Takeaways};
  \foreach \y/\title/\desc in {
      -0.5/Planar/{point spacing \(\ll\) surface detail},
      -0.875/Depth/{noise \(\ll\) gel indentation},
      -1.25/Objects/{no sharp edges; much stiffer than the gel, or all equally stiff},
      -1.625/Touch map/{gated by contact},
      -2.0/Fails on/{shallow contact; same shape, different softness}} {
    \node[font=\footnotesize, text=black!85, anchor=east] at (8.2,\y) {\title:};
    \node[font=\footnotesize, text=black!65, anchor=west] at (8.3,\y) {\desc};
  }
\end{tikzpicture}
\caption{Overview. (a)~A depth sensor observes the target object held between
virtual touch sensors. (b)~The Geometric Force maps the depth image to a depth
patch, then to a touch signal, then to a force. (c)~It rotates with the
coordinate frame, vanishes without contact, and has an error of at most a
weighted sum of the errors of its three maps (Propositions~\ref{prop:equivariance}
and~\ref{prop:vanishing}, Theorem~\ref{thm:stability}). (d)~Objects of the same shape
get the same predicted force, even when their stiffness, and hence their real
force, differs (Proposition~\ref{prop:irreducible}). (e)~Practical takeaways of the theory.}
\label{fig:overview}
\end{figure*}

A virtual touch sensor keeps the gel of a real touch sensor, so the object
still presses against the same soft surface, with the same force as on the
real sensor. Only the electronics that read the gel's
deformation are replaced, by a touch signal predicted from the target
object's geometry, as measured by a depth sensor. From this signal, a
pretrained decoder such as \sparsh{}~\cite{higuera_sparsh_2024} estimates the
force, just as it does from the signal of the real sensor. Unlike
methods that regress force or pressure directly from
images~\cite{grady2022pressurevision,collins2023visual}, a virtual touch
sensor thus reports the force on the taxels of a real sensor. It can
therefore reuse the force decoder and the touch data recorded with that
sensor. Before such a virtual sensor can be trusted, e.g., to
grasp a raw egg, one must know what its force means, which depth sensors it
needs, which objects it works on, and when it fails. None of the four has a
formal answer yet.

We propose to reason about any such virtual touch sensor through one
quantity, the \emph{Geometric Force}: the force on each taxel of a touch
sensor, predicted from the geometry of the object alone. The ideal
Geometric Force uses the touch map with the smallest mean squared error
between the predicted and the true touch signal. It thus marks the upper bound of what geometry
can reveal about touch, separating the errors that better geometry can remove
from those that only touch data can. In short, a virtual touch sensor sees
the shape of a surface, down to its texture, but not how it yields.

This work makes six contributions:
\begin{itemize}
  \item \textbf{Formal definition.} We define the Geometric Force as a composition of
    three maps, rendering, touch prediction, and force estimation, on a shared
    pixel grid. The touch map is either learned from real
    tactile data or simulated.
  \item \textbf{Structural properties.} We prove that the Geometric Force is
    equivariant to the coordinate frame whenever rendering is frame-invariant,
    and give conditions, met by any contact-gated touch map, under which
    it vanishes exactly out of contact.
  \item \textbf{Stability bounds.} For Lipschitz touch and force maps, the error of
    the Geometric Force grows at most linearly with the errors of its three
    maps. The rendering error, in turn, grows at most linearly with the fill distance
    and sampling precision of the measured points and with the touch-sensor
    pose error.
  \item \textbf{Stiffness limit.} Shape alone cannot determine touch
    for objects of unknown stiffness. The resulting irreducible touch error grows with
    the range of the object-to-gel stiffness ratio. Under
    an elastic-foundation model, only the touch map depends on the object.
  \item \textbf{Real-to-sim-to-real example.} We realize rendering for a hidden surface by
    shape completion and 6D tracking, and show that its completion and tracking
    errors replace the sampling errors in the stability bounds.
  \item \textbf{Practical takeaways.} We derive requirements on the depth
    sensor and the objects. The depth sensor needs point spacing finer than
    the surface detail and depth noise well below the gel indentation. The
    objects should keep their shape until contact, have no sharp edges, and be
    much stiffer than the gel or all about equally stiff. We
    also state when a virtual sensor fails.
\end{itemize}
This paper is theoretical. Experiments that test its results are left to
future work.

\section{BACKGROUND}
\label{sec:background}

The Geometric Force theory builds on one idea: a sensor can tell physical
states apart only as far as its outputs differ. We first trace this idea from
sensing in general to touch. Whatever their physical principle, tactile
sensors enter the theory only through a grid of taxels. We then review the
methods that realize the three maps (Fig.~\ref{fig:pipeline}) and the
alternatives that bypass them.

\textbf{Sensing as a many-to-one map.}
A sensor maps the physical state of the world to an output. Many states
produce the same output, so a sensor can only tell states apart up to this
ambiguity~\cite{lavalle2012sensing}. Whether one sensor can stand in for
another then becomes a question of information: does it separate the states
that a task needs separated~\cite{donald1995information,lavalle2019sensor}?
Prior knowledge can make a sparse sensor sufficient: a force-torque
sensor at the base of a fingertip of known shape also locates a single contact
on the fingertip~\cite{bicchi1993contact}. A virtual touch sensor stands in for a real one, so the same
question applies. It measures only geometry and
predicts touch through a learned or simulated touch map, so it can tell
objects apart only where their geometry differs. This is its limit: objects of the same shape and
different stiffness produce the same geometry, yet a real touch sensor tells
them apart by pressing~\cite{yuan2017hardness}. No touch map can resolve this
ambiguity. The error it leaves is aleatoric~\cite{hullermeier2021aleatoric}:
more training data cannot remove it. The Geometric Force theory defines what a virtual touch sensor computes, makes this limit precise, and bounds every other error.

\textbf{Tactile sensing principles.}
Tactile sensors turn the deformation of a compliant surface into a signal
through different physical effects~\cite{dahiya2010tactile}. Optical sensors
such as \digit{}~\cite{Lambeta2020DIGIT} image a gel from inside; magnetic
skins~\cite{bhirangi2021reskin} measure the field of magnetized particles
embedded in an elastomer; capacitive and piezoresistive arrays measure local
changes in capacitance or resistance. Multimodal fingertips combine several of
these effects~\cite{lambeta_digitizing_2024}. Their raw signals take
different forms, e.g., a high-resolution image for optical sensors or a few
three-axis readings for magnetic ones. The theory casts any of them onto a
grid of taxels, so its definitions and bounds apply to all of them; the optical \digit{} sensor serves as the running example.

\textbf{Rendering the surface to be touched.}
At contact, the touched surface is typically hidden from the depth sensor. It
can be reconstructed while still visible, by shape completion from a single
image~\cite{sam3dteam2025sam3d3dfyimages} or by Gaussian splatting~\cite{kerbl2023gaussian}, then carried to the instant of contact by
6D pose tracking~\cite{wen2024foundationpose} or point-cloud
registration~\cite{park2017colored}. Such pipelines realize the rendering map.
We show that their errors reach the force only through the depth-patch error, which we bound by the completion and tracking errors.

\textbf{Predicting the touch signal.}
For an optical sensor, contact appears as the change of an image against a
background recorded while the sensor is free. Cross-modal methods
learn to predict such signals from
images~\cite{li2019connecting,ayad_imagine2touch_2024,tactgen2025}, also with
training in simulation~\cite{wu_vitacgen_2025}.
Cross-sensor methods translate the signal of one tactile sensor into that of
another~\cite{rodriguez_touch2touch_2024}. Tactile simulators render them from the geometry of a rigid contacting
object~\cite{wang2022tacto,si2022taxim} or simulate soft objects whose
material is given~\cite{si2024difftactile}; the rigid ones enable sim-to-real
policy learning~\cite{lin_tactile_2022,su_sim2real_tactile}. Learned and
simulated predictors alike realize the touch map. The learned ones show empirically that touch can be predicted on their
data, but not what such a prediction can recover, how its errors propagate,
or when it must fail. We provide this characterization. In particular, any touch map, learned or
simulated, mispredicts objects whose stiffness differs from the one it
assumes; a rigid simulator, e.g., overestimates the indentation of soft
objects.

\textbf{Estimating the force.}
Each of these sensors yields the contact force through a decoder of its
signal. For optical sensors, the decoder inverts the gel deformation, by
inverse finite elements~\cite{ma2019dense} or by learned
regression~\cite{yuan2017gelsight}. Recent decoders build on self-supervised
touch representations, which also bridge different
sensors~\cite{zhao_transferable_2024,hou_unitac-nv_2025}.
\sparsh{}~\cite{higuera_sparsh_2024} decodes a force field from the
background-subtracted image of an optical sensor. Sparsh-skin~\cite{sharma_self-supervised_2025} and
\sparsh{}-X~\cite{higuera2025sparshx} learn such representations for magnetic
skins and multimodal fingertips, respectively. Such decoders
realize the force map and are specific to the sensor, not to the touched
object. We show that under elastic contact the force map is object-agnostic, so a pretrained decoder carries over to a virtual touch
sensor unchanged.

\textbf{Estimating force without touch.}
Vision-based methods regress contact force or pressure from images, e.g.,
for human hands~\cite{grady2022pressurevision} or robot
grippers~\cite{collins2023visual}. In robot-assisted surgery, the force is
instead computed from the observed tissue deformation through a
biomechanical model whose stiffness is given~\cite{haouchine2018vision}.
A virtual touch sensor reports the force on the taxels of a real sensor
instead. We show that geometry determines this force only up to the
object's stiffness, which the surgical approach supplies as a prior.

\section{PRELIMINARIES}
\label{sec:conventions}

We set out the notation used throughout the paper. Proofs longer than a few
lines are deferred to the Appendix.

\textbf{Typefaces.} Scalars are italic, vectors bold lowercase, and matrices
bold uppercase. Fields, i.e., functions that assign a value to every point of a
domain, such as every pixel of an image, are bold italic capitals, e.g.,
\(\Dp\), whether scalar-valued (\emph{scalar fields}) or vector-valued
(\emph{vector fields}). On a pixel grid, a scalar field is a matrix with one
entry per pixel; a vector field is a three-dimensional array with one vector
per pixel. Uppercase Greek letters denote infinite sets, e.g., a surface
\(\Surf \subset \R^3\); calligraphic letters denote finite sets and tuples,
e.g., \(\Obj\) and \(\Pose\). The pixel grid \(\Grid\), a finite set, keeps
its customary Greek letter. A hat marks an
estimate, a tilde a perturbed counterpart, and a star an ideal reference. A bar
on a vector \(\mathbf{y} = (y_1, y_2, y_3)^{\top} \in \R^{3}\) drops its last
entry, \(\bar{\mathbf{y}} = (y_1, y_2)^{\top} \in \R^{2}\); on a map \(\phi\)
into \(\R^{3}\), it drops the last entry of each value,
\(\bar{\phi}(x) = \overline{\phi(x)}\).

\textbf{Maps.} \(\phi : A \to B\) denotes a map that assigns to each element of
\(A\) one element of \(B\), and \(\phi(A) = \{\phi(x) : x \in A\}\) the image
of a set. Composition is applied right to left:
\((\phi \circ \psi)(x) = \phi(\psi(x))\). The indicator
\(\mathbf{1}[\cdot]\) equals \(1\) where its condition holds and \(0\)
elsewhere.

\textbf{Rigid motions.} \(\mathrm{SE}(3)\) is the group of rigid motions of
\(\R^3\), i.e., rotations followed by translations. A pose
\((\mathbf{R}, \mathbf{t}) \in \mathrm{SE}(3)\) consists of a rotation matrix
\(\mathbf{R}\), with \(\mathbf{R}^{\top}\mathbf{R} = \mathbf{I}\) and hence
\(\mathbf{R}^{-1} = \mathbf{R}^{\top}\), and a translation \(\mathbf{t}\); it
maps a point \(\mathbf{x}\) to \(\mathbf{R}\mathbf{x} + \mathbf{t}\). The
\emph{rotation angle} \(\theta \in [0, \pi]\) of a rotation \(\mathbf{Q}\) is
the angle by which it turns space about its axis; it satisfies
\(\norm{(\mathbf{Q} - \mathbf{I})\mathbf{y}} \le \theta\norm{\mathbf{y}}\)
for every vector \(\mathbf{y}\). Rigid motions preserve all distances.

\textbf{Norms.} \(\norm{\mathbf{x}}_2 = (\sum_i x_i^2)^{1/2}\) is the
Euclidean length of a vector; we write \(\norm{\cdot}\) when no other norm is
meant. For a field \(\fld{X} : \Grid \to \R^k\),
the max norm
\(\ninf{\fld{X}} = \max_{(a,b)} \norm{\fld{X}(a,b)}\) is the length of its
longest vector over all pixels. For the difference of two fields, it is
their largest pixel-wise error. The Frobenius norm
\(\norm{\fld{X}}_F = \bigl(\sum_{(a,b)} \norm{\fld{X}(a,b)}^2\bigr)^{1/2}\)
combines the vectors of all pixels into one length.

\textbf{Distances between sets.} The distance from a point \(\mathbf{x}\) to a
set \(B\), \(\inf_{\mathbf{y} \in B}\norm{\mathbf{x} - \mathbf{y}}\), is how
far \(\mathbf{x}\) lies from the nearest point of \(B\). The \emph{Hausdorff distance}
\begin{equation}
  d_H(A, B) = \max\Bigl\{\sup_{\mathbf{x} \in A}\inf_{\mathbf{y} \in B}
  \norm{\mathbf{x} - \mathbf{y}},\;
  \sup_{\mathbf{y} \in B}\inf_{\mathbf{x} \in A}
  \norm{\mathbf{x} - \mathbf{y}}\Bigr\}
\end{equation}
is the largest distance from a point of either set to the other set; two
surfaces with \(d_H \le \eta\) differ nowhere by more than \(\eta\). Here,
\(\sup\) and \(\inf\) are the least upper and greatest lower bounds, which
equal \(\max\) and \(\min\) when these are attained.

\textbf{Lipschitz continuity.} A map \(\phi\) is \emph{\(L\)-Lipschitz} with
respect to a norm if
\(\norm{\phi(x) - \phi(y)} \le L\norm{x - y}\) for all inputs \(x, y\): a
change in the input changes the output by at most \(L\) times as much. For example, an
\(L\)-Lipschitz function \(\gamma : \R^2 \to \R\) has slope at most \(L\).

\textbf{Expectations.} For random quantities, \(\mathbb{E}[\cdot]\) is the
expectation, i.e., the average over the data distribution. For random fields \(\Sg\) and \(\Dp\), the conditional expectation
\(\mathbb{E}[\Sg \mid \Dp]\) is the average of \(\Sg\) over all samples that
share a given value of \(\Dp\). The trace of the conditional covariance,
\(\operatorname{tr}\operatorname{Cov}(\Sg \mid \Dp)\), sums the variances of
the components of \(\Sg\) around this average; it measures how much \(\Sg\)
still varies once \(\Dp\) is known. \(\operatorname{Var}\) denotes the variance of a scalar.

\section{GEOMETRIC FORCE}
\label{sec:theory}

\begin{figure*}[t]
\centering
\begin{tikzpicture}[>={Stealth[length=5pt, width=4pt]},
    /utils/exec={%
      \colorlet{meas}{black!6}\colorlet{measL}{black!45}%
      \colorlet{geo1}{orange!7}\colorlet{geo2}{orange!13}\colorlet{geo3}{orange!20}%
      \colorlet{geoL}{orange!55!black}%
      \colorlet{dep}{blue!7}\colorlet{depL}{blue!50!black}%
      \colorlet{tch}{teal!10}\colorlet{tchL}{teal!55!black}%
      \colorlet{frc}{yellow!18}\colorlet{frcL}{yellow!35!black!70}%
      \colorlet{call}{blue!3.5}\colorlet{callL}{blue!22}%
      \colorlet{arr}{black!70}},
    box/.style={rounded corners=3pt, line width=0.6pt, minimum width=1.35cm,
                minimum height=0.82cm, align=center, font=\small},
    pill/.style={rounded corners=5pt, line width=0.5pt, draw=black!35,
                fill=white, minimum width=0.85cm, minimum height=0.58cm,
                font=\small},
    flow/.style={->, line width=0.8pt, draw=arr},
    sub/.style={font=\scriptsize, align=center, text=black!60},
    map/.style={font=\scriptsize, text=black!85}]
  \def\py{-3.15}
  \node[box, fill=meas, draw=measL] (in) at (0.3,0)    {\(\Config = (\Imgz, \Img, \Cam, \Pose)\)};
  \node[box, fill=dep,  draw=depL]  (D)  at (10.1,0) {\(\Dp\)};
  \node[box, fill=tch,  draw=tchL]  (S)  at (12.4,0) {\(\Sg\)};
  \node[box, fill=frc,  draw=frcL]  (F)  at (14.7,0){\(\F\)};
  \draw[flow] (in) -- (D) node[midway, above=1pt, map] {\(\Phir\)};
  \draw[flow] (D)  -- (S) node[midway, above=1pt, map] {\(\Phit\)};
  \draw[flow] (S)  -- (F) node[midway, above=1pt, map] {\(\Phif\)};
  \node[sub, below=3pt of in] {depth measurements\\and sensor poses};
  \node[sub, below=3pt of D]  {depth patch};
  \node[sub, below=3pt of S]  {touch signal\\(learned or simulated)};
  \node[sub, below=3pt of F]  {force field\\(sensor-specific)};
  \begin{scope}[on background layer]
    \fill[call, rounded corners=5pt] (-1.15,-1.95) rectangle (13.6,-5.85);
    \fill[tch!60] (11.32,-2.4) rectangle (13.45,-5.7);
    \draw[callL, line width=0.5pt, rounded corners=5pt]
          (-1.15,-1.95) rectangle (13.6,-5.85);
  \end{scope}
  \node[font=\scriptsize, text=blue!45!black, anchor=north west]
        at (-0.98,-2.04) {\(\Phit \circ \Phir\), e.g., realized as real-to-sim-to-real};
  \node[font=\scriptsize, text=blue!45!black, anchor=south west] at (-0.98,-5.78) {real-to-sim};
  \node[font=\scriptsize, text=tchL, anchor=south west] at (11.38,-5.78) {sim-to-real};
  \foreach \a/\b in {in/D, D/S} {
    \draw[decorate, decoration={brace, amplitude=5pt, mirror},
          line width=0.6pt, draw=blue!45!black]
          ($(\a.east)+(0.12,-0.26)$) -- ($(\b.west)+(-0.12,-0.26)$)
          coordinate[midway] (brM);
    \draw[->, line width=0.6pt, draw=blue!45!black]
          ($(brM)+(0,-0.24)$) -- ($(brM |- 0,-1.95)+(0,0.04)$);
  }
  \node[box, fill=meas, draw=measL] (p0) at (0.3,\py)   {\((\Imgz, \Cam)\)};
  \node[box, fill=geo1, draw=geoL]  (Op) at (2.75,\py)  {\(\Objp\)};
  \node[box, fill=geo2, draw=geoL]  (S0) at (5.2,\py)   {\(\SurfH_0\)};
  \node[box, fill=geo3, draw=geoL]  (S1) at (7.65,\py)  {\(\SurfH,\ \Obj\)};
  \node[box, fill=dep,  draw=depL]  (Dd) at (10.1,\py)  {\(\Dp\)};
  \node[box, fill=tch,  draw=tchL]  (Sb) at (12.4,\py)  {\(\Sg\)};
  \draw[flow] (p0) -- (Op) node[midway, above=1pt, map] {\(\Phil\)};
  \draw[flow] (Op) -- (S0) node[midway, above=1pt, map] {\(\Phic\)};
  \draw[flow] (S0) -- (S1) node[midway, above=1pt, map] {\(\Phitr\)};
  \draw[flow] (S1) -- (Dd) node[midway, above=1pt, map] {\(\Phid\)};
  \draw[flow] (Dd) -- (Sb) node[midway, above=1pt, map] {\(\Phit\)};
  \coordinate (tr) at ($(S0)!0.5!(S1)$);
  \coordinate (dp) at ($(S1)!0.5!(Dd)$);
  \node[pill] (Y) at ($(tr)+(0,-1.5)$) {\(\Img\)};
  \draw[flow] (Y) -- (tr);
  \node[sub, anchor=north] at ($(Y.south)+(0,-0.02)$)
        {measurement at contact\\(likely occluded)};
  \node[pill] (P) at ($(dp)+(0,-1.5)$) {\(\Pose\)};
  \draw[flow] (P) -- (dp);
  \node[sub, anchor=north] at ($(P.south)+(0,-0.02)$) {touch-sensor pose};
  \node[sub, below=3pt of p0] {reference measurement\\+ depth-sensor pose};
  \node[sub, below=3pt of Op] {partial\\point cloud};
  \node[sub, below=3pt of S0] {reference\\shape};
  \node[sub, below=3pt of S1, xshift=-7pt] {tracked\\shape};
  \node[sub, below=3pt of Dd] {rendered\\depth patch};
  \node[sub, below=3pt of Sb] {real\\touch signal};
\end{tikzpicture}
\caption{The Geometric Force (Definition~\ref{def:geoforce}) as a composition of three maps (top). The
input is the geometric configuration \(\Config\): a reference depth measurement \(\Imgz\), taken before contact, a
depth measurement \(\Img\) at contact, the pose \(\Cam\) of the depth sensor, and
the pose \(\Pose\) of the touch sensor. The rendering map \(\Phir\) turns them
into a depth patch \(\Dp\) seen from the touch sensor; the touch map \(\Phit\)
predicts a touch signal \(\Sg\); the force map \(\Phif\) estimates the force
field \(\F\). The touch map, learned or simulated, carries all object-specific information; the force map depends only on the sensor.
The depth patch, touch signal, and force field share the same pixel grid.
Bottom: one realization of \(\Phit \circ \Phir\), in a real-to-sim-to-real
fashion. Real-to-sim: \(\Imgz\) is lifted to a partial point cloud
\(\Objp\), the object's shape is completed once into the reference shape
\(\SurfH_0\), the object is tracked to the instant of contact from the likely
occluded measurement \(\Img\), giving the tracked shape \(\SurfH\), which is
sampled densely into the point cloud \(\Obj\) and rendered from the touch-sensor pose
\(\Pose\). Sim-to-real: the touch map turns the rendered depth patch into the
touch signal of the real sensor. Alternatively, \(\Phir\) may be a single
network learned end to end; \(\Phit\) may be a tactile simulator with a learned
correction.}
\label{fig:pipeline}
\end{figure*}
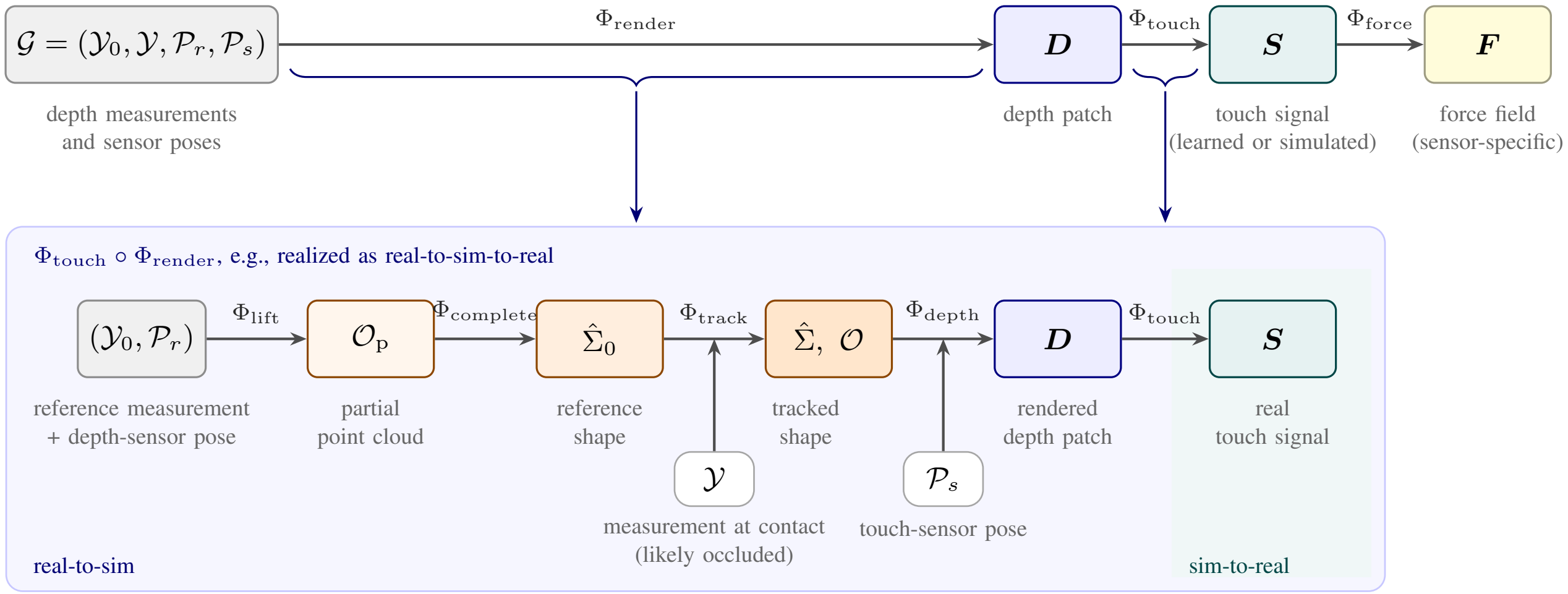
\begin{figure}[t]
\centering
\begin{tikzpicture}[scale=1.1, >=Stealth,
    vec/.style={->, thick},
    lbl/.style={font=\small}]
  \coordinate (W)  at (0,0);
  \coordinate (T)  at (2.9,0.5);
  \coordinate (eu) at (0.9,-0.18);
  \coordinate (ev) at (0.45,0.42);
  \coordinate (nn) at (-0.05,0.95);
  \fill[blue!6] ($(T)-1.6*(eu)-1.0*(ev)$) -- ($(T)+1.3*(eu)-1.0*(ev)$)
        -- ($(T)+1.3*(eu)+1.2*(ev)$) -- ($(T)-1.6*(eu)+1.2*(ev)$) -- cycle;
  \draw[blue!50] ($(T)-1.6*(eu)-1.0*(ev)$) -- ($(T)+1.3*(eu)-1.0*(ev)$)
        -- ($(T)+1.3*(eu)+1.2*(ev)$) -- ($(T)-1.6*(eu)+1.2*(ev)$) -- cycle;
  \node[font=\scriptsize, align=center, blue!60!black, anchor=north]
        at ($(T)+1.15*(eu)-1.0*(ev)$)
        {viewing plane \(z=0\)\\(inside the sensor, at its camera)};
  \fill[teal!35, fill opacity=0.35]
        ($(T)-1.6*(eu)-1.0*(ev)+0.35*(nn)$) -- ($(T)+1.3*(eu)-1.0*(ev)+0.35*(nn)$)
        -- ($(T)+1.3*(eu)+1.2*(ev)+0.35*(nn)$) -- ($(T)-1.6*(eu)+1.2*(ev)+0.35*(nn)$)
        -- cycle;
  \node[font=\scriptsize, align=left, teal!50!black, anchor=west, inner sep=1pt]
        (gellbl) at (4.75,1.05) {gel surface\\\(z=z_0\)};
  \draw[black!45, thin] (gellbl.west) -- ($(T)+1.2*(eu)+0.9*(ev)+0.35*(nn)$);
  \draw[thick, draw=orange!55!black, fill=orange!12] (3.8,3.05) ellipse (1.05 and 0.62);
  \node[lbl] at (3.8,3.15) {object};
  \node[font=\scriptsize] at (5.15,3.5) {surface \(\Surf\)};
  \coordinate (X) at (3.85,2.43);
  \fill (X) circle (1.3pt);
  \node[lbl, anchor=south west] at (X) {\(\mathbf{x}\)};
  \draw[->, gray] (W) -- (0.6,0) node[right, lbl] {};
  \draw[->, gray] (W) -- (0.3,0.25);
  \draw[->, gray] (W) -- (0,0.6);
  \node[font=\scriptsize, gray, align=right, anchor=east, inner sep=1pt] at (-0.12,0.28) {coordinate\\frame};
  \draw[vec, black!60] (W) -- (T) node[pos=0.3, above, lbl] {\(\mathbf{t}\)};
  \draw[vec, black!60] (W) -- (X);
  \draw[vec, red!70!black, dashed] (T) -- (X)
        node[pos=0.55, right, lbl] {\(\mathbf{x}-\mathbf{t}\)};
  \draw[vec, blue!70!black] (T) -- ($(T)+(eu)$) node[below, lbl] {\(\mathbf{e}_u\)};
  \draw[vec, blue!70!black] (T) -- ($(T)+(ev)$) node[above right, lbl, xshift=-2pt] {\(\mathbf{e}_v\)};
  \draw[vec, blue!70!black] (T) -- ($(T)+(nn)$) node[left, lbl] {\(\mathbf{n}\)};
  \fill (T) circle (1.3pt);
  \coordinate (P) at ($(T)-1.2*(eu)-0.8*(ev)$);
  \draw[densely dotted, thick] (T) -- ($(T)-1.2*(eu)$) -- (P);
  \fill[blue!70!black] (P) circle (1.3pt);
  \node[font=\scriptsize, align=center, inner sep=1pt] (Plbl) at (0.8,-0.62)
        {pixel \(\mathbf{x}(a,b)\)\\on the viewing plane};
  \draw[gray] (Plbl.north) -- (P);
\end{tikzpicture}
\caption{Sensor coordinate frame (Definition~\ref{def:config}). The pose \(\Pose = (\mathbf{R}, \mathbf{t})\) places the
sensor's viewing plane in the coordinate frame: \(\mathbf{t}\) is the position of its
origin and the axes \(\mathbf{e}_u, \mathbf{e}_v, \mathbf{n}\), the columns of
\(\mathbf{R}\), are vectors expressed in the coordinate frame. The object surface
\(\Surf\) is a set of points. A surface point \(\mathbf{x}\) is seen from the
sensor through the vector \(\mathbf{x} - \mathbf{t}\), whose components along
\(\mathbf{e}_u, \mathbf{e}_v, \mathbf{n}\) are its sensor coordinates
\((u, v, z)\), Eq.~\eqref{eq:sensor_coords}. Pixel \((a,b)\) lies in the
viewing plane at the point
\(\mathbf{x}(a,b) = \mathbf{t} + u_a\mathbf{e}_u + v_b\mathbf{e}_v\); the
dotted lines show its sensor coordinates \(u_a\) and \(v_b\) along
\(\mathbf{e}_u\) and \(\mathbf{e}_v\). The pixel rays start on the viewing plane and pass through the gel to the
object (Fig.~\ref{fig:depth}).}
\label{fig:frame}
\end{figure}
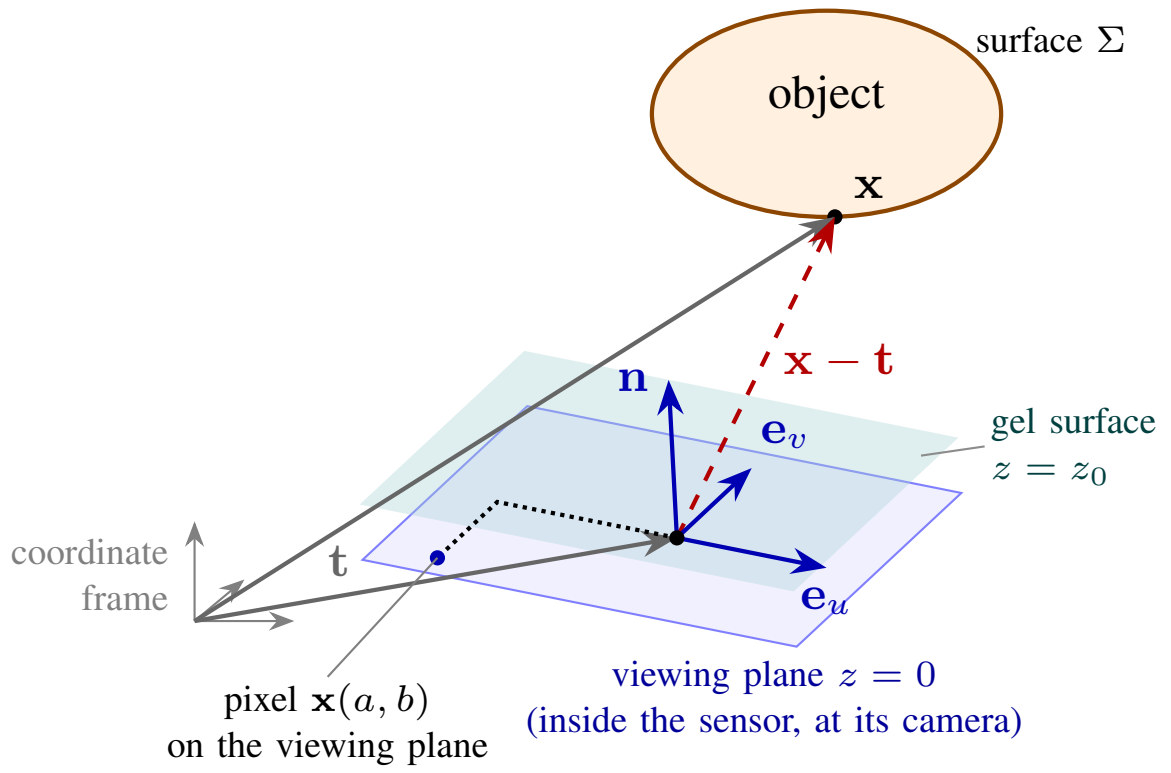

\subsection{Geometric Configuration}

We consider a rigid object with surface \(\Surf \subset \R^{3}\), observed by
a \emph{depth sensor}, i.e., any device that measures distances to surfaces,
such as an \rgbd{} camera or a stereo camera, with pose
\(\Cam = (\mathbf{R}_{r}, \mathbf{t}_{r}) \in \mathrm{SE}(3)\), and a virtual
touch sensor with pose \(\Pose = (\mathbf{R}, \mathbf{t}) \in \mathrm{SE}(3)\).
All quantities are expressed in one coordinate frame. In this frame,
\(\mathbf{t} \in \R^{3}\) is the origin of the touch sensor's viewing plane and
\(\mathbf{R} = [\mathbf{e}_u \; \mathbf{e}_v \; \mathbf{n}]\) its orientation,
where the unit vectors \(\mathbf{e}_u, \mathbf{e}_v\) span the viewing plane
and \(\mathbf{n}\) is its unit normal, pointing towards the scene. The
\emph{sensor coordinates} of a point \(\mathbf{x}\) are its coordinates
in the 3D coordinate frame attached to the viewing plane, with origin \(\mathbf{t}\) and
axes \(\mathbf{e}_u, \mathbf{e}_v, \mathbf{n}\),
\begin{equation}
  \pi_{\Pose}(\mathbf{x}) = \mathbf{R}^{\top}(\mathbf{x} - \mathbf{t})
  = (u, v, z)^{\top},
  \label{eq:sensor_coords}
\end{equation}
see Fig.~\ref{fig:frame}: \((u, v)\) is the position along the viewing plane
and \(z\) the depth in front of it; dropping the depth,
\(\bar{\pi}_{\Pose}(\mathbf{x}) = (u, v)^{\top}\) is the position of \(\mathbf{x}\)
on the viewing plane. Applied to a set, \(\pi_{\Pose}\) acts on
each of its points; e.g., \(\pi_{\Pose}(\Surf)\) is the surface \(\Surf\)
expressed in sensor coordinates. The touch sensor observes a pixel grid
\(\Grid = \{1, \dots, h\} \times \{1, \dots, w\}\) with pixel pitch \(\lambda\).
Pixel \((a,b)\) has center \(\mathbf{p}_{ab} = (u_a, v_b)\) in the viewing
plane, i.e., it is located at the point
\(\mathbf{x}(a,b) = \mathbf{t} + u_a \mathbf{e}_u + v_b \mathbf{e}_v\).
The undeformed sensing surface (for an optical sensor, the gel) is the plane
\(z = z_0 > 0\). Depths beyond \(z_{\max} > z_0\) lie outside the sensing range.

\begin{definition}[Geometric configuration \(\Config\)]
\label{def:config}
A \emph{geometric configuration} is a tuple
\(\Config = (\Imgz, \Img, \Cam, \Pose)\) of a reference depth measurement \(\Imgz\), taken earlier while the object is
unoccluded; a depth measurement \(\Img\) at the \emph{query instant}, at which
the force is evaluated, typically at contact; the pose of the depth sensor;
and the pose of the touch sensor at the query instant. It is \emph{frozen}:
the object geometry is taken as undeformed, without the contact deformation of object and sensor.
See Fig.~\ref{fig:pipeline}.
\end{definition}

For simplicity, we take the depth sensor to be static; a moving one only
requires its pose at each measurement. The virtual touch sensor replaces only
the readout of a real sensor: its sensing surface, e.g., the gel, stays in
place, so that contact with the object is the same as with the real sensor.

\subsection{Depth Patch and Rendering}
\label{sec:depth-rendering}

\begin{figure}[t]
\centering
\begin{tikzpicture}[scale=0.95, >=Stealth,
    math/.style={font=\small},
    word/.style={font=\scriptsize, text=black!80},
    level/.style={font=\scriptsize, align=left, anchor=west, text=black!80},
    dim/.style={{Stealth[length=3pt]}-{Stealth[length=3pt]}, line width=0.5pt,
                draw=black!80},
    /utils/exec={%
      \colorlet{objfill}{orange!12}\colorlet{objline}{orange!55!black}%
      \colorlet{ovfill}{orange!45}%
      \colorlet{gelfill}{teal!10}\colorlet{gelline}{teal!60!black}%
      \colorlet{body}{black!15}\colorlet{wall}{black!30}%
      \colorlet{plane}{blue!55!black}}]
  \begin{scope}
    \clip (0,0) rectangle (5,3.9);
    \fill[objfill] (2.5,3.7) circle (2.5);
  \end{scope}
  \fill[body, rounded corners=3pt] (-0.15,-0.55) rectangle (5.15,0);
  \fill[wall] (-0.15,0) rectangle (0,1.6);
  \fill[wall] (5,0) rectangle (5.15,1.6);
  \fill[gelfill] (0,1.15) rectangle (5,1.6);
  \fill[black!50] (2.25,-0.02) -- (2.75,-0.02) -- (2.62,-0.2) -- (2.38,-0.2) -- cycle;
  \node[word] at (2.5,-0.38) {camera};
  \node[word] at (0.85,-0.28) {sensor body};
  \begin{scope}
    \clip (0,0) rectangle (5,1.6);
    \fill[ovfill] (2.5,3.7) circle (2.5);
  \end{scope}
  \begin{scope}
    \clip (0,0) rectangle (5,3.9);
    \draw[objline, thick] (2.5,3.7) circle (2.5);
  \end{scope}
  \draw[plane, thick] (0,0) -- (5,0);
  \draw[gelline, thick, dashed] (0,1.6) -- (5,1.6);
  \draw[black!55, densely dotted] (0,3.9) -- (5,3.9);
  \foreach \y in {0,1.6,3.9} { \draw[black!35, densely dotted] (5.15,\y) -- (5.4,\y); }
  \node[level] at (5.4,3.9) {\(z = z_{\max}\)\\render limit};
  \node[level] at (5.4,1.6) {\(z = z_0\)\\gel surface};
  \node[level] at (5.4,0)   {\(z = 0\)\\viewing plane};
  \draw[dim] (2.5,0) -- (2.5,1.2);
  \draw[dim] (2.5,1.2) -- (2.5,1.6);
  \node[math, anchor=east] at (2.42,0.6) {\(\Dp(a,b)\)};
  \node[font=\footnotesize, anchor=east, inner sep=1pt] at (2.46,1.41) {\(\pen(a,b)\)};
  \node[font=\small] at (2.5,3.05) {object};
  \node[word] at (4.3,2.95) {surface \(\Surf\)};
  \node[word, text=orange!35!black] at (3.2,1.47) {overlap};
  \draw[->, plane] (-0.55,0) -- (-0.55,0.65) node[above, math] {\(\mathbf{n}\)};
  \draw[->, plane] (-0.55,0) -- (-0.2,0);
  \node[math, text=plane, anchor=north] at (-0.42,-0.04) {\(\mathbf{e}_u\)};
\end{tikzpicture}
\caption{Side view of the touch sensor and the object (Definitions~\ref{def:depth} and~\ref{def:penetration}). Its viewing plane
\(z = 0\) lies inside the sensor, at the camera; the undeformed gel surface
lies in front of it at \(z = z_0\). From a pixel on the viewing
plane, the depth \(\Dp(a,b)\) is the distance along \(\mathbf{n}\) to the
first point of the object surface \(\Surf\). Because the geometry is frozen, the object can
overlap the gel (dark orange region); the penetration \(\pen(a,b) = z_0 - \Dp(a,b)\)
measures this overlap, which a real contact resolves by deforming gel and
object. Pixels whose rays reach the object only beyond \(z_0\) are not in
contact. The far limit \(z_{\max}\) is a rendering range, not a sensor
property.}
\label{fig:depth}
\end{figure}
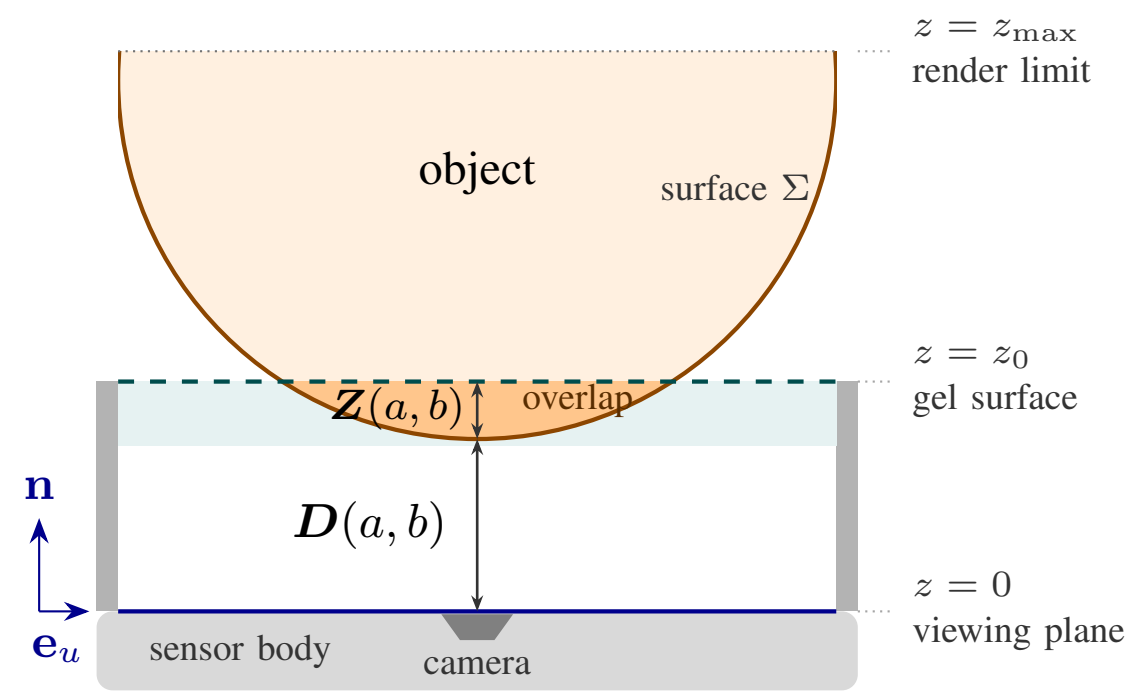

A touch sensor responds to the object's geometry as seen from its viewing
plane, which we describe by a depth patch.

\begin{definition}[Depth patch \(\Dp_{\Surf,\Pose}\) of a surface]
\label{def:depth}
The depth patch \(\Dp_{\Surf,\Pose} : \Grid \to [0, z_{\max}]\) of a surface
\(\Surf\) at pose \(\Pose\) is a scalar field defined by casting, from the center of each
pixel, a ray along \(\mathbf{n}\): \(\Dp_{\Surf,\Pose}(a,b)\) is the distance
the ray travels before it first hits \(\Surf\), or \(z_{\max}\) if it hits
nothing up to that depth:
\begin{equation}
  \Dp_{\Surf,\Pose}(a,b) =
  \begin{cases}
    \min \mathcal{Z}_{ab} & \text{if } \mathcal{Z}_{ab} \neq \emptyset, \\
    z_{\max} & \text{otherwise},
  \end{cases}
\end{equation}
where \(\mathcal{Z}_{ab} = \{ z \in [0, z_{\max}] : (\mathbf{p}_{ab}, z) \in
\pi_{\Pose}(\Surf) \}\) collects the depths at which this ray hits \(\Surf\).
See Fig.~\ref{fig:depth}.
\end{definition}

The \emph{true depth patch} \(\Dp^{*} = \Dp_{\Surf,\Pose}\) at the query
instant cannot be observed directly: at contact, the touched region is likely
hidden from the depth sensor. It has to be estimated from the
measurements.

\begin{definition}[Rendering map \(\Phir\) and depth-patch error \(\delta\)]
A \emph{rendering map} \(\Phir\) assigns to a geometric configuration an
estimate \(\hat{\Dp} = \Phir(\Config)\) of the true depth
patch. Its \emph{depth-patch error} \(\delta\) is the largest depth error
over all pixels,
\begin{equation}
  \delta = \max_{(a,b)} \bigl\lvert\hat{\Dp}(a,b) - \Dp^{*}(a,b)\bigr\rvert.
\end{equation}
\end{definition}

Since \(\delta\) is measured against the true depth patch, it includes the
effect of errors in the inputs, such as an inaccurate touch-sensor pose. A
rendering map can be realized
in different ways, e.g., by reconstructing the object as explicit geometry and
rendering it in simulation, as worked out in Section~\ref{sec:r2s2r}, or by a
single network learned end to end. Our main results depend on \(\Phir\) only
through \(\delta\).

\begin{definition}[Penetration field \(\pen\)]
\label{def:penetration}
The \emph{penetration field} \(\pen\) of a depth patch \(\Dp\) is the scalar field
\begin{equation}
  \pen(a,b) = \max\bigl(0,\, z_0 - \Dp(a,b)\bigr).
\end{equation}
The configuration is
\emph{in contact} if \(\pen \neq 0\). A surface that only touches the sensing
surface, \(\Dp(a,b) = z_0\), has zero penetration.
See Fig.~\ref{fig:depth}.
\end{definition}

Since the geometry is frozen, \(\pen\) measures the overlap between the
undeformed object and the undeformed sensing surface. A real contact resolves
this overlap by deforming the sensor, the object, or both
(Fig.~\ref{fig:depth}).

\subsection{Touch and Force}

A real touch sensor produces a raw output \(\Sg_{\mathrm{raw}}\) in its raw
output space \(\mathcal{S}_{\mathrm{raw}}\), which need not live on \(\Grid\). A
gridding map \(\Gamma : \mathcal{S}_{\mathrm{raw}} \to (\Grid \to \R^{c})\)
casts it onto the grid, with \(c\) channels per pixel, e.g., \(c = 3\) for an
RGB tactile image. We define the \emph{touch signal}, a vector field with \(c\) channels,
as the background-subtracted output
\(\Sg = \Gamma(\Sg_{\mathrm{raw}}) - \Gamma(\Sg_{\mathrm{raw}}^{b})\), where
\(\Sg_{\mathrm{raw}}^{b} \in \mathcal{S}_{\mathrm{raw}}\) is the raw output of the
free sensor, i.e., not in interaction with any object. Thus \(\Sg = 0\) means that the sensor shows no
disturbance.

The \emph{touch map} \(\Phit : \R^{h \times w} \to \R^{h \times w \times c}\)
predicts the touch signal from the depth patch. It is the central component
of the Geometric Force, since it grounds the purely geometric input in the
physical response of the sensor. It can be learned from paired real data
\((\Dp, \Sg)\), or realized by a tactile
simulator~\cite{wang2022tacto,si2022taxim} that computes the sensor response
from the penetrating geometry. A simulator can be followed by a learned
\emph{correction}: a network trained on paired real data to map the
simulated signal to the real one. When learned, it depends on \(\Gamma\) only through its
training targets. The \emph{force map}
\(\Phif : \R^{h \times w \times c} \to \R^{h \times w \times 3}\) estimates a force field \(\f\) in sensor coordinates, where \(\f(a,b)\) is the
force acting on pixel \((a,b)\), expressed in the basis
\((\mathbf{e}_u, \mathbf{e}_v, \mathbf{n})\).

\begin{definition}[Geometric Force \(\F\)]
\label{def:geoforce}
The \emph{Geometric Force} of a geometric configuration \(\Config\) is the
vector field \(\F : \Grid \to \R^{3}\),
located at the points \(\mathbf{x}(a,b)\),
\begin{equation}
  \begin{gathered}
    \f = \Phif \circ \Phit \circ \Phir(\Config), \\
    \F(a,b) = \mathbf{R}\, \f(a,b).
  \end{gathered}
  \label{eq:geoforce}
\end{equation}
We write \(\F = \Phi(\Config)\); see Fig.~\ref{fig:pipeline}.
\end{definition}

The rotation \(\mathbf{R}\) expresses the force, given in sensor
coordinates, in the coordinate frame. With \((f_u, f_v, f_n)\) the entries of \(\f(a,b)\), the
force at pixel \((a,b)\) is
\(\F(a,b) = f_u\mathbf{e}_u + f_v\mathbf{e}_v + f_n\mathbf{n}\). Its \emph{normal component} is
\(\fld{F}^{\perp}(a,b) = \F(a,b) \cdot \mathbf{n} = f_n\). The translation
\(\mathbf{t}\) does not enter, since a force has a direction and a magnitude
but no position. Because each force acts on the gel at its own pixel, their
sum \(\mathbf{f}_{\mathrm{tot}} = \sum_{(a,b) \in \Grid} \F(a,b) \in \R^{3}\)
is the \emph{total force}.

By construction, \(\Dp\), \(\Sg\) and \(\F\) share the same pixel grid; as
arrays, \(\Dp \in \R^{h \times w}\), \(\Sg \in \R^{h \times w \times c}\) and
\(\F \in \R^{h \times w \times 3}\). We are mainly concerned with
\(\fld{F}^{\perp}\). Force decoders that predict only the total force are
covered by our results by summing over \(\Grid\), at the cost of a factor
\(hw\) in the bounds of Section~\ref{sec:stability}.

\section{STRUCTURAL PROPERTIES}
\label{sec:properties}

We show that the Geometric Force is equivariant to the coordinate frame whenever rendering is frame-invariant, and that it vanishes out of contact.

\subsection{Invariance to the Coordinate Frame}

To describe a change of the coordinate frame, we use a rigid motion
\(g = (\mathbf{R}_g, \mathbf{t}_g) \in \mathrm{SE}(3)\). It moves a point
\(\mathbf{x}\) to \(g\mathbf{x} = \mathbf{R}_g \mathbf{x} + \mathbf{t}_g\).
It moves a point set or a surface by moving each of its points. It moves a
pose by rotating its axes and moving its origin, so the touch-sensor pose
becomes \(g\Pose = (\mathbf{R}_g \mathbf{R}, \mathbf{R}_g \mathbf{t} + \mathbf{t}_g)\);
the depth-sensor pose \(\Cam\) moves in the same way. It moves a geometric
configuration by moving its two poses,
\(g\Config = (\Imgz, \Img, g\Cam, g\Pose)\): the measurements are expressed in
the coordinate frame of the depth sensor and do not change.

\begin{definition}[Invariance to the coordinate frame]
A rendering map is \emph{invariant to the coordinate frame} if
\begin{equation}
  \Phir(g\Config) = \Phir(\Config) \quad \text{for all } g.
\end{equation}
In words, moving the coordinate frame does not change the rendered depth patch.
\end{definition}

\begin{proposition}[Equivariance to the coordinate frame]
\label{prop:equivariance}
If \(\Phir\) is frame-invariant, then, for every pixel,
\begin{equation}
  \Phi(g\Config) = \mathbf{R}_g\, \Phi(\Config).
\end{equation}
In particular, the
normal component \(\fld{F}^{\perp}\) does not change.
See Fig.~\ref{fig:overview}c.
\end{proposition}

Consequently, the Geometric Force depends only on the relative poses of depth
sensor, object, and touch sensor: an error shared by all poses, such as a
misplaced coordinate frame, does not affect it.

\subsection{Vanishing out of Contact}

A virtual touch sensor should report no force out of contact, which requires two assumptions on the touch map and the force map.

\begin{assumption}[Contact-consistent touch map]
\label{ass:nocontact}
If the penetration field of \(\Dp\) is \(\pen = 0\), then
\(\Phit(\Dp) = 0\).
\end{assumption}

\begin{assumption}[Calibrated force map]
\label{ass:forcecal}
\(\Phif(0) = 0\).
\end{assumption}

\begin{proposition}[Zero force out of contact]
\label{prop:vanishing}
Under Assumptions~\ref{ass:nocontact} and~\ref{ass:forcecal}, if the rendered
depth patch shows no contact, then \(\F = 0\).
See Fig.~\ref{fig:overview}c.
\end{proposition}

\begin{proof}
If \(\pen = 0\), Assumption~\ref{ass:nocontact} gives
\(\Sg = \Phit(\Dp) = 0\). Assumption~\ref{ass:forcecal} then gives
\(\f = \Phif(0) = 0\), hence \(\F = 0\).
\end{proof}

In practice, both maps meet these assumptions only approximately, so the
virtual touch sensor may report a force where there is no contact, a
\emph{phantom force}. The force map is easy to correct: subtracting its output for the free
sensor, \(\Phif(0)\), from every estimate makes Assumption~\ref{ass:forcecal}
hold exactly. The touch map is harder, since it must output zero not for one
input but for every depth patch without contact. The following construction
achieves this.

\begin{proposition}[Contact consistency of gated touch maps]
\label{prop:gated}
Let \(\Phit(\Dp) = \psi(\pen)\), where \(\pen\) is the penetration field of
\(\Dp\) and \(\psi : \R_{\ge 0}^{h \times w} \to \R^{h \times w \times c}\),
defined on fields with non-negative entries, satisfies \(\psi(0) = 0\). Then
Assumption~\ref{ass:nocontact} holds.
\end{proposition}

Examples of such \(\psi\) are networks without additive biases or
normalization offsets, whose piecewise-linear activations make them positively
homogeneous, and arbitrary networks whose output is multiplied pixel-wise by
the contact mask \(\mathbf{1}[\pen > 0]\). Gating also makes the prediction depend only on what presses into the
sensor: a depth at or beyond the gel, \(\Dp(a,b) \ge z_0\), gives
\(\pen(a,b) = 0\), so geometry that does not penetrate the gel cannot affect
the predicted touch signal.

\section{STABILITY BOUNDS}
\label{sec:stability}

Each of the three maps of the Geometric Force is imperfect: the rendered depth patch
deviates from the true one, and the touch and force maps from ideal ones. We ask how
these errors reach the force. As a reference, we take the ideal Geometric
Force \(\fstar = \Phif^{*} \circ \Phit^{*}(\Dp^{*})\), computed from the true
depth patch with reference maps \(\Phit^{*}\) and \(\Phif^{*}\);
Section~\ref{sec:limits} identifies \(\Phit^{*}\) as the best possible touch
map given geometry alone. Errors are measured at the worst pixel, with the
norm \(\ninf{\cdot}\) of Section~\ref{sec:conventions}.

\subsection{Force Error}

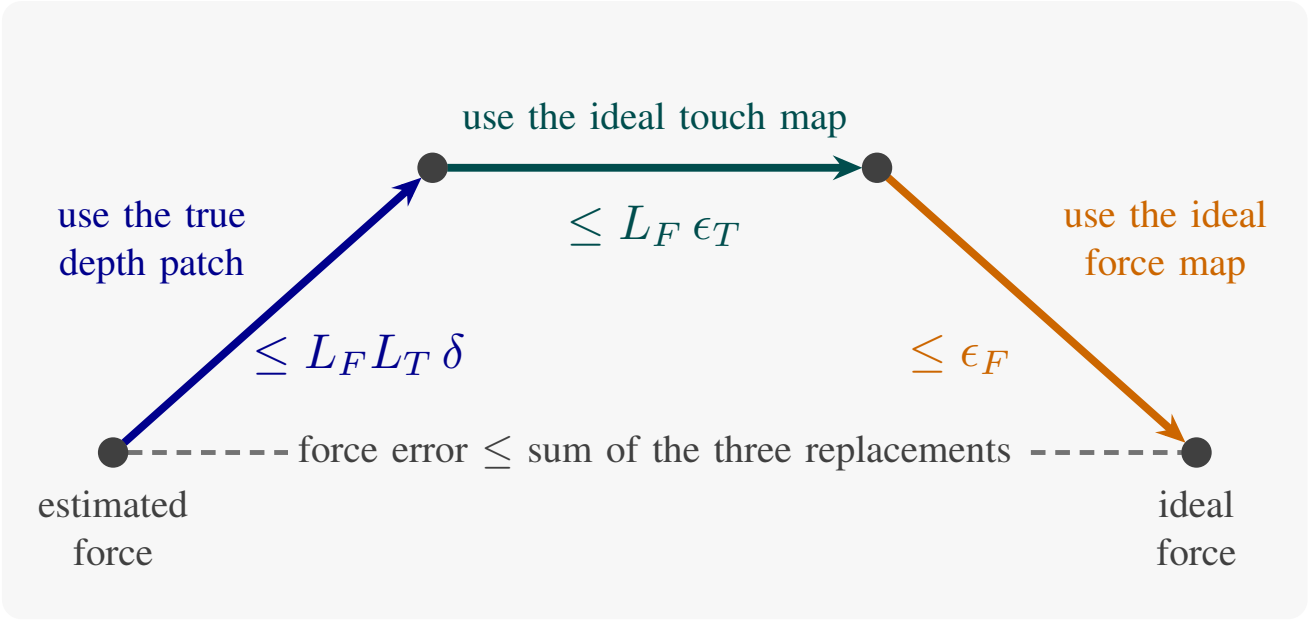
\begin{figure}[t]
\centering
\resizebox{\columnwidth}{!}{%
\begin{tikzpicture}[>={Stealth[length=6pt, width=5pt]},
    /utils/exec={%
      \colorlet{depL}{blue!55!black}\colorlet{tchL}{teal!60!black}%
      \colorlet{frcL}{orange!80!black}\colorlet{panelfill}{black!3}},
    pt/.style={circle, fill=black!75, inner sep=2.2pt},
    ptname/.style={font=\footnotesize, text=black!75, align=center},
    steplbl/.style={font=\footnotesize, align=center},
    bound/.style={font=\normalsize}]
  \fill[panelfill, rounded corners=4pt] (-0.4,-0.9) rectangle (9.0,3.55);
  \coordinate (A) at (0.4,0.3);
  \coordinate (B) at (2.7,2.35);
  \coordinate (C) at (5.9,2.35);
  \coordinate (D) at (8.2,0.3);
  \draw[->, depL, line width=1.6pt, shorten >=3pt, shorten <=3pt] (A) -- (B)
    node[midway, above left=2pt, steplbl, text=depL] {use the true\\depth patch};
  \draw[->, tchL, line width=1.6pt, shorten >=3pt, shorten <=3pt] (B) -- (C)
    node[midway, above=2pt, steplbl, text=tchL] {use the ideal touch map};
  \draw[->, frcL, line width=1.6pt, shorten >=3pt, shorten <=3pt] (C) -- (D)
    node[midway, above right=2pt, steplbl, text=frcL] {use the ideal\\force map};
  \node[bound, text=depL] at ($(A)!0.5!(B)+(0.62,-0.32)$) {\(\le L_F L_T\,\delta\)};
  \node[bound, text=tchL] at ($(B)!0.5!(C)+(0,-0.42)$) {\(\le L_F\,\epsilon_T\)};
  \node[bound, text=frcL] at ($(C)!0.5!(D)+(-0.55,-0.32)$) {\(\le \epsilon_F\)};
  \draw[black!55, densely dashed, line width=1pt, shorten >=3pt, shorten <=3pt] (A) -- (D);
  \node[ptname, fill=panelfill, inner sep=2pt] at ($(A)!0.5!(D)$) {force error \(\le\) sum of the three replacements};
  \node[pt] at (A) {};
  \node[pt] at (B) {};
  \node[pt] at (C) {};
  \node[pt] at (D) {};
  \node[ptname, anchor=north] at ($(A)+(0,-0.15)$) {estimated\\force};
  \node[ptname, anchor=north] at ($(D)+(0,-0.15)$) {ideal\\force};
\end{tikzpicture}}
\caption{Where the force stability bound of Theorem~\ref{thm:stability} comes from. The estimated force uses the
rendered depth patch and the touch and force maps at hand; the ideal force
uses the true depth patch and the ideal maps. Replacing these three parts by
their ideal counterparts one at a time passes through two intermediate forces. Each replacement moves the force by at most the error of that
part, amplified by the maps after it. By the triangle inequality, the direct
gap (dashed) is at most their sum.}
\label{fig:triangle}
\end{figure}

\begin{theorem}[Stability]
\label{thm:stability}
Let \(\fhat = \Phif \circ \Phit(\hat{\Dp})\), where \(\hat{\Dp}\) is rendered by
any rendering map with depth-patch error \(\delta\). Let \(\Phit\) be
\(L_T\)-Lipschitz and \(\Phif\) be \(L_F\)-Lipschitz with respect to
\(\ninf{\cdot}\), and let their approximation errors over the relevant
inputs be
\begin{equation}
  \begin{aligned}
    \epsilon_T &= \sup_{\Dp}\ninf{\Phit(\Dp) - \Phit^{*}(\Dp)}, \\
    \epsilon_F &= \sup_{\Sg}\ninf{\Phif(\Sg) - \Phif^{*}(\Sg)}.
  \end{aligned}
\end{equation}
Then
\begin{equation}
  \ninf{\fhat - \fstar} \le L_F L_T\,\delta + L_F\,\epsilon_T + \epsilon_F,
  \label{eq:stability}
\end{equation}
and the same bound holds for the normal component,
\(\max_{(a,b)}\lvert\hat{\fld{F}}^{\perp}(a,b) - \fld{F}^{*\perp}(a,b)\rvert\).
See Fig.~\ref{fig:triangle}.
\end{theorem}

For a contact-gated touch map (Proposition~\ref{prop:gated}), \(L_T\) is the
Lipschitz constant of \(\psi\), since computing the penetration from the
depth patch does not amplify depth errors.

\subsection{Depth-Patch Error}
\label{sec:depth-error}

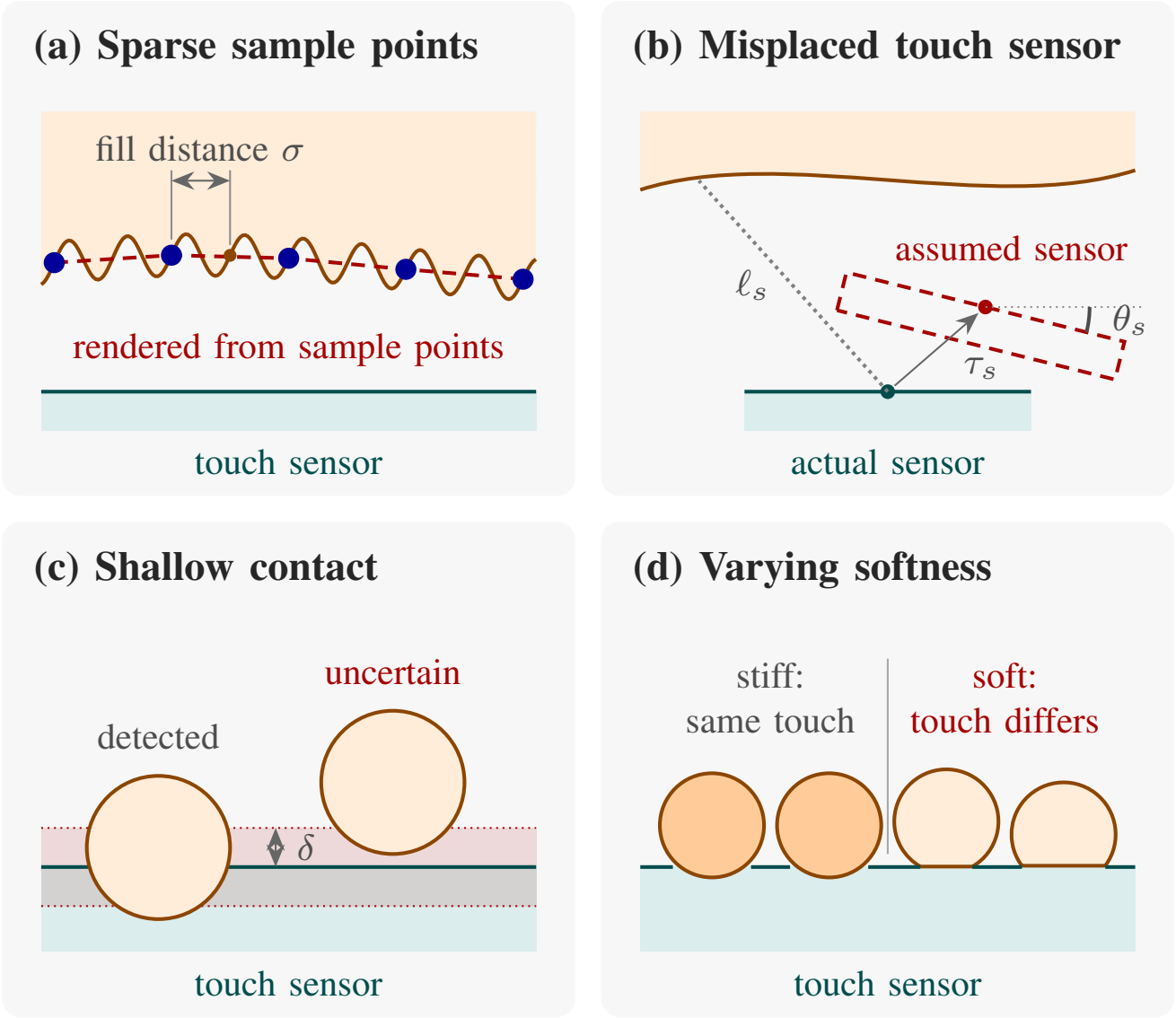
\begin{figure}[t]
\centering
\resizebox{\columnwidth}{!}{%
\begin{tikzpicture}[>={Stealth[length=5pt, width=4pt]},
    /utils/exec={%
      \colorlet{objfill}{orange!15}\colorlet{objline}{orange!55!black}%
      \colorlet{rigidfill}{orange!40}%
      \colorlet{gelfill}{teal!14}\colorlet{gelline}{teal!60!black}%
      \colorlet{samp}{blue!60!black}%
      \colorlet{err}{red!65!black}\colorlet{panelfill}{black!3}},
    head/.style={font=\small\bfseries, text=black!85, anchor=north west},
    lbl/.style={font=\footnotesize, text=black!70},
    gel/.style={fill=gelfill},
    gelsurf/.style={gelline, thick}]
  \foreach \px/\py in {0/4.0, 4.6/4.0, 0/0, 4.6/0} {
    \fill[panelfill, rounded corners=4pt] (\px,\py) rectangle ({\px+4.4},{\py+3.8});
  }
  \begin{scope}[shift={(0,4.0)}]
    \node[head] at (0.12,3.7) {(a) Sparse sample points};
    \fill[gel] (0.3,0.5) rectangle (4.1,0.8);
    \draw[gelsurf] (0.3,0.8) -- (4.1,0.8);
    \node[lbl, text=gelline, anchor=north] at (2.2,0.48) {touch sensor};
    \fill[objfill] plot[domain=0.3:4.1, samples=200]
      (\x, {1.75+0.1*sin(60*\x)+0.16*sin(800*(\x-0.4))}) -- (4.1,2.95) -- (0.3,2.95) -- cycle;
    \draw[objline, thick] plot[domain=0.3:4.1, samples=200]
      (\x, {1.75+0.1*sin(60*\x)+0.16*sin(800*(\x-0.4))});
    \draw[err, thick, densely dashed]
      plot[samples at={0.4,1.3,2.2,3.1,4.0}] (\x, {1.75+0.1*sin(60*\x)});
    \foreach \x in {0.4,1.3,2.2,3.1,4.0} {
      \fill[samp] (\x, {1.75+0.1*sin(60*\x)}) circle (2.3pt);
    }
    \draw[black!45, thin] (1.3,{1.75+0.1*sin(78)+0.12}) -- (1.3,2.5);
    \draw[black!45, thin] (1.75,{1.75+0.1*sin(105)+0.08}) -- (1.75,2.5);
    \fill[objline] (1.75,{1.75+0.1*sin(105)}) circle (1.4pt);
    \draw[<->, black!60] (1.3,2.42) -- (1.75,2.42);
    \node[lbl, anchor=south] at (1.52,2.45) {fill distance \(\sigma\)};
    \node[lbl, text=err] at (2.2,1.12) {rendered from sample points};
  \end{scope}
  \begin{scope}[shift={(4.6,4.0)}]
    \node[head] at (0.12,3.7) {(b) Misplaced touch sensor};
    \fill[objfill] (0.3,2.35) .. controls (1.4,2.7) and (3.0,2.15) .. (4.1,2.5)
      -- (4.1,2.95) -- (0.3,2.95) -- cycle;
    \draw[objline, thick] (0.3,2.35) .. controls (1.4,2.7) and (3.0,2.15) .. (4.1,2.5);
    \fill[gel] (1.1,0.5) rectangle (3.3,0.8);
    \draw[gelsurf] (1.1,0.8) -- (3.3,0.8);
    \fill[gelline] (2.2,0.8) circle (1.6pt);
    \node[lbl, text=gelline, anchor=north] at (2.2,0.48) {actual sensor};
    \draw[black!50, densely dotted, thick] (2.2,0.8) -- (0.75,2.42);
    \node[lbl, anchor=east] at (1.42,1.62) {\(\ell_s\)};
    \begin{scope}[shift={(2.95,1.45)}, rotate=-14]
      \draw[err, thick, densely dashed] (-1.1,-0.3) rectangle (1.1,0);
      \fill[err] (0,0) circle (1.6pt);
    \end{scope}
    \draw[black!45, densely dotted] (2.95,1.45) -- (4.15,1.45);
    \draw[black!70, thick] (2.95,1.45) ++(0:0.8) arc (0:-14:0.8);
    \node[lbl, anchor=west] at (3.8,1.35) {\(\theta_s\)};
    \draw[->, black!60, shorten >=2pt, shorten <=2pt] (2.2,0.8) -- (2.95,1.45);
    \node[lbl, anchor=west] at (2.66,0.98) {\(\tau_s\)};
    \node[lbl, text=err, anchor=south] at (3.15,1.68) {assumed sensor};
  \end{scope}
  \begin{scope}[shift={(0,0)}]
    \node[head] at (0.12,3.7) {(c) Shallow contact};
    \fill[gel] (0.3,0.5) rectangle (4.1,1.15);
    \fill[err, opacity=0.12] (0.3,0.85) rectangle (4.1,1.45);
    \draw[err, densely dotted] (0.3,0.85) -- (4.1,0.85);
    \draw[err, densely dotted] (0.3,1.45) -- (4.1,1.45);
    \draw[gelsurf] (0.3,1.15) -- (4.1,1.15);
    \node[lbl, text=gelline, anchor=north] at (2.2,0.48) {touch sensor};
    \fill[objfill] (1.2,1.3) circle (0.55);
    \draw[objline, thick] (1.2,1.3) circle (0.55);
    \fill[objfill] (3.0,1.8) circle (0.55);
    \draw[objline, thick] (3.0,1.8) circle (0.55);
    \node[lbl, anchor=south] at (1.2,1.93) {detected};
    \node[lbl, text=err, anchor=south] at (3.0,2.43) {uncertain};
    \draw[<->, black!60] (2.1,1.15) -- (2.1,1.45);
    \node[lbl, anchor=west] at (2.14,1.3) {\(\delta\)};
  \end{scope}
  \begin{scope}[shift={(4.6,0)}]
    \node[head] at (0.12,3.7) {(d) Varying softness};
    \fill[gel] (0.3,0.5) rectangle (4.1,1.15);
    \node[lbl, text=gelline, anchor=north] at (2.2,0.48) {touch sensor};
    \foreach \cx in {0.85,1.75} {
      \fill[rigidfill] (\cx,1.47) circle (0.4);
      \draw[objline, thick] (\cx,1.47) circle (0.4);
    }
    \fill[objfill] ($(2.65,1.5)+(-60:0.4)$) arc (-60:240:0.4) -- cycle;
    \draw[objline, thick] ($(2.65,1.5)+(-60:0.4)$) arc (-60:240:0.4) -- cycle;
    \fill[objfill] ($(3.55,1.4)+(-36.9:0.4)$) arc (-36.9:216.9:0.4) -- cycle;
    \draw[objline, thick] ($(3.55,1.4)+(-36.9:0.4)$) arc (-36.9:216.9:0.4) -- cycle;
    \draw[gelsurf] (0.3,1.15) -- (0.55,1.15);
    \draw[gelsurf] (1.15,1.15) -- (1.45,1.15);
    \draw[gelsurf] (2.05,1.15) -- (2.45,1.15);
    \draw[gelsurf] (2.85,1.15) -- (3.23,1.15);
    \draw[gelsurf] (3.87,1.15) -- (4.1,1.15);
    \draw[black!35] (2.2,1.25) -- (2.2,2.75);
    \node[lbl, align=center] at (1.3,2.45) {stiff:\\same touch};
    \node[lbl, text=err, align=center] at (3.1,2.45) {soft:\\touch differs};
  \end{scope}
\end{tikzpicture}}
\caption{What limits a virtual touch sensor. Each panel shows a touch sensor
(teal) and an object (orange). (a)~The surface rendered from sparse sample
points (dashed) misses the detail between them, with an error that grows
with their fill distance \(\sigma\) (Proposition~\ref{prop:sampling}).
(b)~Rendering from an assumed sensor pose, shifted by \(\tau_s\) and tilted by
\(\theta_s\), changes the depth patch by at most \((1+L)(\tau_s + \theta_s\ell_s)\),
where \(\ell_s\) bounds the distance from the sensor to the surface
(Corollary~\ref{cor:pose}).
(c)~Penetration deeper than the depth-patch error \(\delta\) is always
detected, whereas a surface less than \(\delta\) from the gel, on either
side, may be missed or invented (Proposition~\ref{prop:contact}). (d)~Objects much stiffer than the
gel press it alike, whereas softer objects of the same shape press it
differently, which depth does not reveal (Corollary~\ref{cor:stiffness}).}
\label{fig:errors}
\end{figure}

Theorem~\ref{thm:stability} leaves the depth-patch error \(\delta\) open. We
bound it for rendering maps that render points sampled from the object
surface, e.g., the points measured by the depth sensor, from an estimate of
the touch-sensor pose.

\begin{figure}[t]
\centering
\resizebox{\columnwidth}{!}{%
\begin{tikzpicture}[
    /utils/exec={%
      \colorlet{objfill}{orange!15}\colorlet{objline}{orange!55!black}%
      \colorlet{gelfill}{teal!14}\colorlet{gelline}{teal!60!black}%
      \colorlet{samp}{blue!60!black}%
      \colorlet{err}{red!65!black}\colorlet{panelfill}{black!3}},
    head/.style={font=\small\bfseries, text=black!85, anchor=north west},
    lbl/.style={font=\footnotesize, text=black!70},
    ray/.style={black!45, line width=0.6pt},
    splat/.style={samp!55, line width=1.8pt, line cap=butt},
    dim/.style={black!60, {Stealth[length=3pt, width=3pt]}-{Stealth[length=3pt, width=3pt]}}]
  \foreach \px/\name in {0/(a) Points, 4.6/(b) Splats} {
    \begin{scope}[shift={(\px,0)}]
      \fill[panelfill, rounded corners=4pt] (0,0) rectangle (4.4,4.0);
      \node[head] at (0.12,3.9) {\name};
      \fill[objfill] plot[domain=0.3:4.1, samples=60] (\x, {1.45+0.11*(\x-2.2)^2})
        -- (4.1,2.75) -- (0.3,2.75) -- cycle;
      \draw[objline, thick] plot[domain=0.3:4.1, samples=60] (\x, {1.45+0.11*(\x-2.2)^2});
      \draw[black!45, densely dashed] (0.3,3.1) -- (4.1,3.1);
      \node[lbl, anchor=south west] at (0.3,3.1) {\(z_{\max}\)};
      \fill[gelfill] (0.3,0.5) rectangle (4.1,0.8);
      \draw[gelline, thick] (0.3,0.8) -- (4.1,0.8);
      \node[lbl, text=gelline, anchor=north] at (2.2,0.48) {touch sensor};
    \end{scope}
  }
  \foreach \x in {0.550,1.020,1.490,1.960,2.430,2.900,3.370,3.840} {
    \draw[ray] (\x,0.8) -- (\x,3.1);
    \fill[err] (\x,3.1) circle (1.5pt);
    \fill[gelline] (\x,0.8) circle (1.3pt);
  }
  \foreach \x/\z in {0.785/1.670, 1.255/1.548, 1.725/1.475, 2.195/1.450, 2.665/1.474, 3.135/1.546, 3.605/1.667} {
    \fill[samp] (\x,\z) circle (2pt);
  }
  \node[lbl, text=err, anchor=south east] at (4.1,3.1) {every ray misses};
  \begin{scope}[shift={(4.6,0)}]
    \foreach \x/\z in {0.785/1.670, 1.255/1.548, 1.725/1.475, 2.195/1.450, 2.665/1.474, 3.135/1.546, 3.605/1.667} {
      \draw[splat] ({\x-0.3},\z) -- ({\x+0.3},\z);
    }
    \foreach \x/\z in {0.550/1.670, 1.020/1.548, 1.490/1.475, 1.960/1.450, 2.430/1.450, 2.900/1.474, 3.370/1.546, 3.840/1.667} {
      \draw[ray] (\x,0.8) -- (\x,\z);
      \fill[gelline] (\x,0.8) circle (1.3pt);
      \fill[black!75] (\x,\z) circle (1.3pt);
    }
    \draw[black!65, line width=0.35pt] (3.605,1.73) -- (3.605,1.98);
    \draw[black!65, line width=0.35pt] (3.905,1.73) -- (3.905,1.98);
    \draw[black!65, line width=0.4pt,
          {Stealth[length=2.2pt, width=2pt]}-{Stealth[length=2.2pt, width=2pt]},
          shorten <=0.18pt, shorten >=0.18pt] (3.605,1.92) -- (3.905,1.92);
    \node[lbl, anchor=east, inner sep=1.5pt] at (3.6,1.92) {\(\rho\)};
    \foreach \x/\z in {0.785/1.670, 1.255/1.548, 1.725/1.475, 2.195/1.450, 2.665/1.474, 3.135/1.546, 3.605/1.667} {
      \fill[samp] (\x,\z) circle (2pt);
    }
  \end{scope}
\end{tikzpicture}}
\caption{Rendering a depth patch from sample points (Definition~\ref{def:point-depth}), seen from the side: an
object (orange) above a touch sensor (teal), with sample points (blue) that
lie between the rays of its pixels. (a)~Bare sample points are never hit, so
every pixel is left at \(z_{\max}\). (b)~Drawn as discs of radius \(\rho\) facing
the sensor, the sample points cover every pixel. Each pixel takes the depth of
the nearest disc its ray hits.}
\label{fig:splat}
\end{figure}
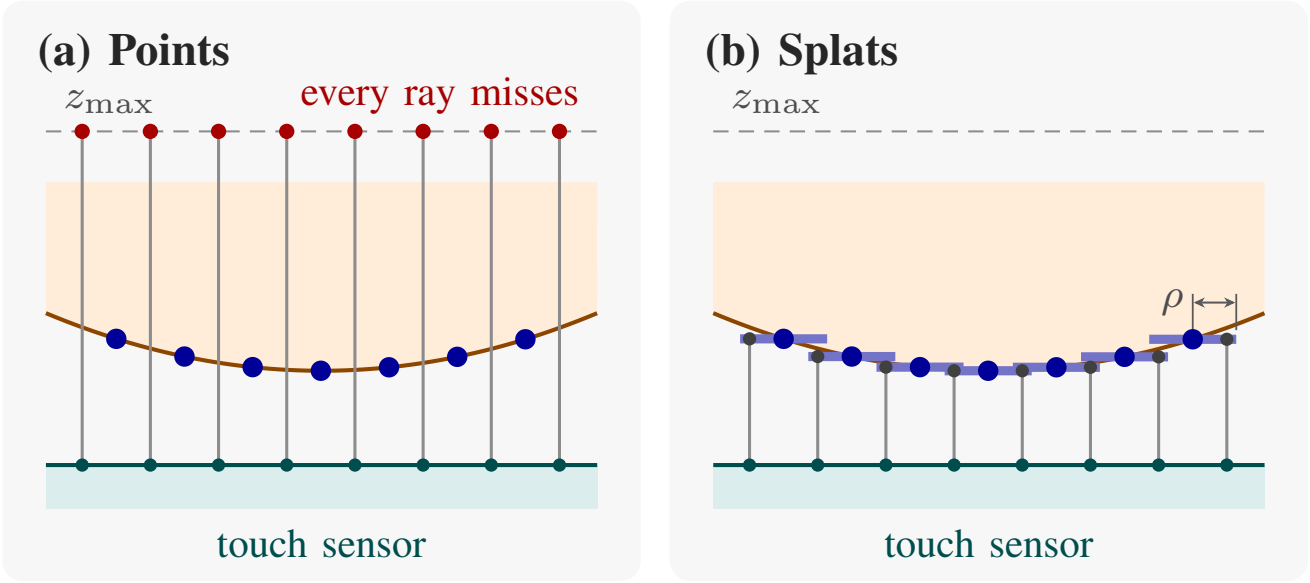

\begin{definition}[Depth patch \(\Dp_{\Obj,\Pose}\) of a point set]
\label{def:point-depth}
Let \(\Obj = \{\mathbf{o}_1, \dots, \mathbf{o}_m\} \subset \R^3\) be a
finite point set, whose elements we call \emph{sample points}. Its depth patch
\(\Dp_{\Obj,\Pose}\) at pose \(\Pose\) is defined as in
Definition~\ref{def:depth}, with each sample point drawn
as a disc facing the sensor, of radius \(\rho > 0\) called the
\emph{splat radius} (Fig.~\ref{fig:splat}). Writing a sample point in sensor coordinates as
\(\pi_{\Pose}(\mathbf{o}_i) = (\bar{\pi}_{\Pose}(\mathbf{o}_i), z_i)\), the ray of pixel \((a,b)\) hits the
sample point if the pixel center \(\mathbf{p}_{ab}\) lies within \(\rho\) of
\(\bar{\pi}_{\Pose}(\mathbf{o}_i)\):
\begin{equation}
  \Dp_{\Obj,\Pose}(a,b) =
  \begin{cases}
    \min \mathcal{Z}^{\Obj}_{ab} & \text{if } \mathcal{Z}^{\Obj}_{ab} \neq \emptyset, \\
    z_{\max} & \text{otherwise},
  \end{cases}
\end{equation}
with \(\mathcal{Z}^{\Obj}_{ab} = \{ z_i : \norm{\bar{\pi}_{\Pose}(\mathbf{o}_i) - \mathbf{p}_{ab}} \le \rho,\;
0 \le z_i \le z_{\max} \}\). We write \(\Phid(\Obj, \Pose) = \Dp_{\Obj,\Pose}\).
\end{definition}

\begin{figure}[t]
\centering
\resizebox{\columnwidth}{!}{%
\begin{tikzpicture}[
    /utils/exec={%
      \colorlet{objfill}{orange!15}\colorlet{objline}{orange!55!black}%
      \colorlet{plane}{blue!55!black}%
      \colorlet{err}{red!65!black}\colorlet{panelfill}{black!3}},
    head/.style={font=\small\bfseries, text=black!85, anchor=north west},
    lbl/.style={font=\footnotesize, text=black!70},
    ray/.style={black!45, line width=0.6pt},
    win/.style={black!60, line width=0.6pt, |-|},
    dim/.style={black!60, line width=0.4pt,
                {Stealth[length=3pt, width=2.6pt]}-{Stealth[length=3pt, width=2.6pt]}}]
  \fill[panelfill, rounded corners=4pt] (0,0) rectangle (4.4,4.0);
  \node[head] at (0.12,3.9) {(a) Holds};
  \fill[objfill] plot[domain=0.3:4.1, samples=60] (\x, {2.05+0.1*(\x-2.2)^2+0.05*(\x-2.2)})
    -- (4.1,2.85) -- (0.3,2.85) -- cycle;
  \draw[objline, thick] plot[domain=0.3:4.1, samples=60] (\x, {2.05+0.1*(\x-2.2)^2+0.05*(\x-2.2)});
  \begin{scope}
      \draw[black!45, densely dashed] (0.3,3.15) -- (4.1,3.15);
      \node[lbl, anchor=south west] at (0.3,3.15) {\(z_{\max}\)};
      \draw[plane, thick] (0.3,1.25) -- (4.1,1.25);
      \foreach \x in {0.6,3.8} { \draw[black!35, densely dotted] (\x,0.45) -- (\x,1.25); }
      \foreach \x in {1.2,3.2} { \draw[black!35, densely dotted] (\x,0.85) -- (\x,1.25); }
      \draw[win] (1.2,0.85) -- (3.2,0.85);
      \node[lbl, fill=panelfill, inner sep=1.5pt] at (2.2,0.85) {\(W_0\)};
      \draw[win] (0.6,0.45) -- (3.8,0.45);
      \node[lbl, fill=panelfill, inner sep=1.5pt] at (2.2,0.45) {\(W_\beta\)};
      \draw[dim] (3.2,0.62) -- (3.8,0.62);
      \node[lbl, anchor=west] at (3.82,0.62) {\(\beta\)};
      \draw[ray] (1.2,1.25) -- (1.2,2.100);
      \fill[black!75] (1.2,2.100) circle (1.3pt);
      \draw[ray] (1.7,1.25) -- (1.7,2.050);
      \fill[black!75] (1.7,2.050) circle (1.3pt);
      \draw[ray] (2.2,1.25) -- (2.2,2.050);
      \fill[black!75] (2.2,2.050) circle (1.3pt);
      \draw[ray] (3.2,1.25) -- (3.2,2.200);
      \fill[black!75] (3.2,2.200) circle (1.3pt);
      \draw[black!75, line width=0.8pt] (2.7,1.25) -- (2.7,2.100);
      \fill[black!75] (2.7,2.100) circle (1.6pt);
      \node[lbl, anchor=west] at (2.74,1.675) {\(\gamma(\mathbf{p})\)};
      \foreach \x in {1.2,1.7,2.2,2.7,3.2} { \fill[plane] (\x,1.25) circle (1.5pt); }
      \node[lbl, anchor=north, inner sep=1pt] at (2.7,1.22) {\(\mathbf{p}\)};
      \node[lbl] at (3.35,2.55) {slope \(\le L\)};
  \end{scope}
  \begin{scope}[shift={(4.6,0)}]
    \fill[panelfill, rounded corners=4pt] (0,0) rectangle (4.4,4.0);
    \node[head] at (0.12,3.9) {(b) Fails};
    \fill[objfill] (2.2,2.3) circle (0.75);
    \draw[objline, thick] (2.2,2.3) circle (0.75);
      \draw[black!45, densely dashed] (0.3,3.15) -- (4.1,3.15);
      \node[lbl, anchor=south west] at (0.3,3.15) {\(z_{\max}\)};
      \draw[plane, thick] (0.3,1.25) -- (4.1,1.25);
      \foreach \x in {0.6,3.8} { \draw[black!35, densely dotted] (\x,0.45) -- (\x,1.25); }
      \foreach \x in {1.2,3.2} { \draw[black!35, densely dotted] (\x,0.85) -- (\x,1.25); }
      \draw[win] (1.2,0.85) -- (3.2,0.85);
      \node[lbl, fill=panelfill, inner sep=1.5pt] at (2.2,0.85) {\(W_0\)};
      \draw[win] (0.6,0.45) -- (3.8,0.45);
      \node[lbl, fill=panelfill, inner sep=1.5pt] at (2.2,0.45) {\(W_\beta\)};
      \draw[dim] (3.2,0.62) -- (3.8,0.62);
      \node[lbl, anchor=west] at (3.82,0.62) {\(\beta\)};
    \draw[ray] (1.2,1.25) -- (1.2,3.15);
    \fill[err] (1.2,3.15) circle (1.5pt);
    \draw[ray] (1.7,1.25) -- (1.7,1.741);
    \fill[black!75] (1.7,1.741) circle (1.3pt);
    \draw[ray] (2.2,1.25) -- (2.2,1.550);
    \fill[black!75] (2.2,1.550) circle (1.3pt);
    \draw[ray] (2.7,1.25) -- (2.7,1.741);
    \fill[black!75] (2.7,1.741) circle (1.3pt);
    \draw[ray] (3.2,1.25) -- (3.2,3.15);
    \fill[err] (3.2,3.15) circle (1.5pt);
    \foreach \x in {1.2,1.7,2.2,2.7,3.2} { \fill[plane] (\x,1.25) circle (1.5pt); }
    \node[lbl, text=err, anchor=south east] at (4.1,3.15) {edge rays miss};
    \node[lbl, anchor=west] (sil) at (3.35,2.0) {silhouette};
    \draw[black!45, thin] (sil.west) -- (2.905,2.043);
  \end{scope}
\end{tikzpicture}}
\caption{The surface assumption (Assumption~\ref{ass:graph}), seen from the side. The window \(W_0\)
spans the pixel centers on the viewing plane (blue); \(W_\beta\) extends it by \(\beta\). (a)~The assumption holds: every ray from \(W_\beta\) first
hits the surface at a depth \(\gamma(\mathbf{p}) \le z_{\max}\), which
changes with slope at most \(L\). (b)~It fails for an object smaller than
the window: rays near the edge miss the object; near its silhouette the
depth changes arbitrarily fast.}
\label{fig:graph}
\end{figure}
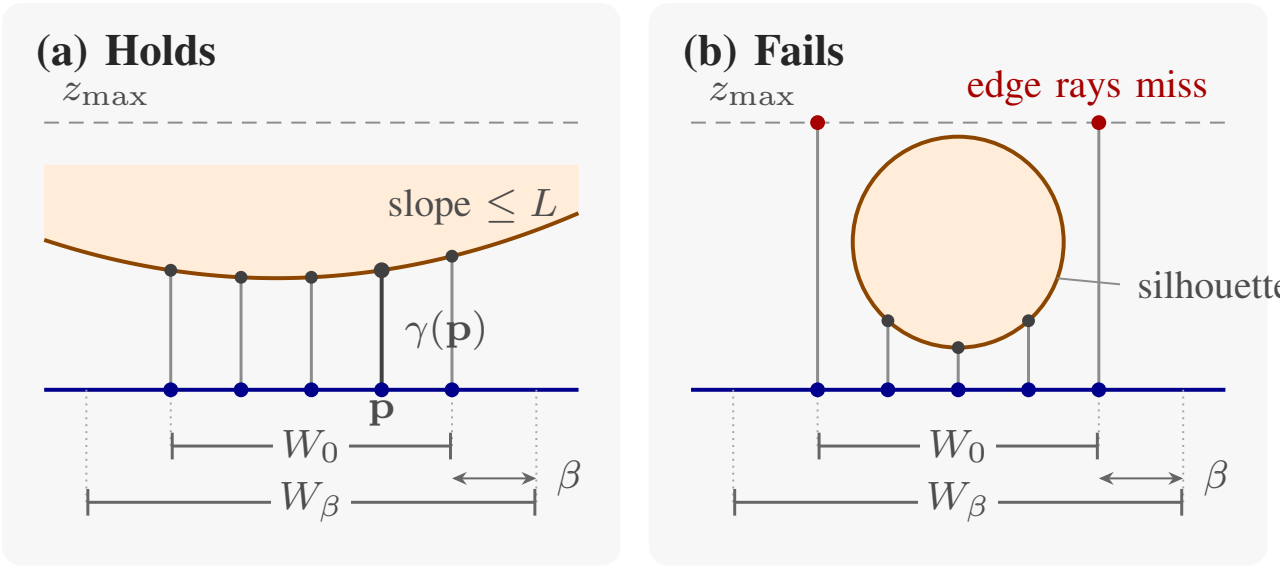

For \(\beta \ge 0\), let the \emph{window} \(W_\beta \subset \R^2\) be the set
of points of the viewing plane within distance \(\beta\) of the convex hull of
the pixel centers (Fig.~\ref{fig:graph}).

\begin{assumption}[Regular surface]
\label{ass:graph}
At pose \(\Pose\), over the window \(W_\beta\):
(i)~no point of the surface lies behind the viewing plane, i.e.,
\(\pi_{\Pose}(\Surf) \cap (W_\beta \times (-\infty, 0)) = \emptyset\);
(ii)~every ray from \(\mathbf{p} \in W_\beta\) hits the surface within
\(z_{\max}\), first at the depth
\begin{equation}
  \gamma(\mathbf{p}) = \min\bigl\{ z \ge 0 : (\mathbf{p}, z) \in
  \pi_{\Pose}(\Surf)\bigr\};
\end{equation}
(iii)~this depth changes with slope at most \(L\), i.e.,
\(|\gamma(\mathbf{p}) - \gamma(\mathbf{q})| \le L \norm{\mathbf{p} - \mathbf{q}}\).
See Fig.~\ref{fig:graph}.
\end{assumption}

In words, the object covers the whole window. The sensor sees its surface
without gaps or jumps, with slope at most \(L\) (Fig.~\ref{fig:graph}a). The
margin \(\beta\) is set by each result below to the largest sideways distance
between a pixel center and a surface point that can affect its rendered
depth. In Proposition~\ref{prop:sampling}, a sample point affects the pixel if
it lies within the splat radius \(\rho\) of the pixel center. Since the point
also lies within the sampling precision \(\epsilon\) of the surface, a distance
in 3D that bounds the sideways offset too, \(\beta = \rho + \epsilon\). The assumption fails at the
silhouette of an object smaller than the window (Fig.~\ref{fig:graph}b),
where the errors are confined to within \(\beta\) of the silhouette. We call
\(\Surf_{\Pose} = \{\pi_{\Pose}^{-1}(\mathbf{p}, \gamma(\mathbf{p})) :
\mathbf{p} \in W_0\}\) the part of \(\Surf\) \emph{seen by the touch sensor}.

\begin{figure}[t]
\centering
\resizebox{\columnwidth}{!}{%
\begin{tikzpicture}[
    /utils/exec={%
      \colorlet{objfill}{orange!15}\colorlet{objline}{orange!55!black}%
      \colorlet{gelfill}{teal!14}\colorlet{gelline}{teal!60!black}%
      \colorlet{samp}{blue!60!black}%
      \colorlet{err}{red!65!black}\colorlet{panelfill}{black!3}},
    head/.style={font=\small\bfseries, text=black!85, anchor=north west},
    lbl/.style={font=\footnotesize, text=black!70, align=center},
    sym/.style={font=\small, text=black!85},
    dist/.style={black!85, line width=1.3pt},
    dim/.style={line width=0.4pt,
      {Stealth[length=2.6pt, width=2.4pt]}-{Stealth[length=2.6pt, width=2.4pt]}}]
  \fill[panelfill, rounded corners=4pt] (0,0.55) rectangle (4.4,4.0);
  \node[head] at (0.12,3.9) {(a) Sampling};
  \fill[objfill] plot[domain=0.3:4.1, samples=80] (\x, {1.75+0.18*sin(1.6*\x r)}) -- (4.1,3.0) -- (0.3,3.0) -- cycle;
  \draw[samp!60, densely dotted, line width=0.6pt] (0.242,2.059) -- (0.293,2.071) -- (0.344,2.084) -- (0.395,2.095) -- (0.446,2.106) -- (0.498,2.116) -- (0.550,2.125) -- (0.602,2.134) -- (0.654,2.141) -- (0.707,2.147) -- (0.760,2.153) -- (0.813,2.157) -- (0.866,2.160) -- (0.919,2.162) -- (0.973,2.163) -- (1.026,2.163) -- (1.079,2.161) -- (1.132,2.158) -- (1.185,2.154) -- (1.238,2.149) -- (1.291,2.143) -- (1.343,2.136) -- (1.396,2.128) -- (1.448,2.119) -- (1.500,2.109) -- (1.551,2.099) -- (1.602,2.088) -- (1.653,2.076) -- (1.704,2.063) -- (1.754,2.050) -- (1.804,2.037) -- (1.853,2.024) -- (1.903,2.010) -- (1.951,1.996) -- (2.000,1.982) -- (2.048,1.968) -- (2.096,1.954) -- (2.143,1.941) -- (2.190,1.928) -- (2.237,1.915) -- (2.283,1.902) -- (2.329,1.890) -- (2.375,1.879) -- (2.420,1.868) -- (2.466,1.858) -- (2.510,1.849) -- (2.555,1.841) -- (2.599,1.833) -- (2.643,1.826) -- (2.687,1.820) -- (2.730,1.815) -- (2.773,1.811) -- (2.817,1.807) -- (2.860,1.805) -- (2.903,1.803) -- (2.946,1.803) -- (2.988,1.803) -- (3.031,1.805) -- (3.075,1.807) -- (3.118,1.811) -- (3.161,1.815) -- (3.204,1.820) -- (3.248,1.826) -- (3.292,1.833) -- (3.336,1.841) -- (3.381,1.849) -- (3.426,1.859) -- (3.471,1.869) -- (3.516,1.879) -- (3.562,1.891) -- (3.608,1.903) -- (3.654,1.915) -- (3.701,1.928) -- (3.748,1.941) -- (3.795,1.955) -- (3.843,1.968) -- (3.891,1.982) -- (3.940,1.996) -- (3.989,2.010) -- (4.038,2.024);
  \draw[samp!60, densely dotted, line width=0.6pt] (0.358,1.607) -- (0.403,1.619) -- (0.449,1.630) -- (0.494,1.640) -- (0.539,1.649) -- (0.583,1.658) -- (0.628,1.666) -- (0.672,1.673) -- (0.715,1.679) -- (0.759,1.684) -- (0.802,1.689) -- (0.845,1.692) -- (0.888,1.695) -- (0.931,1.696) -- (0.974,1.697) -- (1.017,1.697) -- (1.060,1.695) -- (1.103,1.693) -- (1.146,1.690) -- (1.190,1.686) -- (1.233,1.681) -- (1.277,1.675) -- (1.321,1.668) -- (1.365,1.661) -- (1.409,1.652) -- (1.454,1.643) -- (1.499,1.633) -- (1.544,1.623) -- (1.590,1.611) -- (1.636,1.600) -- (1.682,1.587) -- (1.729,1.575) -- (1.776,1.561) -- (1.823,1.548) -- (1.871,1.534) -- (1.919,1.520) -- (1.968,1.506) -- (2.016,1.493) -- (2.065,1.479) -- (2.115,1.465) -- (2.165,1.452) -- (2.215,1.439) -- (2.265,1.426) -- (2.316,1.414) -- (2.367,1.403) -- (2.419,1.392) -- (2.471,1.382) -- (2.523,1.373) -- (2.575,1.365) -- (2.627,1.358) -- (2.680,1.352) -- (2.733,1.346) -- (2.786,1.342) -- (2.839,1.339) -- (2.892,1.338) -- (2.946,1.337) -- (2.999,1.338) -- (3.052,1.339) -- (3.105,1.342) -- (3.158,1.346) -- (3.211,1.352) -- (3.264,1.358) -- (3.316,1.365) -- (3.369,1.373) -- (3.421,1.382) -- (3.472,1.392) -- (3.524,1.403) -- (3.575,1.414) -- (3.626,1.426) -- (3.676,1.439) -- (3.726,1.452) -- (3.776,1.465) -- (3.826,1.479) -- (3.875,1.493) -- (3.924,1.507) -- (3.972,1.521) -- (4.020,1.534) -- (4.068,1.548) -- (4.115,1.562) -- (4.162,1.575);
  \draw[objline, thick] plot[domain=0.3:4.1, samples=80] (\x, {1.75+0.18*sin(1.6*\x r)});
  \draw[dist] (0.920,1.869) -- (1.220,1.987);
  \draw[dist] (2.244,1.672) -- (1.700,1.844);
  \draw[dist] (3.620,1.425) -- (3.565,1.652);
  \draw[objline, fill=white, thick] (2.244,1.672) circle (1.8pt);
  \fill[objline] (3.565,1.652) circle (1.3pt);
  \foreach \x/\z in {0.380/1.893, 0.920/1.869, 1.220/1.987, 1.700/1.844, 2.800/1.545, 3.220/1.637, 3.620/1.425, 4.050/1.825} { \fill[samp] (\x,\z) circle (1.7pt); }
  \node[sym, anchor=south, fill=objfill, inner sep=1pt] at (1.010,2.028) {\(r\)};
  \node[sym, anchor=south, fill=objfill, inner sep=1pt] at (2.052,1.858) {\(\sigma\)};
  \node[sym, anchor=west] at (3.753,1.438) {\(\epsilon\)};
  \begin{scope}[shift={(4.6,0)}]
    \fill[panelfill, rounded corners=4pt] (0,0.55) rectangle (4.4,4.0);
    \node[head] at (0.12,3.9) {(b) Bound};
    \fill[err, opacity=0.10] plot[domain=0.5:3.9, samples=60] (\x, {1.75+0.18*sin(1.6*\x r)+0.570})
      -- plot[domain=3.9:0.5, samples=60] (\x, {1.75+0.18*sin(1.6*\x r)-0.570}) -- cycle;
    \draw[err!70, densely dotted] plot[domain=0.5:3.9, samples=60] (\x, {1.75+0.18*sin(1.6*\x r)+0.570});
    \draw[err!70, densely dotted] plot[domain=0.5:3.9, samples=60] (\x, {1.75+0.18*sin(1.6*\x r)-0.570});
    \draw[objline, thick] plot[domain=0.3:4.1, samples=80] (\x, {1.75+0.18*sin(1.6*\x r)});
    \foreach \x/\z in {0.500/1.869, 0.700/1.869, 0.900/1.869, 1.100/1.869, 1.300/1.844, 1.500/1.844, 1.700/1.844, 1.900/1.844, 2.100/1.844, 2.300/1.545, 2.500/1.545, 2.700/1.545, 2.900/1.545, 3.100/1.425, 3.300/1.425, 3.500/1.425, 3.700/1.425, 3.900/1.425} { \fill[black!80] (\x,\z) circle (1.5pt); }
    \draw[err, dim] (3.9,1.742) -- (3.9,2.312);
    \node[lbl, text=err, anchor=south east] at (3.95,2.362) {\(\max\bigl(L\rho + (1+L)\epsilon,\, \sigma\bigr)\)};
    \draw[objline, thick] (0.25,0.85) -- (0.6,0.85);
    \node[lbl, anchor=west] at (0.65,0.85) {true depth};
    \fill[black!80] (2.1,0.85) circle (1.5pt);
    \node[lbl, anchor=west] at (2.21,0.85) {rendered depth};
  \end{scope}
\end{tikzpicture}}
\caption{Sampling and the bound it implies, seen from the side. (a)~Sample points
(blue) measured on a surface (orange) with noise: the separation distance
\(r\) is the smallest distance between two sample points, the fill distance
\(\sigma\) the largest distance from a surface point (open circle) to its
nearest sample point, and the sampling precision \(\epsilon\) the largest distance
from a sample point to the surface, so that all sample points lie within \(\epsilon\) of
it (dotted). (b)~At every pixel, the depth rendered from these sample points with
splat radius \(\rho \ge \sigma\) (dots) lies within
\(\max(L\rho + (1+L)\epsilon, \sigma)\) of the true depth (shaded band), as
Proposition~\ref{prop:sampling} guarantees.}
\label{fig:sampling}
\end{figure}
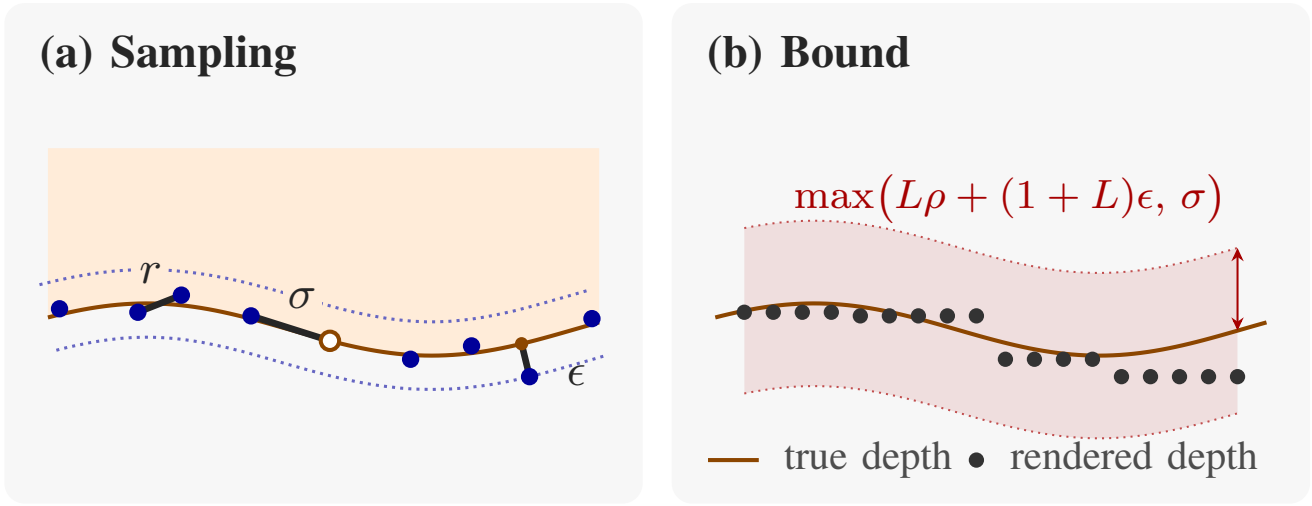

How well the points represent a surface \(\Surf\) depends on how densely and
how accurately they sample it (Fig.~\ref{fig:sampling}).

\begin{definition}[Separation distance \(r\), fill distance \(\sigma\), and sampling precision \(\epsilon\)]
The \emph{separation distance} \(r\) of \(\Obj\) is the smallest distance
between two of its points, its \emph{fill distance} \(\sigma\) at pose \(\Pose\) is the largest
distance from a surface point of \(\Surf_{\Pose}\) to its nearest sample
point, and its \emph{sampling precision} \(\epsilon\) is the largest distance from
a sample point to the surface \(\Surf\):
\begin{equation}
  \begin{aligned}
    r &= \min_{i \ne j} \norm{\mathbf{o}_i - \mathbf{o}_j}, \quad
    \sigma = \sup_{\mathbf{x} \in \Surf_{\Pose}} \min_{i} \norm{\mathbf{x} - \mathbf{o}_i}, \\
    \epsilon &= \max_{i} \inf_{\mathbf{x} \in \Surf} \norm{\mathbf{o}_i - \mathbf{x}}.
  \end{aligned}
\end{equation}
See Fig.~\ref{fig:sampling}a.
\end{definition}

For the points measured by the depth sensor, \(\sigma\) is how finely the
depth sensor samples the part seen by the touch sensor, unbounded if some of
that part is hidden from it, and \(\epsilon\) is its depth precision. The separation distance alone does not bound the
error: points may cluster, with a small \(r\), and still leave parts of the
surface unsampled. We call a point set \emph{quasi-uniform}
if its fill distance is at most a constant multiple of its separation
distance, \(\sigma \le c_q\, r\). For such sets, the standard setting of
scattered-data approximation~\cite{wendland2004scattered}, the two notions
coincide up to the constant \(c_q\).

\begin{proposition}[Bound on \(\delta\) from sampling]
\label{prop:sampling}
Let \(\Obj\) have sampling precision \(\epsilon\) and fill distance
\(\sigma \le \rho\) at \(\Pose\), and let
Assumption~\ref{ass:graph} hold there with \(\beta = \rho + \epsilon\) and
\(\gamma \ge \epsilon\) on \(W_\beta\). Then
\begin{equation}
  \ninf{\Dp_{\Obj,\Pose} - \Dp_{\Surf,\Pose}} \le
  \max\bigl(L\rho + (1 + L)\epsilon,\, \sigma\bigr).
  \label{eq:sampling}
\end{equation}
For sample points on the surface (\(\epsilon = 0\)) that are quasi-uniform with
separation distance \(r\), and \(\rho = c_q r\), the bound is \(\max(L, 1)\,c_q\, r\).
See Figs.~\ref{fig:errors}a and~\ref{fig:sampling}b.
\end{proposition}

The same argument covers a perturbed surface.

\begin{lemma}[Bound on \(\delta\) from surface perturbation]
\label{lem:hausdorff}
Let \(\Surf\) and \(\tilde\Surf\) satisfy Assumption~\ref{ass:graph} at pose
\(\Pose\) with \(\beta = \eta\) and a common slope bound \(L\). If their Hausdorff
distance satisfies \(d_H(\Surf, \tilde\Surf) \le \eta\) over the observed
volume, then
\begin{equation}
  \ninf{\Dp_{\tilde\Surf,\Pose} - \Dp_{\Surf,\Pose}} \le (1 + L)\,\eta.
\end{equation}
\end{lemma}

An error in the touch-sensor pose acts like a perturbation of the surface.

\begin{corollary}[Bound on \(\delta\) from pose error]
\label{cor:pose}
Let \(\PoseT = (\hat{\mathbf{R}}, \hat{\mathbf{t}})\) be an estimate
of \(\Pose\), with rotation error \(\theta_s\), the rotation angle of
\(\mathbf{R}^{\top}\hat{\mathbf{R}}\), and translation error
\(\tau_s = \norm{\hat{\mathbf{t}} - \mathbf{t}}\), and let \(\ell_s\) bound the
distance from the touch-sensor origin of all surface points in the observed volume.
If Assumption~\ref{ass:graph} holds at both poses with \(\beta = \eta\) and
a common slope bound \(L\), then
\begin{equation}
  \ninf{\Dp_{\Surf,\PoseT} - \Dp_{\Surf,\Pose}} \le (1 + L)\,\eta,
  \quad \eta = \tau_s + \theta_s \ell_s.
\end{equation}
See Fig.~\ref{fig:errors}b.
\end{corollary}

\begin{corollary}[Bound on \(\delta\) from sampling and pose error]
\label{cor:samples}
Let \(\Phir\) render a point set \(\Obj\) at an estimated touch-sensor pose
\(\PoseT\) with errors \((\theta_s, \tau_s)\), \(\hat{\Dp} = \Dp_{\Obj,\PoseT}\),
where \(\Obj\) has sampling precision \(\epsilon\) and fill distance
\(\sigma \le \rho\) at \(\PoseT\). Let Assumption~\ref{ass:graph} hold
at \(\Pose\) and \(\PoseT\) with \(\beta = \max(\rho + \epsilon, \tau_s +
\theta_s\ell_s)\) and a common slope bound \(L\), and with \(\gamma \ge \epsilon\) at
\(\PoseT\). Then
\begin{equation}
  \delta \le \max\bigl(L\rho + (1 + L)\epsilon,\, \sigma\bigr)
  + (1 + L)(\tau_s + \theta_s\ell_s).
  \label{eq:delta_samples}
\end{equation}
\end{corollary}

\begin{proof}
By the triangle inequality through \(\Dp_{\Surf,\PoseT}\), apply
Proposition~\ref{prop:sampling} at \(\PoseT\) and Corollary~\ref{cor:pose}.
\end{proof}

With \(\rho = \sigma\), the sampling term of~\eqref{eq:delta_samples} is at most
\(\max(L, 1)\,\sigma + (1 + L)\,\epsilon\). For the points measured by the
depth sensor, the depth-patch error thus grows at most linearly with the spacing and
the precision of the measurement on the surface seen by the touch sensor, and
with the touch-sensor pose error. At contact, however, this surface is
typically hidden from the depth sensor, so that \(\sigma\) is unbounded.
Section~\ref{sec:r2s2r} instead renders a reconstruction.

The depth-patch error also limits how reliably contact itself is rendered.

\begin{proposition}[Bound on the penetration error and contact detection]
\label{prop:contact}
Let \(\hat{\pen}\) and \(\pen^{*}\) be the penetration fields of
\(\hat{\Dp}\) and \(\Dp^{*}\). Then
\(\ninf{\hat{\pen} - \pen^{*}} \le \delta\). In particular, every pixel with
true penetration \(\pen^{*}(a,b) > \delta\) is rendered in contact, and every
pixel rendered in contact has true depth \(\Dp^{*}(a,b) < z_0 + \delta\).
See Fig.~\ref{fig:errors}c.
\end{proposition}

\begin{proof}
The penetration is a 1-Lipschitz function of the depth, which gives the
bound. The two consequences follow:
\begin{equation*}
  \begin{aligned}
    \pen^{*}(a,b) > \delta \;&\Rightarrow\;
      \hat{\pen}(a,b) \ge \pen^{*}(a,b) - \delta > 0, \\
    \hat{\pen}(a,b) > 0 \;&\Rightarrow\; \hat{\Dp}(a,b) < z_0
      \;\Rightarrow\; \Dp^{*}(a,b) < z_0 + \delta.
  \end{aligned}
  \qedhere
\end{equation*}
\end{proof}

Since a touch sensor responds to indentations of its gel, \(\delta\) must be small compared with the indentations to be resolved.

\section{STIFFNESS LIMIT}
\label{sec:limits}

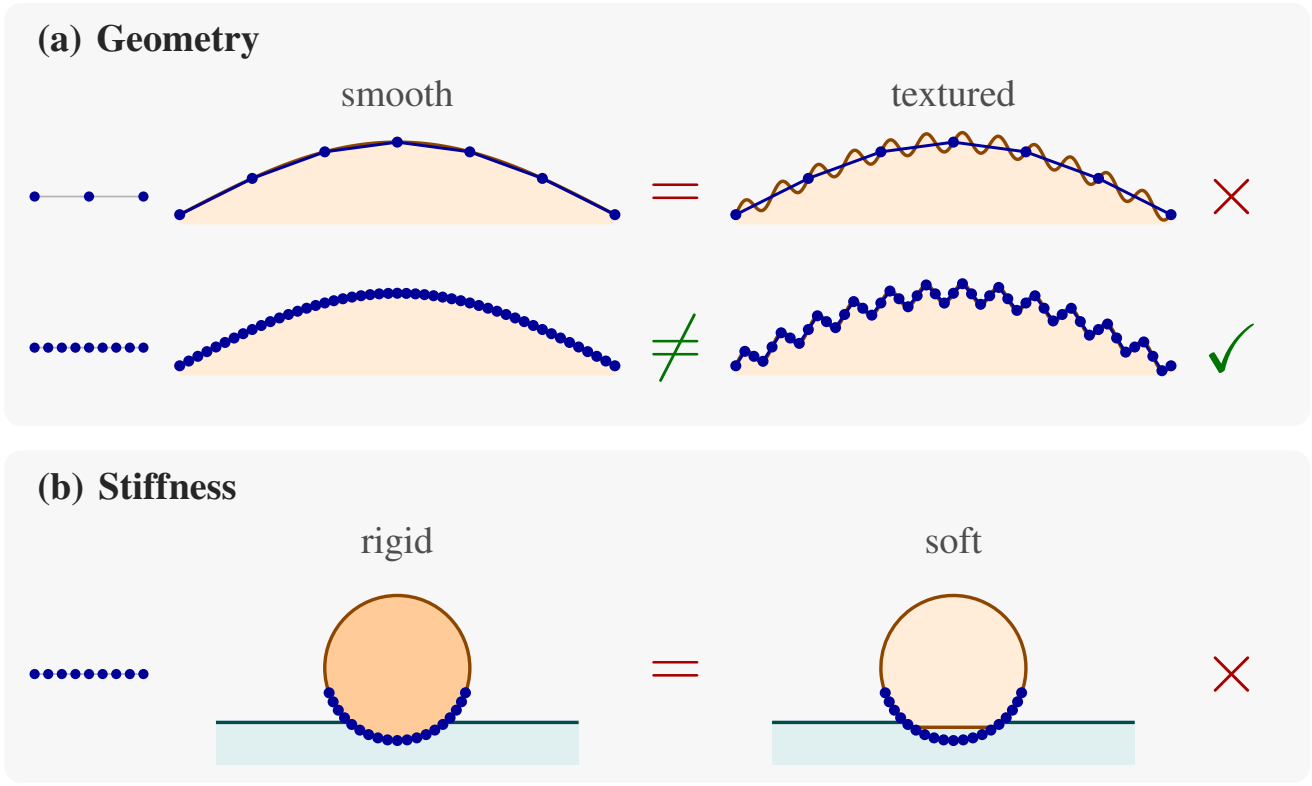
\begin{figure}[t]
\centering
\newcommand{\sampicon}[3]{%
  \draw[black!30] ({#1-0.45},#2) -- ({#1+0.45},#2);
  \foreach \i in {0,...,#3} {
    \fill[samp] ({#1-0.45+0.9*\i/#3},#2) circle (1.2pt);
  }}
\resizebox{\columnwidth}{!}{%
\begin{tikzpicture}[>=Stealth,
    label/.style={font=\small, text=black!70},
    rowlab/.style={label, anchor=east},
    colhead/.style={label, anchor=base},
    head/.style={font=\small\bfseries, text=black!85, anchor=base west},
    rel/.style={font=\LARGE, anchor=center},
    verdict/.style={font=\LARGE, anchor=center},
    /utils/exec={%
      \colorlet{objfill}{orange!15}\colorlet{objline}{orange!55!black}%
      \colorlet{samp}{blue!60!black}%
      \colorlet{gelfill}{teal!12}\colorlet{gelline}{teal!60!black}%
      \colorlet{yes}{green!45!black}\colorlet{no}{red!65!black}%
      \colorlet{panelfill}{black!3}\colorlet{rigidfill}{orange!40}}]
  \fill[panelfill, rounded corners=4pt] (-1.45,-0.5) rectangle (9.35,3.0);
  \fill[panelfill, rounded corners=4pt] (-1.45,-3.45) rectangle (9.35,-0.7);
  \node[head] at (-1.3,2.6) {(a) Geometry};
  \node[colhead] at (1.8,2.15) {smooth};
  \node[colhead] at (6.4,2.15) {textured};
  \foreach \ox/\amp in {0/0, 4.6/0.08} {
    \foreach \oy/\step in {1.25/0.6, 0/0.075} {
      \fill[objfill] plot[domain=0:3.6, samples=200, smooth]
        ({\ox+\x}, {\oy+0.6*sin(50*\x)+\amp*sin(1200*\x)})
        -- ({\ox+3.6},{\oy-0.08}) -- ({\ox},{\oy-0.08}) -- cycle;
      \draw[objline, thick] plot[domain=0:3.6, samples=200, smooth]
        ({\ox+\x}, {\oy+0.6*sin(50*\x)+\amp*sin(1200*\x)});
      \draw[samp, line width=0.7pt] plot[domain=0:3.6, samples={3.6/\step+1}]
        ({\ox+\x}, {\oy+0.6*sin(50*\x)+\amp*sin(1200*\x)});
      \foreach \k in {0,\step,...,3.61} {
        \fill[samp] ({\ox+\k}, {\oy+0.6*sin(50*\k)+\amp*sin(1200*\k)}) circle (1.3pt);
      }
    }
  }
  \sampicon{-0.75}{1.4}{2}
  \sampicon{-0.75}{0.15}{8}
  \node[rel, text=no]      at (4.1,1.4)  {\(=\)};
  \node[rel, text=yes]     at (4.1,0.15) {\(\neq\)};
  \node[verdict, text=no]  at (8.7,1.4)  {\(\times\)};
  \node[verdict, text=yes] at (8.7,0.15) {\(\checkmark\)};
  \node[head] at (-1.3,-1.1) {(b) Stiffness};
  \node[colhead] at (1.8,-1.55) {rigid};
  \node[colhead] at (6.4,-1.55) {soft};
  \begin{scope}[yshift=-2.95cm]
  \foreach \cx in {1.8,6.4} {
    \fill[gelfill] ({\cx-1.5},-0.35) rectangle ({\cx+1.5},0);
  }
  \draw[gelline, thick] (0.3,0) -- ({1.8-0.397},0);
  \draw[gelline, thick] ({1.8+0.397},0) -- (3.3,0);
  \fill[rigidfill] (1.8,0.45) circle (0.6);
  \draw[objline, thick] (1.8,0.45) circle (0.6);
  \draw[gelline, thick] (4.9,0) -- ({6.4-0.346},0);
  \draw[gelline, thick] ({6.4+0.346},0) -- (7.9,0);
  \fill[objfill] ($(6.4,0.45)+(-54.8:0.6)$) arc (-54.8:234.8:0.6) -- cycle;
  \draw[objline, thick] ($(6.4,0.45)+(-54.8:0.6)$) arc (-54.8:234.8:0.6) -- cycle;
  \draw[objline, densely dashed] ($(6.4,0.45)+(234.8:0.6)$) arc (234.8:305.2:0.6);
  \foreach \cx in {1.8,6.4} {
    \draw[samp, line width=0.7pt] plot[domain=200:340, samples=19]
      ({\cx+0.6*cos(\x)}, {0.45+0.6*sin(\x)});
    \foreach \a in {200,207.78,...,340.1} {
      \fill[samp] ({\cx+0.6*cos(\a)}, {0.45+0.6*sin(\a)}) circle (1.3pt);
    }
  }
  \sampicon{-0.75}{0.4}{8}
  \node[rel, text=no]     at (4.1,0.4) {\(=\)};
  \node[verdict, text=no] at (8.7,0.4) {\(\times\)};
  \end{scope}
\end{tikzpicture}}
\caption{What geometry can and cannot resolve (Proposition~\ref{prop:irreducible}). The dots at the left of each
row show the sampling density of the depth sensor; \(=\) or \(\neq\) states
whether the Geometric Force is the same for both objects, and \(\checkmark\)
or \(\times\) whether this matches the real touch. (a)~A smooth and a
textured surface give the same sample points, and hence the same force, at coarse
resolution, but not at fine resolution. (b)~A rigid and a soft object of the
same shape give the same sample points at any resolution, yet indent the gel
differently: the Geometric Force cannot tell them apart, so better geometry cannot remove this error.}
\label{fig:teaser}
\end{figure}

The Geometric Force is computed from the shape of the object, whereas the
real touch signal also depends on its stiffness. We quantify the resulting
irreducible touch error and show where stiffness enters.

\begin{proposition}[Irreducible touch error]
\label{prop:irreducible}
Let \((\Dp, \Sg)\) be random with \(\mathbb{E}\norm{\Sg}^2 < \infty\), drawn
from the data distribution, with \(\norm{\cdot}\) the Frobenius norm, and let
\(\mu(\Dp) = \mathbb{E}[\Sg \mid \Dp]\). For every measurable \(\Phit\),
\begin{multline}
  \mathbb{E}\norm{\Phit(\Dp) - \Sg}^2 = \mathbb{E}\norm{\Phit(\Dp) - \mu(\Dp)}^2
  \\ + \mathbb{E}\bigl[\operatorname{tr}\operatorname{Cov}(\Sg \mid \Dp)\bigr].
\end{multline}
Hence the mean-squared-error minimizer is \(\Phit^{*} = \mu\), whose error
\(\mathbb{E}[\operatorname{tr}\operatorname{Cov}(\Sg \mid \Dp)]\) vanishes if
and only if \(\Sg\) is almost surely a function of \(\Dp\).
See Fig.~\ref{fig:teaser}b.
\end{proposition}

\begin{proof}
Expand
\begin{equation*}
  \Phit(\Dp) - \Sg = (\Phit(\Dp) - \mu(\Dp)) + (\mu(\Dp) - \Sg).
\end{equation*}
Squaring and taking the expectation given \(\Dp\) gives
\begin{equation*}
  \begin{aligned}
  \mathbb{E}\bigl[\norm{\Phit(\Dp) - \Sg}^2 \mid \Dp\bigr]
  &= \norm{\Phit(\Dp) - \mu(\Dp)}^2 \\
  &\quad + \operatorname{tr}\operatorname{Cov}(\Sg \mid \Dp).
  \end{aligned}
\end{equation*}
The cross term drops out, since
\(\mathbb{E}[\mu(\Dp) - \Sg \mid \Dp] = 0\). The first term vanishes only for
\(\Phit = \mu\).
\end{proof}

A touch map trained with a squared loss therefore approximates the
conditional mean touch signal: the Geometric Force is the force \emph{under
the material prior of the training distribution}. The error is
irreducible only for a touch map that sees the depth patch alone. By the law
of total variance, more information can only lower it: for any context
\(\mathbf{c}\),
\(\mathbb{E}[\operatorname{tr}\operatorname{Cov}(\Sg \mid \Dp, \mathbf{c})]
\le \mathbb{E}[\operatorname{tr}\operatorname{Cov}(\Sg \mid \Dp)]\). The
context can be semantic, such as the object category. Training on a single
category, e.g., one fruit, has the same effect at the cost of generality. The
context can also be geometric. A fine depth patch determines any subsampling
of it, so finer depth resolution can only lower the error, most where surface
detail correlates with the material. Geometry cannot help, however, for
objects that share their shape at every resolution but differ in stiffness.
For them, Proposition~\ref{prop:factorization} below shows that the error
equals the variance of a stiffness ratio times the squared penetration, so it
is zero out of contact but positive in contact.

To see where object dependence enters, we consider an idealized contact model.

\begin{figure}[t]
\centering
\begin{tikzpicture}[scale=0.95, >=Stealth,
    math/.style={font=\small},
    word/.style={font=\scriptsize, text=black!80},
    level/.style={font=\scriptsize, align=left, anchor=west, text=black!80},
    dim/.style={{Stealth[length=3pt]}-{Stealth[length=3pt]}, line width=0.5pt,
                draw=black!80},
    spring/.style={decorate, thick, decoration={coil, aspect=0.45,
                   segment length=4pt, amplitude=6pt,
                   pre length=5pt, post length=5pt}},
    /utils/exec={%
      \colorlet{objfill}{orange!12}\colorlet{objline}{orange!55!black}%
      \colorlet{ovfill}{orange!45}\colorlet{core}{orange!30}%
      \colorlet{gelfill}{teal!10}\colorlet{gelline}{teal!60!black}%
      \colorlet{body}{black!15}}]
  \fill[body, rounded corners=2pt] (0,-0.4) rectangle (1.4,0);
  \fill[gelfill] (0,0) rectangle (1.4,2.0);
  \fill[objfill] (0,0.8) rectangle (1.4,4.2);
  \fill[ovfill] (0,0.8) rectangle (1.4,2.0);
  \draw[objline, thick] (0,0.8) -- (1.4,0.8);
  \node[font=\small] at (0.7,3.1) {object};
  \node[word] at (0.7,0.4) {gel};
  \node[word, text=orange!35!black] at (0.7,1.4) {overlap};
  \draw[dim] (-0.2,0.8) -- (-0.2,2.0);
  \node[math, anchor=east] at (-0.25,1.4) {\(\pen\)};
  \node[word, anchor=north] at (0.7,-0.5) {(a) frozen geometry};
  \fill[body, rounded corners=2pt] (2.8,-0.4) rectangle (4.2,0);
  \fill[gelfill] (2.8,0) rectangle (4.2,1.3);
  \fill[objfill] (2.8,1.3) rectangle (4.2,3.3);
  \fill[core] (2.8,3.3) rectangle (4.2,4.2);
  \draw[black!70, thick] (2.8,1.3) -- (4.2,1.3);
  \draw[black!70, thick] (2.8,3.3) -- (4.2,3.3);
  \draw[draw=gelline, dashed, opacity=0.6] (2.8,2.0) -- (4.2,2.0);
  \draw[draw=objline, dashed, opacity=0.6] (2.8,0.8) -- (4.2,0.8);
  \draw[->, gelline, thick] (4.0,2.0) -- (4.0,1.36);
  \draw[->, objline, thick] (4.0,0.8) -- (4.0,1.24);
  \draw[spring, gelline] (3.5,0) -- (3.5,1.3);
  \draw[spring, objline] (3.5,1.3) -- (3.5,3.3);
  \node[math, text=gelline, anchor=east] at (3.24,0.5) {\(K_g\)};
  \node[math, text=objline, anchor=east] at (3.24,2.6) {\(K_o\)};
  \node[word] at (3.5,3.75) {object core};
  \node[word, anchor=north] at (3.5,-0.5) {(b) elastic contact};
  \draw[dim] (2.6,1.3) -- (2.6,2.0);
  \node[math, anchor=east] at (2.55,1.65) {\(\pen_g\)};
  \draw[dim] (2.6,0.8) -- (2.6,1.3);
  \node[math, anchor=east] at (2.55,1.05) {\(\pen_o\)};
  \draw[gelline, dashed] (1.4,2.0) -- (2.8,2.0);
  \draw[objline, dashed] (1.4,0.8) -- (2.8,0.8);
  \draw[gelline, dashed] (4.2,2.0) -- (4.6,2.0);
  \draw[objline, dashed] (4.2,0.8) -- (4.6,0.8);
  \node[level] at (4.65,2.0) {gel surface\\at rest};
  \node[level] at (4.65,0.8) {frozen object\\surface};
\end{tikzpicture}
\caption{Elastic-foundation contact at one pixel (Assumption~\ref{ass:winkler} and Proposition~\ref{prop:factorization}). (a)~The frozen object
overlaps the gel by the penetration \(\pen\). (b)~In contact, the gel and the
object act as springs in series. The penetration splits into the gel
indentation \(\pen_g\) and the object indentation \(\pen_o\) at equal
pressure, so \(\pen_g = \kappa\pen\): the arrows show both surfaces pushed
to the interface, compressing both springs. A rigid object does not deform, so
the gel indentation equals the penetration.}
\label{fig:springs}
\end{figure}
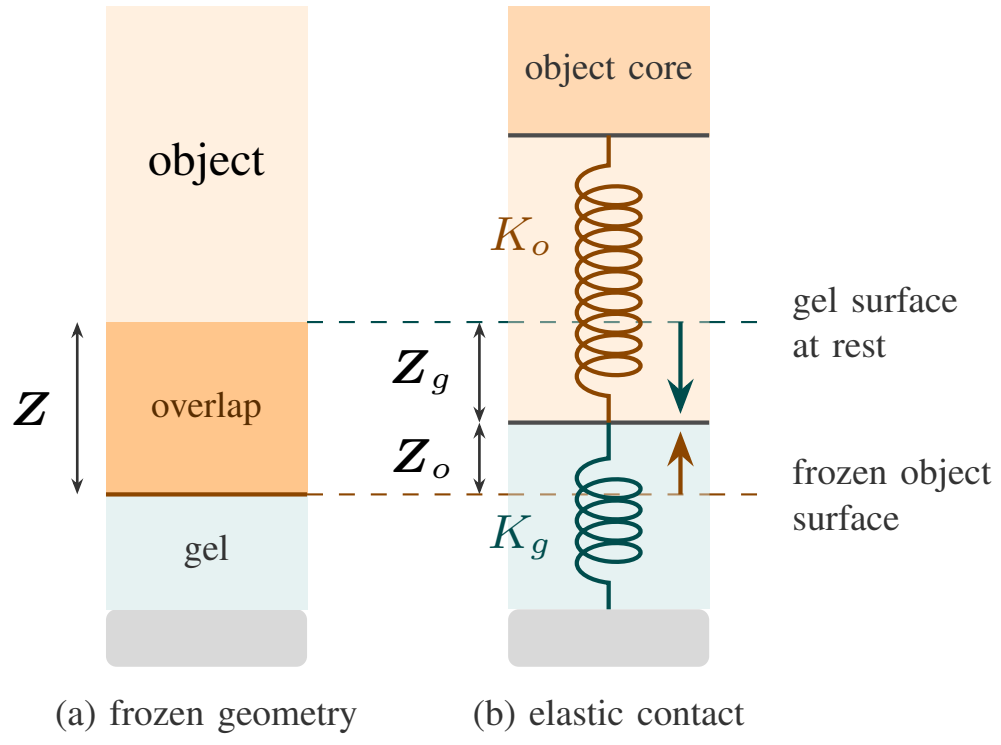

\begin{assumption}[Elastic-foundation contact]
\label{ass:winkler}
Contact is frictionless and quasi-static, i.e., slow enough to be in
equilibrium at every instant, without inertia or rate effects. Each pixel column, of area \(\lambda^2\), behaves as two
linear elastic springs in series without lateral coupling, a Winkler
foundation~\cite{johnson1985contact} (Fig.~\ref{fig:springs}), with moduli per unit area \(K_g\) for
the gel and \(K_o\) for the object. The penetration splits into gel and
object indentation, \(\pen = \pen_g + \pen_o\), at equal pressure on both
sides, \(K_g\,\fld{Z}_g = K_o\,\fld{Z}_o\) pixel-wise. The touch signal is the
gel indentation, a single channel: \(\Sg = \pen_g\).
\end{assumption}

\begin{proposition}[Factorization into touch and force maps]
\label{prop:factorization}
Under Assumption~\ref{ass:winkler}, pixel-wise,
\begin{equation*}
  \begin{gathered}
    \fld{Z}_g = \kappa\,\fld{Z},\\
    \fld{F}^{\perp} = \lambda^2 K_g\,\fld{Z}_g = \lambda^2 K_{\mathrm{eff}}\,\fld{Z},\\
    \kappa = \frac{K_o}{K_g + K_o}, \qquad
    K_{\mathrm{eff}} = \frac{K_g K_o}{K_g + K_o}.
  \end{gathered}
\end{equation*}
Hence \(\Phif^{*}\) is object-agnostic: it maps the touch signal
\(\fld{Z}_g\) to the force through \(K_g\) alone. All object dependence lies
in \(\Phit^{*}\), through \(\kappa \in (0, 1)\). If \(K_o\) is random and independent of \(\Dp\), then
\begin{equation*}
  \begin{gathered}
    \mu(\Dp) = \mathbb{E}[\kappa]\,\fld{Z},\\
    \fld{F}^{\perp} = \lambda^2\,\mathbb{E}[K_{\mathrm{eff}}]\,\fld{Z},\\
    \operatorname{tr}\operatorname{Cov}(\Sg \mid \Dp)
      = \operatorname{Var}(\kappa)\norm{\fld{Z}}^2.
  \end{gathered}
\end{equation*}
In the rigid limit \(K_o \to \infty\), \(\kappa \to 1\) and
\(\fld{Z}_g \to \fld{Z}\).
\end{proposition}

The size of the irreducible touch error depends on how widely the
object-to-gel stiffness ratio \(K_o/K_g\) varies.

\begin{corollary}[Bound on the irreducible touch error from stiffness range]
\label{cor:stiffness}
Under Assumption~\ref{ass:winkler}, let \(K_o\) be random, independent of
\(\Dp\), with values between the softest and the stiffest object
stiffness, \(K_- \le K_o \le K_+\), and let
\(\kappa_\pm = K_\pm / (K_g + K_\pm)\). Then
\begin{equation}
  \mathbb{E}\bigl[\operatorname{tr}\operatorname{Cov}(\Sg \mid \Dp)\bigr]
  \le \tfrac{1}{4}(\kappa_+ - \kappa_-)^2\, \mathbb{E}\norm{\pen}^2,
\end{equation}
with \(\kappa_+ - \kappa_- \le K_g / (K_g + K_-)\).
See Fig.~\ref{fig:errors}d.
\end{corollary}

The irreducible touch error is therefore small when all objects are much stiffer
than the gel, \(K_- \gg K_g\), or when their stiffness varies little. It is largest when some objects are softer than the gel and others stiffer.

Proposition~\ref{prop:factorization} motivates the factorization
\(\Phif \circ \Phit\): the force map depends only on the gel and can be
shared across objects, while all object-specific information enters through
the touch map. The object stiffness in \(K_{\mathrm{eff}}\) thus belongs to the
composition, not to the force map. The irreducible touch error grows quadratically with penetration
and vanishes out of contact. Finally, tactile
simulators such as TACTO~\cite{wang2022tacto} and
Taxim~\cite{si2022taxim} treat the object as rigid: they render the gel deformation from the penetrating geometry alone.
Used as the touch map, such a simulator realizes the rigid limit
\(K_o \to \infty\). For an object of stiffness \(K_o\), it overestimates the gel
indentation by the factor \(1/\kappa = 1 + K_g/K_o\), which is large for
objects much softer than the gel.

\section{REAL-TO-SIM-TO-REAL EXAMPLE}
\label{sec:r2s2r}

Consider a gripper with \digit{} sensors~\cite{Lambeta2020DIGIT} at its
fingertips that grasps a raw egg under force control. From the signals of
these sensors, \sparsh{}~\cite{higuera_sparsh_2024} estimates the contact
force that the controller regulates. \rgbd{} cameras observe the scene.
During data collection, the real touch sensors record the touch data that
train the touch map. At deployment, virtual touch sensors at the same poses
replace them: the electronics are removed, but the gel stays, so the egg
still presses against the same soft surface with the same force. \sparsh{}
then estimates the force from the predicted touch signal instead of the
measured one.

This section realizes rendering and touch prediction, \(\Phit \circ \Phir\),
for such a setting in two parts. The \emph{real-to-sim} part reconstructs the
object from real depth measurements as explicit geometry and renders it in
simulation from the pose of the touch sensor. The \emph{sim-to-real} part
turns this rendering, through the touch map, into the signal that the real
touch sensor would report. We define the maps of this realization, show that
its rendering is invariant to the coordinate frame, and bound its depth-patch
error.

\subsection{Construction}
\label{sec:r2s2r-construction}

The realization consists of five maps (Fig.~\ref{fig:pipeline}, bottom):
four real-to-sim maps that realize \(\Phir\), and one sim-to-real map that
realizes \(\Phit\).

\textbf{Lifting.} A measurement \(\Img = \{\mathbf{y}_k\}_{k \in \mathcal{K}}\),
indexed by a finite set \(\mathcal{K}\), consists of measured points
\(\mathbf{y}_k \in \R^3\) in the coordinate frame of the depth sensor, optionally with
appearance such as color. For a camera with intrinsics \(\mathbf{K}\)
and depth image \(d\), each pixel \(\mathbf{q} = (q_1, q_2)\) gives
\(\mathbf{y} = d(\mathbf{q})\,\mathbf{K}^{-1}(q_1, q_2, 1)^{\top}\). A segmentation of the object selects the subset
\(\mathcal{K}_o \subseteq \mathcal{K}\) of measurements on the object. The \emph{partial point cloud} is
\begin{equation}
  \Objp = \Phil(\Img, \Cam) = \bigl\{ \mathbf{R}_{r} \mathbf{y}_k
  + \mathbf{t}_{r} : k \in \mathcal{K}_o \bigr\}.
\end{equation}
The lifted points lie, up to the depth precision of the depth sensor, on
the part of the object
surface visible from the depth sensor. This part rarely contains the touched
region: at contact, the touched side is likely hidden by the object itself and
by the touch sensor or gripper.

\textbf{Completion.} Let \(\Surf_0\) be the object surface at the reference
instant. A \emph{shape completion} map \(\Phic\) reconstructs it in full from
the reference measurement, in which the object is unoccluded: the
\emph{reference shape}
\begin{equation}
  \SurfH_0 = \Phic\bigl(\Phil(\Imgz, \Cam), \Imgz\bigr)
\end{equation}
is a closed surface that estimates \(\Surf_0\), with \emph{completion error}
\(\eta_0 = d_H(\SurfH_0, \Surf_0)\), the Hausdorff distance.

\textbf{Tracking.} Between the reference and the query instant, the rigid
object moves by an unknown rigid motion
\(\mathcal{T}_o = (\mathbf{R}_o, \mathbf{t}_o) \in \mathrm{SE}(3)\), applied to
points as \(\mathbf{x} \mapsto \mathbf{R}_o\mathbf{x} + \mathbf{t}_o\). Its
surface at the query instant is thus \(\Surf = \mathcal{T}_o\Surf_0\). A \emph{tracking} map
\(\Phitr\) estimates this motion from the measurement \(\Img\), which is
partial and likely occluded at contact,
\begin{equation}
  \hat{\mathcal{T}}_o = \Phitr\bigl(\Phil(\Img, \Cam), \SurfH_0\bigr).
\end{equation}
The \emph{tracked shape} is \(\SurfH = \hat{\mathcal{T}}_o\SurfH_0\).
Tracking leaves the residual motion
\(\Delta\mathcal{T}_o = \hat{\mathcal{T}}_o\mathcal{T}_o^{-1}\), which maps
the true surface \(\Surf\) to \(\hat{\mathcal{T}}_o\Surf_0\). Two numbers measure
this residual motion, the \emph{tracking error}: its rotation angle
\(\theta_o\), and the distance
\(\tau_o = \norm{\Delta\mathcal{T}_o\mathbf{c} - \mathbf{c}}\) by which it
moves the centroid \(\mathbf{c}\) of \(\Surf\). The rotation moves points in
proportion to their distance from \(\mathbf{c}\), which is at most the
object's radius
\(\ell_o = \max_{\mathbf{x} \in \Surf} \norm{\mathbf{x} - \mathbf{c}}\).
The tracking map can be realized, e.g., by a model-based 6D
tracker~\cite{wen2024foundationpose} or by registering the visible points to
\(\SurfH_0\)~\cite{park2017colored}; once the object is rigidly grasped,
\(\hat{\mathcal{T}}_o\) can also follow the gripper through the robot's forward
kinematics. If the object is unoccluded at the query instant, one may take
\(\Imgz = \Img\) and \(\hat{\mathcal{T}}_o\) the identity.

\textbf{Depth rendering.} The tracked shape is then sampled uniformly into
the \emph{final point cloud} \(\Obj \subset \SurfH\), quasi-uniform with
separation distance \(r\), and rendered from the touch sensor's viewing plane by
Definition~\ref{def:point-depth}. In summary, the realization renders
\(\Phir(\Config) = \Phid(\Obj, \Pose)\).

\textbf{Sim-to-real.} Finally, the touch map turns the rendered depth patch
into the touch signal of the real sensor, \(\Sg = \Phit(\hat{\Dp})\). Learned
from pairs of rendered depth patches and real touch signals, it learns what
the rendered, undeformed geometry lacks: how gel and object deform in contact,
and how the gel's deformation appears in the signal of the real sensor. It
learns the deformation of the object only on average over the training
objects, since geometry does not reveal their stiffness
(Section~\ref{sec:limits}). A tactile simulator with a learned
correction can take its place (Section~\ref{sec:theory}).

\subsection{Invariance to the Coordinate Frame}

The rendering map of this realization is invariant to the coordinate frame
when moving the input of shape completion and tracking by \(g\) also moves
their output by \(g\).

\begin{assumption}[Equivariant completion and tracking]
\label{ass:equivariant}
Completion and tracking follow every rigid motion \(g\):
\begin{equation}
  \begin{aligned}
    \Phic(g\Objp, \Imgz) &= g\,\Phic(\Objp, \Imgz), \\
    \Phitr(g\Objp, g\SurfH_0) &= g\,\Phitr(\Objp, \SurfH_0)\,g^{-1},
  \end{aligned}
\end{equation}
and the point cloud sampled from \(g\SurfH\) is \(g\Obj\).
\end{assumption}

\begin{lemma}[Invariance of real-to-sim-to-real rendering]
\label{lem:r2s2r_inv}
Under Assumption~\ref{ass:equivariant}, the rendering map of
Section~\ref{sec:r2s2r-construction} is invariant to the coordinate frame.
\end{lemma}

By Proposition~\ref{prop:equivariance}, the Geometric Force of this
realization is therefore equivariant to the coordinate frame.
Assumption~\ref{ass:equivariant} holds by construction when completion and
tracking are computed in the frame of the depth sensor and mapped to the
coordinate frame with \(\Cam\).

\subsection{Depth-Patch Error}
\label{sec:r2s2r-error}

The bounds of Section~\ref{sec:depth-error} apply to this realization with
the final point cloud in place of the measured points. The final point cloud
lies on the tracked shape. The tracked shape, in turn, differs from the true surface by the errors of completion and
tracking.

\begin{lemma}[Bound on the shape error from completion and tracking]
\label{lem:shape}
The tracked shape satisfies
\begin{equation}
  d_H(\SurfH, \Surf) \le \eta_0 + \tau_o + \theta_o\,\ell_o =: \eta_{\mathrm{shape}}.
\end{equation}
See Fig.~\ref{fig:pipeline}, bottom.
\end{lemma}

Both errors depend on the depth sensor, whose spacing and precision
determine the lifted points from which \(\SurfH_0\) is completed and
\(\hat{\mathcal{T}}_o\) is estimated; we take them as given.

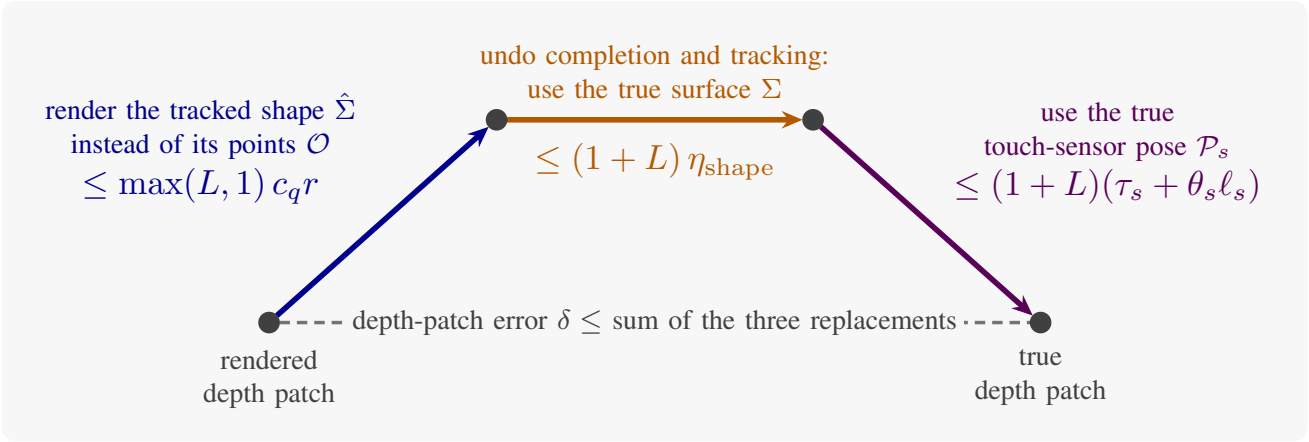
\begin{figure}[t]
\centering
\resizebox{\columnwidth}{!}{%
\begin{tikzpicture}[>={Stealth[length=6pt, width=5pt]},
    /utils/exec={%
      \colorlet{smpL}{blue!55!black}\colorlet{shpL}{orange!70!black}%
      \colorlet{posL}{violet!70!black}\colorlet{panelfill}{black!3}},
    pt/.style={circle, fill=black!75, inner sep=2.2pt},
    ptname/.style={font=\footnotesize, text=black!75, align=center},
    steplbl/.style={font=\footnotesize, align=center},
    bound/.style={font=\normalsize}]
  \fill[panelfill, rounded corners=4pt] (-2.3,-0.9) rectangle (10.9,3.55);
  \coordinate (A) at (0.4,0.3);
  \coordinate (B) at (2.7,2.35);
  \coordinate (C) at (5.9,2.35);
  \coordinate (D) at (8.2,0.3);
  \draw[->, smpL, line width=1.6pt, shorten >=3pt, shorten <=3pt] (A) -- (B);
  \node[steplbl, text=smpL, anchor=south east] at ($(A)!0.5!(B)+(-0.15,0.05)$)
    {render the tracked shape \(\SurfH\)\\instead of its points \(\Obj\)\\[2pt]{\normalsize\(\le \max(L,1)\,c_q r\)}};
  \draw[->, shpL, line width=1.6pt, shorten >=3pt, shorten <=3pt] (B) -- (C)
    node[midway, above=2pt, steplbl, text=shpL] {undo completion and tracking:\\use the true surface \(\Surf\)};
  \draw[->, posL, line width=1.6pt, shorten >=3pt, shorten <=3pt] (C) -- (D);
  \node[steplbl, text=posL, anchor=south west] at ($(C)!0.5!(D)+(0.15,0.05)$)
    {use the true\\touch-sensor pose \(\Pose\)\\[2pt]{\normalsize\(\le (1+L)(\tau_s + \theta_s\ell_s)\)}};
  \node[bound, text=shpL] at ($(B)!0.5!(C)+(0,-0.42)$) {\(\le (1+L)\,\eta_{\mathrm{shape}}\)};
  \draw[black!55, densely dashed, line width=1pt, shorten >=3pt, shorten <=3pt] (A) -- (D);
  \node[ptname, fill=panelfill, inner sep=2pt] at ($(A)!0.5!(D)$) {depth-patch error \(\delta \le\) sum of the three replacements};
  \node[pt] at (A) {};
  \node[pt] at (B) {};
  \node[pt] at (C) {};
  \node[pt] at (D) {};
  \node[ptname, anchor=north] at ($(A)+(0,-0.15)$) {rendered\\depth patch};
  \node[ptname, anchor=north] at ($(D)+(0,-0.15)$) {true\\depth patch};
\end{tikzpicture}}
\caption{Where the depth-patch bound of Corollary~\ref{cor:r2s2r} comes from. The
rendered depth patch uses the final point cloud \(\Obj\), sampled from the
tracked shape \(\SurfH\), seen from the estimated touch-sensor pose; the true depth patch uses the true surface, seen
from the true pose. Three replacements lead from one to the other, through
two intermediate depth patches: the points by the tracked shape; the tracked
shape by the true surface, which undoes completion and tracking; and the
estimated pose by the true one. Each replacement moves the depth patch by at most the bound shown.
By the triangle inequality, the depth-patch error (dashed) is at most their
sum.}
\label{fig:r2s2r_bound}
\end{figure}

\begin{corollary}[Bound on \(\delta\) for real-to-sim-to-real rendering]
\label{cor:r2s2r}
Let the realization of Section~\ref{sec:r2s2r-construction} render the final point cloud
\(\Obj \subset \SurfH\), quasi-uniform with separation distance \(r\), at an estimated
touch-sensor pose \(\PoseT\) with errors \((\theta_s, \tau_s)\), using
\(\rho = c_q r\). Let Assumption~\ref{ass:graph} hold for \(\Surf\) at
\(\Pose\) and \(\PoseT\), and for \(\SurfH\) at \(\PoseT\), with
a common slope bound \(L\) and \(\beta = \max(\rho, \eta_{\mathrm{shape}}, \tau_s + \theta_s\ell_s)\). Then
\begin{equation}
  \delta \le \max(L, 1)\,c_q\, r + (1 + L)\bigl(\eta_{\mathrm{shape}} + \tau_s +
  \theta_s\ell_s\bigr),
  \label{eq:delta_r2s2r}
\end{equation}
with \(\eta_{\mathrm{shape}}\) from Lemma~\ref{lem:shape}.
See Fig.~\ref{fig:pipeline}, bottom.
\end{corollary}

\begin{proof}
By the triangle inequality (Fig.~\ref{fig:r2s2r_bound}) through \(\Dp_{\SurfH,\PoseT}\) and
\(\Dp_{\Surf,\PoseT}\), apply Proposition~\ref{prop:sampling} to
\(\Obj \subset \SurfH\) with \(\epsilon = 0\), Lemmas~\ref{lem:shape}
and~\ref{lem:hausdorff} to \(\SurfH\) and \(\Surf\), and
Corollary~\ref{cor:pose}.
\end{proof}

Together with Theorem~\ref{thm:stability}, the force error of this realization
grows at most linearly with the separation distance of the final point cloud, the completion
error, the object-tracking error, and the touch-sensor pose error. Since the
final point cloud is sampled from the tracked shape, its separation distance \(r\) can be made small, leaving the error dominated by completion and tracking. The
object-tracking term \(\theta_o\ell_o\) grows with the object's size, so
rotation errors matter most for large objects. The sim-to-real map enters
the force error through \(L_T\) and \(\epsilon_T\) (Theorem~\ref{thm:stability}).

Only Proposition~\ref{prop:sampling} is specific to point clouds.
Lemma~\ref{lem:shape}, Lemma~\ref{lem:hausdorff}, and Corollary~\ref{cor:pose}
hold for any reconstructed surface, e.g., a mesh or a surface extracted from
a Gaussian splatting reconstruction~\cite{kerbl2023gaussian}. For such
representations, \eqref{eq:delta_r2s2r} holds with the sampling term replaced
by \((1 + L)\) times the Hausdorff distance between the rendered surface and
\(\SurfH\), by Lemma~\ref{lem:hausdorff}.

\section{PRACTICAL TAKEAWAYS}
\label{sec:implications}

The results guide a virtual touch sensor from design to deployment: how to
build its three maps, what the depth sensor and the objects must satisfy, how
to check the assumptions, and when the sensor still fails.

\subsection{Designing the Three Maps}

\textbf{Rendering.}
It needs only relative poses (Proposition~\ref{prop:equivariance}), so
the touch-sensor pose can come from the robot's forward kinematics. A surface
hidden at contact can be reconstructed once per object, from a close-range,
unoccluded measurement before contact, e.g., with a text-prompted
segmenter~\cite{carion2025sam3segmentconcepts} and a single-image 3D
reconstruction model~\cite{sam3dteam2025sam3d3dfyimages}. Afterwards, only its motion must be tracked.

\textbf{Touch map.}
It should receive the penetration field rather than the depth
patch: this removes phantom forces by construction
(Proposition~\ref{prop:gated}). It must localize the contact and predict,
pixel by pixel, how the contact changes the signal, which suits dense
image-to-image networks such as a U-Net~\cite{ronneberger2015unet}. A learned
touch map approximates the conditional mean signal of its training data
(Proposition~\ref{prop:irreducible}), so it should be trained on real touch
data from the target objects. Pressing each training object gradually, from
first contact to the end of its elastic range, covers every penetration depth
without damaging the object. The same sweep checks
Assumption~\ref{ass:winkler}. A rigid tactile simulator, in contrast,
needs a learned correction, since it overestimates the indentation of soft
objects (Proposition~\ref{prop:factorization}).

\textbf{Force map.}
It depends only on the sensor, so a pretrained decoder such as
\sparsh{}~\cite{higuera_sparsh_2024} can be used without retraining.

\subsection{Requirements}

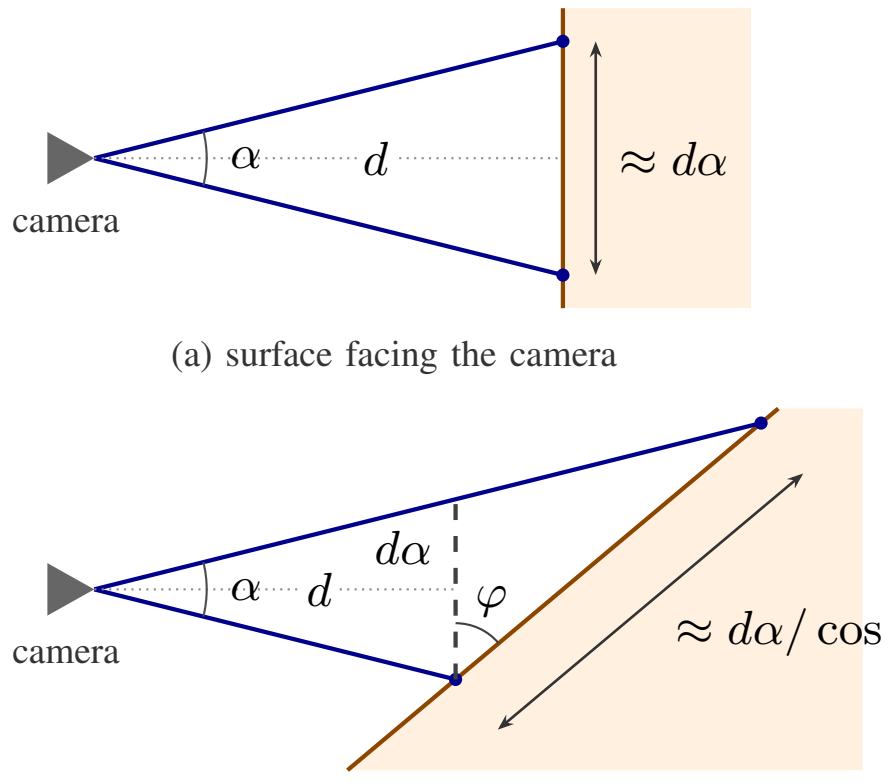
\begin{figure}[t]
\centering
\begin{tikzpicture}[scale=0.62, >=Stealth,
    math/.style={font=\small},
    word/.style={font=\scriptsize, text=black!80},
    dim/.style={{Stealth[length=3pt]}-{Stealth[length=3pt]}, line width=0.5pt,
                draw=black!80},
    /utils/exec={%
      \colorlet{objfill}{orange!12}\colorlet{objline}{orange!55!black}%
      \colorlet{ray}{blue!55!black}}]
  \begin{scope}
    \fill[objfill] (5,-1.6) rectangle (7,1.6);
    \draw[objline, thick] (5,-1.6) -- (5,1.6);
    \fill[black!60] (0,0) -- (-0.5,0.28) -- (-0.5,-0.28) -- cycle;
    \node[word, anchor=north] at (-0.3,-0.4) {camera};
    \draw[ray, thick] (0,0) -- (5,1.247);
    \draw[ray, thick] (0,0) -- (5,-1.247);
    \fill[ray] (5,1.247) circle (2pt);
    \fill[ray] (5,-1.247) circle (2pt);
    \draw[black!45, densely dotted] (0,0) -- (5,0);
    \draw[black!70] (1.164,-0.29) arc[start angle=-14, end angle=14, radius=1.2];
    \node[math, anchor=west] at (1.25,0) {\(\alpha\)};
    \node[math, fill=white, inner sep=1pt] at (3.0,0) {\(d\)};
    \draw[dim] (5.35,1.247) -- (5.35,-1.247);
    \node[math, anchor=west] at (5.4,0) {\(\approx d\alpha\)};
    \node[word, anchor=north] at (3.2,-1.75) {(a) surface facing the camera};
  \end{scope}
  \begin{scope}[yshift=-4.6cm]
    \fill[objfill] (2.702,-1.929) -- (7.298,1.929) -- (8.2,1.929) -- (8.2,-1.929) -- cycle;
    \draw[objline, thick] (2.702,-1.929) -- (7.298,1.929);
    \fill[black!60] (0,0) -- (-0.5,0.28) -- (-0.5,-0.28) -- cycle;
    \node[word, anchor=north] at (-0.3,-0.4) {camera};
    \draw[ray, thick] (0,0) -- (7.114,1.774);
    \draw[ray, thick] (0,0) -- (3.855,-0.961);
    \fill[ray] (7.114,1.774) circle (2pt);
    \fill[ray] (3.855,-0.961) circle (2pt);
    \draw[black!45, densely dotted] (0,0) -- (3.855,0);
    \draw[black!70] (1.164,-0.29) arc[start angle=-14, end angle=14, radius=1.2];
    \node[math, anchor=west] at (1.25,0) {\(\alpha\)};
    \node[math, fill=white, inner sep=1pt] at (2.4,0) {\(d\)};
    \draw[black!75, thick, dashed] (3.855,-0.961) -- (3.855,0.961);
    \node[math, anchor=east] at (3.8,0.45) {\(d\alpha\)};
    \draw[black!75] (3.855,-0.361) arc[start angle=90, end angle=40, radius=0.6];
    \node[math] at (4.24,-0.13) {\(\varphi\)};
    \draw[dim] (7.564,1.237) -- (4.305,-1.497);
    \node[math, anchor=west] at (6.0,-0.45) {\(\approx d\alpha/\cos\varphi\)};
    \node[word, anchor=north] at (3.2,-2.1) {(b) inclined surface};
  \end{scope}
\end{tikzpicture}
\caption{Spacing of measured points, which sets the fill distance \(\sigma\) in Corollary~\ref{cor:samples} and the camera condition~\eqref{eq:budget}. Two neighbouring pixel rays of a camera
diverge by its angular pixel pitch \(\alpha\). (a)~At distance \(d\), they hit
a surface facing the camera about \(d\alpha\) apart. (b)~On a surface
inclined by \(\varphi\), the gap \(d\alpha\) across the rays (dashed) and the
surface between the hits form a right triangle with angle \(\varphi\) between
them. Since \(\cos\varphi\) is the ratio of the gap to the hypotenuse, the
hits lie about \(d\alpha/\cos\varphi\) apart. Steep surfaces therefore call for
a closer camera: by~\eqref{eq:distance}, the largest working distance shrinks
with \(\cos\varphi\).}
\label{fig:spacing}
\end{figure}

\textbf{Depth sensor.}
The depth-patch error \(\delta\) must stay below two scales. By
Theorem~\ref{thm:stability}, it adds at most \(L_F L_T\,\delta\) to the force
error; by Proposition~\ref{prop:contact}, a penetration smaller than \(\delta\)
may be missed. A tolerated force error \(e\) and a smallest penetration
\(Z_{\min}\) to be resolved thus set the target
\(\delta^{*} = \min(e/(L_F L_T), Z_{\min})\). When the measured points are rendered directly, Corollary~\ref{cor:samples} with \(\rho = \sigma\) meets the target if
\begin{equation}
  \max(L, 1)\,\sigma + (1 + L)(\epsilon + \tau_s + \theta_s\ell_s) \le \delta^{*}.
  \label{eq:budget}
\end{equation}
How large is \(\sigma\) for a given camera? Neighbouring pixel rays diverge by
the camera's angular pixel pitch \(\alpha\), so at distance \(d\) they hit a
surface facing the camera about \(d\alpha\) apart. A surface whose normal is
inclined by \(\varphi\) to the viewing direction stretches this gap to about
\(d\alpha/\cos\varphi\) (Fig.~\ref{fig:spacing}), which bounds \(\sigma\).
Condition~\eqref{eq:budget} then limits the working distance:
\begin{equation}
  d \le \frac{\cos\varphi\,\bigl[\delta^{*} - (1 + L)(\epsilon + \tau_s +
  \theta_s\ell_s)\bigr]}{\alpha \max(L, 1)}.
  \label{eq:distance}
\end{equation}
For stereo and structured-light sensors, \(\epsilon\) itself grows with
\(d\). To choose a depth sensor and its placement:
\begin{enumerate}
  \item set \(\delta^{*}\) from the tolerated force error \(e\), the
    smallest penetration \(Z_{\min}\), and estimates of \(L_F\) and \(L_T\);
  \item estimate the slope \(L\) of the expected contact surfaces and the
    touch-sensor pose error \(\tau_s + \theta_s\ell_s\);
  \item check that the sampling precision \(\epsilon\) at the intended
    distance is small enough: the pose and precision terms of~\eqref{eq:budget}
    must use up only part of \(\delta^{*}\), leaving the rest for the point
    spacing;
  \item compute the largest working distance \(d\) from~\eqref{eq:distance},
    i.e., the farthest placement at which the camera's angular pixel pitch
    \(\alpha\) still gives fine enough spacing (Fig.~\ref{fig:spacing}).
\end{enumerate}
When the surface to be touched is hidden and reconstructed,
Corollary~\ref{cor:r2s2r} replaces \(\sigma\) and \(\epsilon\) by the
separation distance \(r\) of the final point cloud, which can be chosen
freely, and by the shape error \(\eta_{\mathrm{shape}}\). The depth sensor then
matters only through the completion and tracking errors.

\textbf{Objects.}
Good objects meet the touch sensor with gently sloped surfaces, since the
slope \(L\) multiplies every geometric error (Assumption~\ref{ass:graph}).
They need not be smooth: texture is fine as long as its slopes are gentle and
the depth sensor resolves it. They must stay rigid between the reference and
the query instant, as tracking assumes. Finally, they should be much stiffer
than the gel or all about equally stiff, which keeps the irreducible touch
error small (Corollary~\ref{cor:stiffness}). For objects both softer and
stiffer than the gel, such as fruit at different ripeness, the touch map needs the object category as an input, or a narrower training distribution (Proposition~\ref{prop:irreducible}).

\textbf{Fingertips.}
Both the touch map and the force map depend on the gel, through its
stiffness \(K_g\) and its surface \(z_0\) (Proposition~\ref{prop:factorization}).
The deployed fingertips must therefore keep the gel of the sensor that
recorded the training data; only the electronics can go. A different gel
requires new touch data.

\subsection{Checking the Assumptions}
\label{sec:checks}

Most assumptions can be checked before deployment.

\textbf{No contact, no force} (Assumptions~\ref{ass:nocontact}
and~\ref{ass:forcecal}). Render depth patches with the object away from the
sensor, so that \(\pen = 0\). The touch map should output zero; its largest
output on such patches measures the phantom touch signal. The force map
should output zero on the zero signal; otherwise, subtract \(\Phif(0)\) from
its output.

\textbf{Stability constants} (Theorem~\ref{thm:stability}). Estimate \(L_T\)
as the largest ratio of output change to input change over many pairs of
held-out depth patches, and \(L_F\) likewise from pairs of touch signals.
Since the true constants are the largest ratios over all pairs, these
estimates are lower bounds. The error of the touch map against real signals, by contrast, overestimates \(\epsilon_T\): it includes the irreducible touch error (Proposition~\ref{prop:irreducible}).

\textbf{Sampling and geometry} (Assumption~\ref{ass:graph}). If rendering
uses points, compute their separation distance \(r\) and fill distance
\(\sigma\), check that \(\sigma \le c_q r\) for a small \(c_q\), and choose
the splat radius \(\rho \ge \sigma\). Estimate \(L\) from finite differences
of the rendered depth patch, and flag pixels near silhouettes, where the
depth jumps to \(z_{\max}\) and the bounds fail.

\textbf{Elastic-foundation contact} (Assumption~\ref{ass:winkler}). Press
objects of known stiffness into the gel by known depths, and reconstruct the
gel indentation from the touch signal. The ratio of gel indentation to
penetration should match \(\kappa = K_o/(K_g + K_o)\) at every depth.
Indentation outside the penetrated pixels measures the neglected lateral
coupling.

\textbf{Coordinate-frame invariance} (Assumption~\ref{ass:equivariant}). Move
the coordinate frame by a random rigid motion, which moves both sensor poses,
and check that the rendered depth patch stays the same. A learned completion
map passes only if it works in the frame of the depth sensor or of the
object, not in an arbitrary world frame. Random sampling of the final point
cloud passes only in distribution, so the sampling should be fixed in the
object frame. Calibration errors are a different matter: they misplace the
touch sensor relative to the object, which the bounds already count as pose
error.

\subsection{Failure Modes}

A virtual touch sensor can fail in six situations:
\begin{itemize}
  \item \emph{Shallow contact.} A penetration smaller than \(\delta\)
    can be missed, and a surface less than \(\delta\) in front of the gel
    can appear to touch it
    (Proposition~\ref{prop:contact}).
  \item \emph{Hidden surfaces.} The error grows with the completion error and
    with the rotation error, which large objects amplify (Corollary~\ref{cor:r2s2r}).
  \item \emph{Edges and steep surfaces.} The bounds grow with the slope \(L\)
    and do not hold at silhouettes (Assumption~\ref{ass:graph}). Surfaces seen
    at a grazing angle are also sampled sparsely (Fig.~\ref{fig:spacing}).
  \item \emph{Objects of varying stiffness.} The irreducible touch error grows
    with the range of \(K_o/K_g\)
    (Corollary~\ref{cor:stiffness}).
  \item \emph{Unreliable learned maps.} Large Lipschitz constants amplify
    \(\delta\), and the errors \(\epsilon_T\) and \(\epsilon_F\) are small
    only on inputs like those seen in training (Theorem~\ref{thm:stability}).
  \item \emph{Effects outside the model.} Friction, shear, deforming
    objects, and fast or history-dependent contact
    (Section~\ref{sec:limitations}).
\end{itemize}

\section{LIMITATIONS}
\label{sec:limitations}

Our analysis is quasi-static. It evaluates the Geometric Force at a frozen
configuration, so it ignores dynamics, the viscoelasticity of soft objects,
and the hysteresis of gel and object, i.e., any dependence of the touch
signal on the history of the contact. A single depth measurement shows where
a surface is, not how it yields, so stiffness stays invisible; a sequence of
measurements, e.g., of the object flattening as the gripper closes, could
reveal it. We model only the normal force, without friction or shear. The
elastic-foundation model neglects lateral coupling in the gel, which spreads
deformation beyond the contact. With coupling, the touch map becomes
non-local and the factorization holds only approximately, although
Theorem~\ref{thm:stability} does not depend on this model. Finally, the
bounds for reconstruction-based rendering take the completion error
\(\eta_0\) as given rather than analyzing shape completion itself. Every
guarantee holds only as far as its assumptions do;
Section~\ref{sec:checks} shows how to check them.

\section{CONCLUSION}
\label{sec:conclusion}

We introduced the Geometric Force, the force on each taxel of a virtual touch
sensor, computed from depth measurements by rendering, touch prediction, and
force estimation. It is equivariant to the coordinate frame, vanishes out of
contact for contact-gated touch maps, and has an error that grows at most
linearly with the errors of its three maps. Its limit is the object's
stiffness: geometry does not reveal it, so no touch map can predict touch
exactly for objects of the same shape and different stiffness. In practice,
the depth sensor needs point spacing finer than the surface detail and depth
noise well below the gel indentation. Suitable objects keep their shape until
contact, have no sharp edges, and are much stiffer than the gel or all about
equally stiff.

Future work can test these results in experiments. For example, a rigid
tactile simulator should overestimate the gel indentation of soft objects by
the \mbox{factor}~\(1 + K_g/K_o\). Future work can also carry the Geometric
Force beyond grasping with standard grippers. One direction is learning from
humans: tracked geometry can label demonstrations with contact forces, which
today are recorded with devices worn by the
demonstrator~\cite{adeniji_feel_2025,liu_vitamin_2025}. Another is
prosthetics. Prosthetic hands rarely carry touch sensors, but a depth camera
on the prosthesis can supply the contact force, both for grip
control and for sensory feedback to the user~\cite{raspopovic2014restoring}.

\section*{APPENDIX: PROOFS}
\label{sec:appendix}

\makeatletter\@secpenalty=0\makeatother
\subsection*{Structural Properties and Stability}

\begin{proof}[Proof of Proposition~\ref{prop:equivariance}]
By Definition~\ref{def:geoforce}, the Geometric Force at a pixel is
\begin{equation*}
  \F(a,b) = \mathbf{R}\,\f(a,b),
\end{equation*}
where the force field \(\f\) in sensor coordinates is computed from the
rendered depth patch. A prime marks a
quantity after the motion \(g\). Since \(\Phir\) is frame-invariant, the
depth patch does not change, hence neither do \(\Sg\) and \(\f\). The motion
turns the touch-sensor rotation \(\mathbf{R}\) into
\(\mathbf{R}_g\mathbf{R}\), so
\begin{equation*}
  \F'(a,b) = \mathbf{R}_g\mathbf{R}\,\f(a,b) = \mathbf{R}_g\F(a,b).
\end{equation*}
The normal \(\mathbf{n}\), the last column of \(\mathbf{R}\), likewise
becomes \(\mathbf{R}_g\mathbf{n}\). Hence the normal component
\(\fld{F}^{\perp} = \F \cdot \mathbf{n}\) is unchanged:
\begin{equation*}
  \fld{F}'^{\perp}(a,b) = \bigl(\mathbf{R}_g\F(a,b)\bigr)\cdot(\mathbf{R}_g\mathbf{n})
  = \fld{F}^{\perp}(a,b).
  \qedhere
\end{equation*}
\end{proof}
\smallskip

\begin{proof}[Proof of Theorem~\ref{thm:stability}]
The estimate \(\fhat = \Phif\Phit(\hat{\Dp})\) differs from the ideal force
\(\fstar = \Phif^{*}\Phit^{*}(\Dp^{*})\) in the depth patch and in the two
maps. Replace them by their ideal counterparts, one at a time
(Fig.~\ref{fig:triangle}). Each replacement contributes one term:
\begin{equation*}
  \begin{aligned}
    T_1 &= \ninf{\Phif\Phit(\hat{\Dp}) - \Phif\Phit(\Dp^{*})},\\
    T_2 &= \ninf{\Phif\Phit(\Dp^{*}) - \Phif\Phit^{*}(\Dp^{*})},\\
    T_3 &= \ninf{\Phif\Phit^{*}(\Dp^{*}) - \Phif^{*}\Phit^{*}(\Dp^{*})}.
  \end{aligned}
\end{equation*}
By the triangle inequality, the force error is at most the sum of the three terms.
The two Lipschitz bounds then give
\begin{equation*}
  \begin{aligned}
    T_1 &\le L_F\ninf{\Phit(\hat{\Dp}) - \Phit(\Dp^{*})} \\
    &\le L_F L_T\,\ninf{\hat{\Dp} - \Dp^{*}} = L_F L_T\,\delta.
  \end{aligned}
\end{equation*}
Likewise, with the definition of \(\epsilon_T\),
\begin{equation*}
  T_2 \le L_F\ninf{\Phit(\Dp^{*}) - \Phit^{*}(\Dp^{*})} \le L_F\,\epsilon_T.
\end{equation*}
The third term compares the two force maps on the same touch signal
\(\Phit^{*}(\Dp^{*})\), so \(T_3 \le \epsilon_F\). Adding the three bounds
gives~\eqref{eq:stability}. The normal component is never longer than the
force vector, so the bound holds for it too.
\end{proof}
\subsection*{Depth-Patch Error}

\begin{proof}[Proof of Proposition~\ref{prop:sampling}]
Fix a pixel \((a,b)\) and let \(\mathbf{p}\) be the position of its
center on the viewing plane. By
Assumption~\ref{ass:graph}, its true depth is
\(\Dp_{\Surf,\Pose}(a,b) = \gamma(\mathbf{p})\). A sample point
\(\mathbf{o}_i\) has position \(\bar{\pi}_{\Pose}(\mathbf{o}_i)\) on the
viewing plane and depth \(z_i\). By Definition~\ref{def:point-depth}, the ray
of the pixel hits \(\mathbf{o}_i\) if these two positions are within the
splat radius \(\rho\) of each other:
\begin{equation*}
  \norm{\bar{\pi}_{\Pose}(\mathbf{o}_i) - \mathbf{p}} \le \rho.
\end{equation*}
The rendered depth \(\Dp_{\Obj,\Pose}(a,b)\) is the smallest depth
within the sensing range \([0, z_{\max}]\) among the sample points that
the ray hits, or \(z_{\max}\) if there is no such depth. We bound the rendered depth from both
sides.

\emph{Lower bound.} Let \(\mathbf{o}_i\) be a sample point that the ray hits.
Let a surface point nearest to \(\mathbf{o}_i\) have position
\(\mathbf{q}\) on the viewing plane and depth \(z\). By the definition of the
sampling precision, this surface point lies within \(\epsilon\) of
\(\mathbf{o}_i\), so
\begin{equation*}
  \norm{\mathbf{q} - \bar{\pi}_{\Pose}(\mathbf{o}_i)} \le \epsilon, \qquad
  \lvert z - z_i \rvert \le \epsilon.
\end{equation*}
Since the ray hits \(\mathbf{o}_i\), the position \(\mathbf{q}\) is within
\(\rho + \epsilon = \beta\) of \(\mathbf{p}\). It thus lies in the window
\(W_\beta\), where Assumption~\ref{ass:graph} holds. The ray from
\(\mathbf{q}\) passes through this surface point, so it first hits the surface
at a depth \(\gamma(\mathbf{q}) \le z\). By the slope bound \(L\), this depth is at most
\(L(\rho + \epsilon)\) below \(\gamma(\mathbf{p})\). Therefore
\begin{equation}
  z_i \ge z - \epsilon \ge \gamma(\mathbf{q}) - \epsilon
  \ge \gamma(\mathbf{p}) - L(\rho + \epsilon) - \epsilon.
  \label{eq:sample-depth}
\end{equation}
The rendered depth \(\Dp_{\Obj,\Pose}(a,b)\) is either the depth \(z_i\) of
such a sample point or \(z_{\max}\), which is at least \(\gamma(\mathbf{p})\).
Hence
\begin{equation*}
  \Dp_{\Obj,\Pose}(a,b) \ge \gamma(\mathbf{p}) - L\rho - (1 + L)\epsilon.
\end{equation*}

\emph{Upper bound.} The surface point with position \(\mathbf{p}\) on the
viewing plane and depth \(\gamma(\mathbf{p})\) belongs to \(\Surf_{\Pose}\),
the part of the surface seen by the touch sensor. By the definition of the
fill distance, some sample point \(\mathbf{o}_i\) lies within \(\sigma\) of
this surface point, so
\begin{equation*}
  \norm{\bar{\pi}_{\Pose}(\mathbf{o}_i) - \mathbf{p}} \le \sigma \le \rho, \qquad
  z_i \le \gamma(\mathbf{p}) + \sigma.
\end{equation*}
The ray of the pixel thus hits \(\mathbf{o}_i\). Its depth \(z_i\) is not
negative, by~\eqref{eq:sample-depth} and the assumption
\(\gamma \ge \epsilon\) on \(W_\beta\). If this depth lies within the
sensing range, it is one of the depths among which the rendered depth is the
smallest. Otherwise, it exceeds \(z_{\max}\), whereas the rendered depth
does not.
In both cases, the rendered depth is at most \(z_i\), so
\begin{equation*}
  \Dp_{\Obj,\Pose}(a,b) \le z_i \le \gamma(\mathbf{p}) + \sigma.
\end{equation*}

Together, the two bounds give~\eqref{eq:sampling} at every pixel. For
quasi-uniform sample points on the surface, \(\epsilon = 0\) and
\(\sigma \le c_q r = \rho\), so the bound becomes
\begin{equation*}
  \max(L\rho, \sigma) \le \max(L, 1)\,c_q r.
  \qedhere
\end{equation*}
\end{proof}
\smallskip

\begin{proof}[Proof of Lemma~\ref{lem:hausdorff}]
Fix a pixel \((a,b)\) and let \(\mathbf{p}\) be the position of its center on
the viewing plane. By Assumption~\ref{ass:graph}, the ray from a position in
the window \(W_\eta\) first hits \(\Surf\) at the depth \(\gamma\) and
\(\tilde\Surf\) at the depth \(\tilde\gamma\) of this position. The two
depths of the pixel are thus \(\gamma(\mathbf{p})\) and
\(\tilde\gamma(\mathbf{p})\).

The ray of the pixel first hits \(\tilde\Surf\) at the point with position
\(\mathbf{p}\) and depth \(\tilde\gamma(\mathbf{p})\). Since the two surfaces are within Hausdorff distance \(\eta\), a point of
\(\Surf\) lies within \(\eta\) of this point. Let it have position \(\mathbf{q}\) on the viewing
plane and depth \(z\), so
\begin{equation*}
  \norm{\mathbf{q} - \mathbf{p}} \le \eta, \qquad
  \lvert z - \tilde\gamma(\mathbf{p}) \rvert \le \eta.
\end{equation*}
The position \(\mathbf{q}\) thus lies in \(W_\eta\). The ray from
\(\mathbf{q}\) passes through this point of \(\Surf\), so it first hits
\(\Surf\) at a depth \(\gamma(\mathbf{q}) \le z\). The slope bound \(L\) keeps this depth within \(L\eta\) of
\(\gamma(\mathbf{p})\), so
\begin{equation*}
  \tilde\gamma(\mathbf{p}) \ge z - \eta \ge \gamma(\mathbf{q}) - \eta
  \ge \gamma(\mathbf{p}) - (1 + L)\eta.
\end{equation*}
Exchanging the roles of \(\Surf\) and \(\tilde\Surf\) gives the reverse
inequality. Hence the two depths differ by at most \((1 + L)\eta\).
\end{proof}
\smallskip

\begin{proof}[Proof of Corollary~\ref{cor:pose}]
Let \(\mathbf{y} = \pi_{\Pose}(\mathbf{x})\) be the sensor coordinates of a
surface point \(\mathbf{x}\) at the true pose. At the estimated pose, its
sensor coordinates are
\begin{equation*}
  \pi_{\PoseT}(\mathbf{x}) = \mathbf{Q}\mathbf{y} + \mathbf{d},
  \quad \mathbf{Q} = \hat{\mathbf{R}}^{\top}\mathbf{R},
  \quad \mathbf{d} = \hat{\mathbf{R}}^{\top}(\mathbf{t} - \hat{\mathbf{t}}).
\end{equation*}
Viewed from a sensor held in place, the pose error thus moves the surface
point from \(\mathbf{y}\) to \(\mathbf{Q}\mathbf{y} + \mathbf{d}\), by the
distance
\begin{equation*}
  \norm{(\mathbf{Q} - \mathbf{I})\mathbf{y} + \mathbf{d}}
  \le \norm{(\mathbf{Q} - \mathbf{I})\mathbf{y}} + \norm{\mathbf{d}}
  \le \theta_s\ell_s + \tau_s = \eta.
\end{equation*}
The first term is at most \(\theta_s\ell_s\), since \(\mathbf{Q}\) rotates by
the angle \(\theta_s\) and the point lies within \(\ell_s\) of the sensor
origin. The second is the translation error \(\tau_s\).
Such a motion changes the surface by at most \(\eta\) in Hausdorff distance, so Lemma~\ref{lem:hausdorff} applies.
\end{proof}
\subsection*{Stiffness Limit}

\begin{proof}[Proof of Proposition~\ref{prop:factorization}]
By Assumption~\ref{ass:winkler}, at every pixel,
\begin{equation*}
  \pen = \pen_g + \pen_o, \qquad K_g\pen_g = K_o\pen_o.
\end{equation*}
Eliminating the object indentation \(\pen_o\) gives
\begin{equation*}
  \pen = \Bigl(1 + \frac{K_g}{K_o}\Bigr)\pen_g, \quad \text{i.e.,} \quad
  \pen_g = \kappa\pen.
\end{equation*}
The pressure on the gel is \(K_g\pen_g\). Multiplied by the pixel area
\(\lambda^2\), it is the normal force on the pixel:
\begin{equation*}
  \fld{F}^{\perp} = \lambda^2 K_g\pen_g = \lambda^2 K_g\kappa\,\pen
  = \lambda^2 K_{\mathrm{eff}}\pen.
\end{equation*}
Now let \(K_o\) be random and independent of \(\Dp\). Given
\(\Dp\), the penetration \(\pen\) is fixed, so only \(\kappa\) is random in
the touch signal \(\Sg = \kappa\pen\). With \(\pen\) read as a vector,
\begin{equation*}
  \mathbb{E}[\Sg \mid \Dp] = \mathbb{E}[\kappa]\,\pen, \qquad
  \operatorname{Cov}(\Sg \mid \Dp) = \operatorname{Var}(\kappa)\,\pen\pen^{\top}.
\end{equation*}
The trace of \(\pen\pen^{\top}\) is \(\norm{\pen}^2\), which gives the stated
error. The ideal force map multiplies the touch signal by \(\lambda^2 K_g\).
Applied to the mean, it gives the stated force, since \(K_{\mathrm{eff}} = K_g\kappa\). Finally,
\(\kappa \to 1\) as \(K_o \to \infty\).
\end{proof}
\smallskip

\begin{proof}[Proof of Corollary~\ref{cor:stiffness}]
By Proposition~\ref{prop:factorization}, the error given \(\Dp\) is the
variance of \(\kappa\) times the squared penetration. Since \(\kappa\) is
independent of \(\Dp\), the expectation over \(\Dp\) only replaces the
squared penetration by its mean. The ratio \(\kappa\) increases with
\(K_o\), so it stays between \(\kappa_-\) and \(\kappa_+\). A random
variable confined to an interval has variance at most a quarter of the
squared length of that interval.
Finally,
\begin{equation*}
  \kappa_+ - \kappa_- = \frac{K_g(K_+ - K_-)}{(K_g + K_-)(K_g + K_+)}
  \le \frac{K_g}{K_g + K_-},
\end{equation*}
since \(K_+ - K_- \le K_g + K_+\).
\end{proof}
\subsection*{Real-to-Sim-to-Real Example}

\begin{proof}[Proof of Lemma~\ref{lem:r2s2r_inv}]
We show that moving the coordinate frame by \(g\) leaves the rendered depth
patch unchanged. The motion \(g\) moves the depth-sensor pose, so the lifted
points move by \(g\) as well:
\begin{equation*}
  \Phil(\Img, g\Cam) = \bigl\{\mathbf{R}_g\mathbf{R}_{r} \mathbf{y}_k
  + \mathbf{R}_g\mathbf{t}_{r} + \mathbf{t}_g\bigr\} = g\,\Phil(\Img, \Cam),
\end{equation*}
and likewise for the reference measurement. By
Assumption~\ref{ass:equivariant}, completion, tracking, and sampling follow
suit, so the final point cloud becomes \(g\Obj\). Rendering at the moved
touch-sensor pose \(g\Pose\) removes \(g\) again: every point
\(\mathbf{x}\) has the same sensor coordinates as before the motion,
\begin{equation*}
  \pi_{g\Pose}(g\mathbf{x}) = (\mathbf{R}_g\mathbf{R})^{\top}
  (\mathbf{R}_g\mathbf{x} + \mathbf{t}_g - \mathbf{R}_g\mathbf{t} - \mathbf{t}_g)
  = \pi_{\Pose}(\mathbf{x}).
  \qedhere
\end{equation*}
\end{proof}
\smallskip

\begin{proof}[Proof of Lemma~\ref{lem:shape}]
The tracked shape is \(\SurfH = \hat{\mathcal{T}}_o\SurfH_0\) and the true
surface is \(\Surf = \mathcal{T}_o\Surf_0\). By the triangle inequality,
\begin{equation*}
  d_H(\hat{\mathcal{T}}_o\SurfH_0, \mathcal{T}_o\Surf_0) \le
  d_H(\hat{\mathcal{T}}_o\SurfH_0, \hat{\mathcal{T}}_o\Surf_0) +
  d_H(\hat{\mathcal{T}}_o\Surf_0, \mathcal{T}_o\Surf_0),
\end{equation*}
where \(\hat{\mathcal{T}}_o\Surf_0\) is the true reference surface under
the estimated motion. The first term is the completion error
\(\eta_0 = d_H(\SurfH_0, \Surf_0)\), since a rigid motion preserves
distances. The second term compares the true surface \(\Surf\) with its image
\(\hat{\mathcal{T}}_o\Surf_0 = \Delta\mathcal{T}_o\Surf\) under the residual
motion \(\Delta\mathcal{T}_o\), whose rotation we write as
\(\Delta\mathbf{Q}\). Let \(\mathbf{c}\) be the centroid of the true surface.
The residual motion moves every point \(\mathbf{y}\) of \(\Surf\) by
\begin{equation*}
  \begin{aligned}
    \norm{\Delta\mathcal{T}_o\mathbf{y} - \mathbf{y}}
    &\le \norm{(\Delta\mathbf{Q} - \mathbf{I})(\mathbf{y} - \mathbf{c})} +
    \norm{\Delta\mathcal{T}_o\mathbf{c} - \mathbf{c}} \\
    &\le \theta_o\ell_o + \tau_o.
  \end{aligned}
\end{equation*}
The first term is at most \(\theta_o\ell_o\), since \(\Delta\mathbf{Q}\)
rotates by the angle \(\theta_o\) and the point lies within \(\ell_o\) of
the centroid. The second term is the
definition of \(\tau_o\). Such a motion changes the surface by at most \(\theta_o\ell_o + \tau_o\) in
Hausdorff distance.
\end{proof}

\section*{ACKNOWLEDGMENT}
The author thanks Prasanna Bhat for work on practical experimentation
related to this theory. The theory
grows out of four years of the author's research on artificial touch, including a short
Geo-Force manuscript, the OnTouch code and experiments, and
Imagine2touch~\cite{ayad_imagine2touch_2024}. Starting from these, Claude (Anthropic), an AI assistant, drafted the
text. The author then set its structure, shaped its meaning, flow, and style word by
word, and verified every claim. Later, the author
specified the figures, Claude drew them, and the author refined every
detail. The
book-like layout, with colored boxes for definitions, assumptions, and
results, is the author's design. The author takes full responsibility for the content of this paper.

\bibliographystyle{IEEEtran}
\bibliography{refs}

\end{document}